%% file: main.tex
\documentclass[twocolumn]{galbot}
\usepackage{wrapfig}
\usepackage[nospread]{flushend}
\usepackage{textcomp}
\usepackage{stfloats}
\usepackage{url}
\usepackage{verbatim}
\usepackage{graphicx}
\usepackage{titlesec}
\usepackage{tocloft}
\usepackage{adjustbox}
\usepackage{multirow}
\usepackage{tikz}
\usepackage{comment}
\usepackage{amsmath,amssymb}
\usepackage{colortbl}
\usepackage{color}
\usepackage{booktabs} 
\usepackage{hyperref}
\usepackage{subcaption} 
\RequirePackage{xspace}
\makeatletter
\DeclareRobustCommand\onedot{\futurelet\@let@token\@onedot}
\def\@onedot{\ifx\@let@token.\else.\null\fi\xspace}

\makeatother

\usepackage{makecell}

\usepackage{pifont}
\usepackage{bbding}
\usepackage{fontawesome}
\usepackage{xspace}

\usepackage{float}

\newlength\savewidth

\newcolumntype{x}[1]{>{\centering\arraybackslash}p{#1pt}}
\newcolumntype{y}[1]{>{\raggedright\arraybackslash}p{#1pt}}
\newcolumntype{z}[1]{>{\raggedleft\arraybackslash}p{#1pt}}

\renewcommand{\paragraph}[1]{\vspace{1mm}\noindent\textbf{#1}}

\usepackage{xcolor}
\usepackage{array}
\usepackage{bbm}
\usepackage{collcell,xfp}
\usepackage{pgf}
\usepackage[most]{tcolorbox}
\usepackage{csquotes}
\usepackage[noorphans,vskip=1em,leftmargin=1em]{quoting}
\usepackage{enumitem} 
\usepackage{forest}
\usepackage{caption}
\usepackage{longtable}
\usepackage[T1]{fontenc}

\usepackage{needspace}
\renewcommand{\paragraph}[1]{\par\addvspace{1mm}\Needspace*{2\baselineskip}\phantomsection\noindent\textbf{#1}}

\usepackage{algorithm}
\usepackage{listings}

\definecolor{codeblue}{rgb}{0.25, 0.5, 0.5}
\definecolor{codekw}{rgb}{0.35, 0.35, 0.75}
\lstdefinestyle{Pytorch}{
    language = Python,
    backgroundcolor = \color{white},
    basicstyle = \fontsize{9pt}{8pt}\selectfont\ttfamily\bfseries,
    columns = fullflexible,
    aboveskip=1pt,
    belowskip=1pt,
    breaklines = true,
    captionpos = b,
    commentstyle = \color{codeblue},
    keywordstyle = \color{codekw},
}

\definecolor{green}{HTML}{009000}
\definecolor{red}{HTML}{ea4335}

\definecolor{linecolor1}{RGB}{246, 248, 239}
\definecolor{linecolor2}{RGB}{230, 234, 217}
\definecolor{linecolor3}{RGB}{211, 222, 190}

\newcommand{\astra}{GPT-6 Astra\xspace}
\newcommand{\pifive}{\ensuremath{\pi_{0.5}}\xspace}

\newcommand{\navdash}{\textendash}
\AtEndEnvironment{figure}{\par\vspace{8pt}}
\AtEndEnvironment{figure*}{\par\vspace{8pt}}
\AddToHook{cmd/section/before}{\Needspace*{4\baselineskip}}

\titleformat*{\section}{\Large\sffamily\bfseries\raggedright}
\titleformat*{\subsection}{\large\sffamily\bfseries\raggedright}
\titlespacing*{\section}{0pt}{2ex minus .2ex}{1.5ex}
\titlespacing*{\subsection}{0pt}{2ex minus .2ex}{1ex}
\titlespacing*{\subsubsection}{0pt}{1.5ex minus .2ex}{1ex}
\input{table-style}
\usepackage{catchfile}

\usepackage{fix-cm}
\hypersetup{colorlinks=true,linkcolor=galbot,citecolor=galbot,urlcolor=galbotnavy,
  pdftitle={Systematically Exploring the Capabilities of GPT-6 Astra as Embodied Policies},
  pdfsubject={Six-domain evaluation of embodied reasoning, policy cooperation, and control limitations},
  pdfkeywords={GPT-6 Astra, embodied intelligence, capability boundaries, manipulation, HumanoidBench}}

\title{Systematically Exploring the Capabilities of\\GPT-6 Astra as Embodied Policies}

\author{Galbot Team}

\date{September 2026}
\code{\href{https://github.com/anonymous-report-421/GPT-as-Policy}{GPT-as-Policy}}
\page{\href{https://galaxygeneralrobotics.github.io/astra-policy/}{Project page}}

\input{sec/0_abstract}

\begin{document}
\maketitle
\pagestyle{plain}
\setlength{\parskip}{4pt plus 1pt minus .5pt}
\makeatletter
\def\@textbottom{\vskip\z@\@plus2pt}
\makeatother

\input{sec/1_intro}
\input{sec/2_framework}
\input{sec/4_experiments}

\input{sec/4_analysis}
\input{sec/5_conclusion}
\input{sec/6_contributors}

\raggedcolsend
\setlength{\bibsep}{0pt plus .2pt}
\bibliographystyle{assets/plainnat}
\bibliography{main}
\flushcolsend

\clearpage
\raggedcolsend
\appendix
\setlength{\parskip}{2pt plus .5pt}
\input{sec/3_method}
\input{sec/X_suppl}
\input{sec/Z_humanoid_suppl}
\input{sec/Y_harness_suppl}
\input{sec/Z_dexterous_suppl}
\input{sec/Z_dexterous_suppl1}

\input{sec/Z_navigation_mobile_suppl}

\raggedbottom
\end{document}

%% file: table-style.tex
\definecolor{tablehead}{HTML}{E3EFE9}
\definecolor{tablestripe}{HTML}{F4F6F5}
\definecolor{tablesummary}{HTML}{EAF3EE}
\definecolor{tablegroup}{HTML}{F0F3F1}
\definecolor{tablerule}{HTML}{61776C}
\newcommand{\reporthead}[1]{{\sffamily\bfseries #1}}
\newcommand{\reporttablestyle}{%
  \small
  \renewcommand{\arraystretch}{1.12}%
  \setlength{\extrarowheight}{0.6pt}%
  \setlength{\tabcolsep}{6pt}%
  \setlength{\heavyrulewidth}{0.6pt}%
  \setlength{\lightrulewidth}{0.3pt}%
  \setlength{\cmidrulewidth}{0.3pt}%
  \setlength{\aboverulesep}{0pt}%
  \setlength{\belowrulesep}{0pt}%
  \arrayrulecolor{tablerule}%
}
\AtBeginEnvironment{table}{\reporttablestyle}
\AtBeginEnvironment{table*}{\reporttablestyle}
\AtEndEnvironment{table}{\par\vspace{8pt}}
\AtEndEnvironment{table*}{\par\vspace{8pt}}

%% file: sec/0_abstract.tex
\abstract{
GPT-6 Astra exhibits a remarkable ability to generate numerical robot actions, extending its role beyond high-level planning.
To assess Astra's capabilities as general-purpose embodied policies, we conduct comprehensive evaluations across six domains, examining direct control, cooperation with learned policies, and feedback-driven adaptation.
In gripper manipulation, Astra can correct task targets and prepare contact conditions for subsequent policy execution; hybrid control with $\pi_{0.5}$ achieves 48\% success on the evaluated RoboDojo subset.
In dexterous manipulation, hybrid control achieves 50\% success in ten experience-guided DexJoCo trials, while direct in-hand control struggles to coordinate finger contacts.
In mobile manipulation, hybrid control reaches 38.7\% success on the evaluated RoboCasa365.
In navigation, Astra leads our local comparisons, reaching 92\% success on RxR instruction following and 82\% on HM3D object search, although search incurs substantial detours.
In locomotion, dense motion-reference generation remains unreliable: none of five sequential attempts on a single obstacle course reaches the goal, despite improvements in stability and forward progress.
In humanoid loco-manipulation, Astra exceeds baseline methods on 13 of 30 HumanoidBench tasks with pretrained whole-body controllers.
These findings reveal a gap between useful task decisions and reliable physical control.
Inference latency further constrains practical control: across 50 RoboDojo instances per condition, policy-assisted and direct control consume 624.8 million and 1.132 billion tokens. A 30-second locomotion run requires 250 model calls averaging 39.86 seconds each, with physics paused during inference.
}

%% file: sec/1_intro.tex
\section{Introduction}
\label{sec:intro}

Community experiments with Astra reveal possibilities ranging from 3D scene understanding and real-to-sim workflows~\cite{hkusail2026real2sim} to robot manipulation~\cite{zhang2026earlyastra}. These demonstrations motivate a systematic assessment of Astra as an embodied policy: \emph{Which tasks can it perform, where does it remain unreliable, and what resources do its decisions require?} The assessment must account for both successes and failures.

We evaluate \astra~\cite{openai2026astra} across six domains: gripper manipulation, dexterous manipulation, mobile manipulation, navigation, locomotion, and humanoid loco-manipulation. Our primary aim is to characterize performance and capability boundaries through task outcomes, comparisons with existing methods, failure analysis, and inference demands. We then analyze how interfaces, whole-body controllers, and feedback shape these outcomes.

Existing work spans spatial understanding~\cite{qi2024shapellm,jia2026omnispatial} and reasoning for robot actions~\cite{qi2025sofar}. Vision--language--action models such as \pifive map observations and instructions to actions learned from robot experience~\cite{pi05}. SayCan selects grounded skills~\cite{ahn2022saycan}, Code as Policies composes perception and control through programs~\cite{liang2023code}, and Prompt a Robot to Walk and Natural Language as Policies explore numerical feedback control~\cite{wang2024prompt,mikami2024natural}. Recent Astra evaluations report uneven manipulation performance across tasks~\cite{zhang2026earlyastra}. Our evaluation covers both Astra-generated robot commands, termed \emph{Direct}, and cooperation with learned action policies or whole-body controllers, termed \emph{Hybrid}. Figure~\ref{fig:reasoning-boundary} summarizes these control paths.

\input{figures/reasoning-boundary}

The domains expose complementary demands. Gripper and dexterous manipulation test grasping, local correction, and changing contact. Mobile manipulation adds base--arm coordination and multistage goals. Navigation tests spatial decisions and stopping, while locomotion requires coordinated body motion. Humanoid loco-manipulation combines task progression with pretrained whole-body controllers. We retain each study's metrics and protocols to make the scope of its conclusions explicit.

The results reveal substantial but uneven capabilities. Astra leads our local navigation comparisons on success and path efficiency, and Hybrid systems achieve higher aggregate performance in several manipulation studies. Yet direct in-hand control trails task-specific RL, dense-reference locomotion remains unreliable despite repeated attempts, and assistance can fail on cases completed by a standalone policy. Token use and inference latency qualify these outcomes and must be measured separately from motion duration.

To interpret these boundaries, we examine the division of control. Once a reasoning model can generate numerical actions, task success alone does not reveal which responsibilities it assumes, which capabilities interfaces and whole-body controllers supply, or how feedback connects them. In successful manipulation cases, Astra can prepare conditions for a policy's next action without generating the full trajectory. Conversely, preserving a grasp can impede required contact changes, and delegating motion can retain substantial proposal-review costs. This analysis helps explain the measured strengths and limitations while distinguishing Astra's decisions from the capabilities of the complete system.

%% file: figures/reasoning-boundary.tex
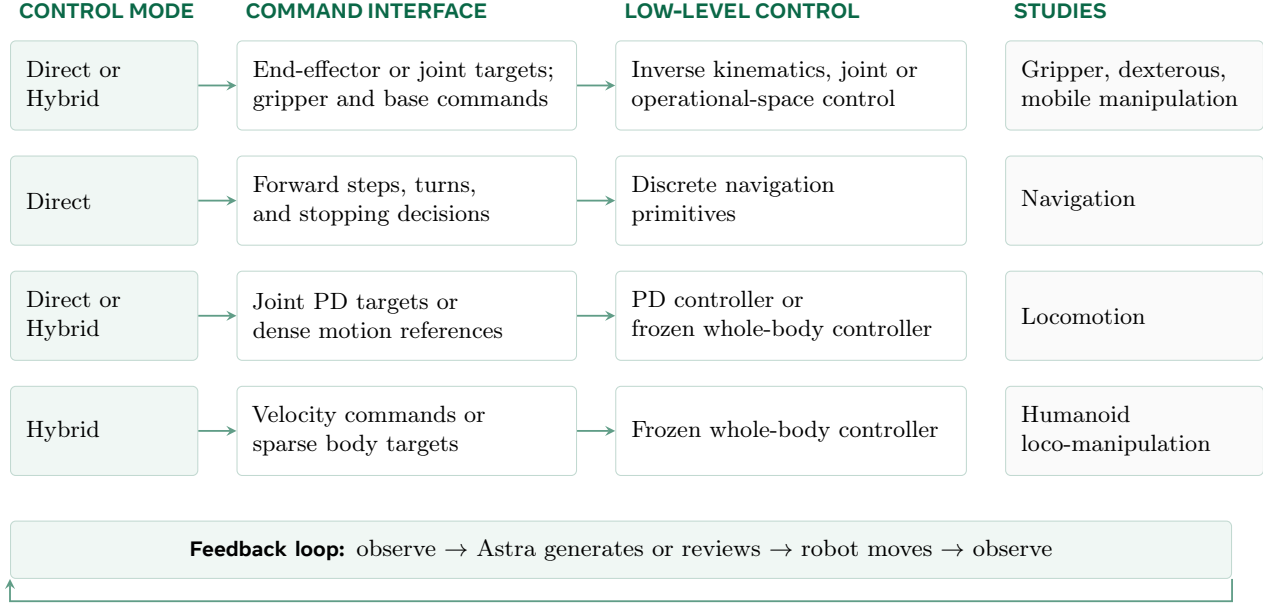
\begin{figure*}[t]
\centering
\resizebox{\textwidth}{!}{%
\begin{tikzpicture}[
  cell/.style={draw=galbot!25,rounded corners=2pt,minimum height=1.20cm,
    text width=4.15cm,inner sep=6pt,align=left,font=\small},
  source/.style={cell,text width=2.10cm,fill=galbotlight},
  support/.style={cell,text width=4.30cm},
  study/.style={cell,text width=3.05cm,fill=black!2},
  head/.style={font=\small\sffamily\bfseries,text=galbot,anchor=west},
  flow/.style={->,>=stealth,line width=.7pt,draw=galbot!65}]
\node[head] at (0,1.00) {CONTROL MODE};
\node[head] at (3.05,1.00) {COMMAND INTERFACE};
\node[head] at (8.15,1.00) {LOW-LEVEL CONTROL};
\node[head] at (13.40,1.00) {STUDIES};
\node[source,anchor=west] (a1) at (0,0) {Direct or\\Hybrid};
\node[cell,anchor=west] (b1) at (3.05,0)
  {End-effector or joint targets;\\gripper and base commands};
\node[support,anchor=west] (c1) at (8.15,0)
  {Inverse kinematics, joint or\\operational-space control};
\node[study,anchor=west] at (13.40,0)
  {Gripper, dexterous,\\mobile manipulation};
\node[source,anchor=west] (a2) at (0,-1.55) {Direct};
\node[cell,anchor=west] (b2) at (3.05,-1.55)
  {Forward steps, turns,\\and stopping decisions};
\node[support,anchor=west] (c2) at (8.15,-1.55)
  {Discrete navigation\\primitives};
\node[study,anchor=west] at (13.40,-1.55) {Navigation};
\node[source,anchor=west] (a3) at (0,-3.10) {Direct or\\Hybrid};
\node[cell,anchor=west] (b3) at (3.05,-3.10)
  {Joint PD targets or\\dense motion references};
\node[support,anchor=west] (c3) at (8.15,-3.10)
  {PD controller or\\frozen whole-body controller};
\node[study,anchor=west] at (13.40,-3.10) {Locomotion};
\node[source,anchor=west] (a4) at (0,-4.65) {Hybrid};
\node[cell,anchor=west] (b4) at (3.05,-4.65)
  {Velocity commands or\\sparse body targets};
\node[support,anchor=west] (c4) at (8.15,-4.65)
  {Frozen whole-body controller};
\node[study,anchor=west] at (13.40,-4.65) {Humanoid\\loco-manipulation};
\foreach \i in {1,2,3,4} {
  \draw[flow] (a\i.east) -- (b\i.west);
  \draw[flow] (b\i.east) -- (c\i.west);
}
\node[draw=galbot!25,fill=galbotlight,rounded corners=2pt,
  text width=15.95cm,inner sep=7pt,align=center,font=\small,
  anchor=west] (feedback) at (0,-6.25)
  {\textbf{Feedback loop:} observe $\rightarrow$ Astra generates or reviews $\rightarrow$ robot moves $\rightarrow$ observe};
\draw[flow] (feedback.south east) -- ++(0,-.30) -| (feedback.south west);
\end{tikzpicture}}
\caption{\textbf{Control modes, interfaces, and low-level control.} Direct uses Astra-generated commands with analytic control. Hybrid combines Astra with a learned task policy or whole-body controller: Astra reviews policy proposals or supplies commands or motion references to frozen controllers. Observations of the resulting motion inform subsequent decisions.}
\label{fig:reasoning-boundary}
\end{figure*}

%% file: sec/2_framework.tex
\section{Evaluation Framework}
\label{sec:method}

\subsection{How Astra Produces Robot Actions}
An \emph{action} is a command submitted to the robot's control interface, such as a target pose, a joint increment, or a navigation move. Astra receives the observations allowed by the study, selects an action, and uses the resulting feedback to choose the next one. The interface checks command bounds, and the environment determines task success. Tools support calculations, image inspection, and notes.

\emph{Direct} uses Astra-generated robot commands without a learned action policy or whole-body controller. Inverse kinematics and proportional--derivative control can implement these commands. \emph{Hybrid} adds a learned task policy or whole-body controller.

Hybrid takes two forms in this report. In manipulation, Astra accepts, modifies, or replaces a task policy's action proposals; RoboCasa also permits subgoal rewriting. In humanoid control, Astra supplies dense motion references, velocity commands, or sparse body targets to frozen whole-body controllers. Locomotion includes both direct joint PD targets and references executed by ScaleTrack. Figure~\ref{fig:reasoning-boundary} shows these control paths. Each study specifies what Astra decides, how commands become motion, and what feedback informs the next action.

Within a trial, Astra uses feedback to generate or review the next action. Some studies retain notes or refine task guidance between trials with model weights fixed. Separate development studies involve researcher changes to prompts, interfaces, or whole-body controller settings. These procedures are described with their results and detailed in Appendix~\ref{app:interfaces}.

\subsection{What the Comparisons Establish}
Table~\ref{tab:scope} identifies the tasks, sample sizes, and comparison conditions. Where methods start from the same physical state, their outcomes can be compared on the same task instance. Each study specifies its observations, controllers, and inference budgets.

\begin{table*}[t]
\caption{\textbf{Evaluation scope and comparison conditions.} Direct uses Astra-generated commands with analytic control; Hybrid adds a learned task policy or whole-body controller. Sample sizes are reported separately for each study.}
\label{tab:scope}
\centering\small
\setlength{\tabcolsep}{5.00pt}
\begin{tabular}{>{\raggedright\arraybackslash}p{\dimexpr0.170\textwidth-2\tabcolsep\relax}>{\raggedright\arraybackslash}p{\dimexpr0.300\textwidth-2\tabcolsep\relax}>{\raggedright\arraybackslash}p{\dimexpr0.530\textwidth-2\tabcolsep\relax}}
\toprule
\rowcolor{tablehead}
\reporthead{Domain} & \reporthead{Tasks and sample sizes} & \reporthead{What is compared}\\
\midrule
\rowcolor{white}
Gripper manipulation & RoboDojo and RoboLab each cover 10 tasks with 5 results per task and method. & RoboDojo compares paired Direct and Hybrid runs with published policy scores. RoboLab compares Direct and Hybrid with three zero-shot policies. Control settings and evaluation procedures appear in Appendix~\ref{app:manipulation}. \\
\rowcolor{tablestripe}
Dexterous manipulation & 10 manipulation tasks with 5 cases each; 4 in-hand tasks with 5 initial states each. & Manipulation compares Direct, Hybrid, and a standalone policy on the same cases. In-hand control compares Astra and task-specific RL from identical physical states, with different observations and action rates. \\
\rowcolor{white}
Mobile manipulation & 15 RoboCasa tasks with 5 episodes per task and method. & Direct, Hybrid, and a locally evaluated standalone policy share initial states and 20-step action segments. The Astra conditions differ in policy access, context management, and feedback guards. \\
\rowcolor{tablestripe}
Navigation & 4 dataset subsets with 50 episodes per system in each. & Astra and released navigation policies use the same episode lists, scoring rules, and action budgets. Camera views and the conversion of outputs to navigation actions differ between systems. \\
\rowcolor{white}
Locomotion & 5 sequential Astra attempts on one course. Two reference interfaces, each tested on 6 courses in 2 physics backends: 12 rollouts per interface. & Astra and PASSAGE~\cite{ma2026passagescalingscenealignedmotion} share the initial state, goal, and frozen tracker. Separate analytic tests compare five-point and whole-body reference interfaces without Astra. \\
\rowcolor{tablestripe}
Humanoid loco-manipulation & HumanoidBench covers 30 tasks. SIMPLE covers 6 L2 tasks with 10 scenes each. & HumanoidBench compares Astra returns with published DreamerV3, TD-MPC2, and SAC results. SIMPLE uses the official codebase and protocol for comparison. \\
\bottomrule
\end{tabular}
\end{table*}

We report fixed-configuration evaluations and development trials separately. The former measure performance under specified conditions; the latter track changes in behavior as prompts, interfaces, or whole-body controller settings are refined. Transfer trials assess the resulting behavior on additional states.

Performance depends on the complete system, including the available commands, written guidance, and whole-body controller, even when Astra's weights remain fixed. For example, the grasp interface determines which actions are possible, while gait calibration changes how commands produce motion.

\subsection{Measuring Progress and Timing}
We distinguish completing a task from making partial progress or merely avoiding failure. The metrics must capture the intended change in physical state. We interpret rotation error together with angular speed, navigation success with path efficiency, and humanoid reward with observed progress and termination.

Controller frequency and reasoning speed are separate quantities. A fast controller can apply actions while Astra takes much longer to choose the next command. When physics pauses during inference, the robot-motion clock excludes decision latency. Resource summaries distinguish robot control steps, action segments, model requests, and cached input, uncached input, and output tokens. Astra can review many policy proposals while changing few actions. Section~\ref{sec:resources} summarizes inference demands.

%% file: sec/4_experiments.tex
\input{sec/4a_manipulation}
\input{sec/4c_1_dexhand_fused}
\input{sec/4g_mobile_manipulation}
\input{sec/4f_navigation}
\input{sec/4b_1_humanoid_locomotion}
\input{sec/4b_humanoid}

%% file: sec/4a_manipulation.tex
\section{Gripper Manipulation}
\label{sec:manipulation}

We compare two closed-loop control architectures for manipulation. \emph{Direct} uses GPT 6 Astra alone: given images, proprioception, the task instruction, and execution history, it generates bimanual end-effector (EEF) targets and gripper commands, executes one to five control steps, and then observes the new state. \emph{Hybrid} uses a learned policy, $\pifive$, to propose a 50-step joint-space action sequence. GPT 6 Astra reviews the proposal and either executes its first one to fifteen steps or replaces it with a one-to-five-step EEF correction; the two branches are mutually exclusive. Both architectures use the same task descriptions, success criteria, GPT 6 Astra reasoning setting, and EEF execution interface. 

\begin{figure*}[t]
  \centering
  \includegraphics[width=\textwidth]{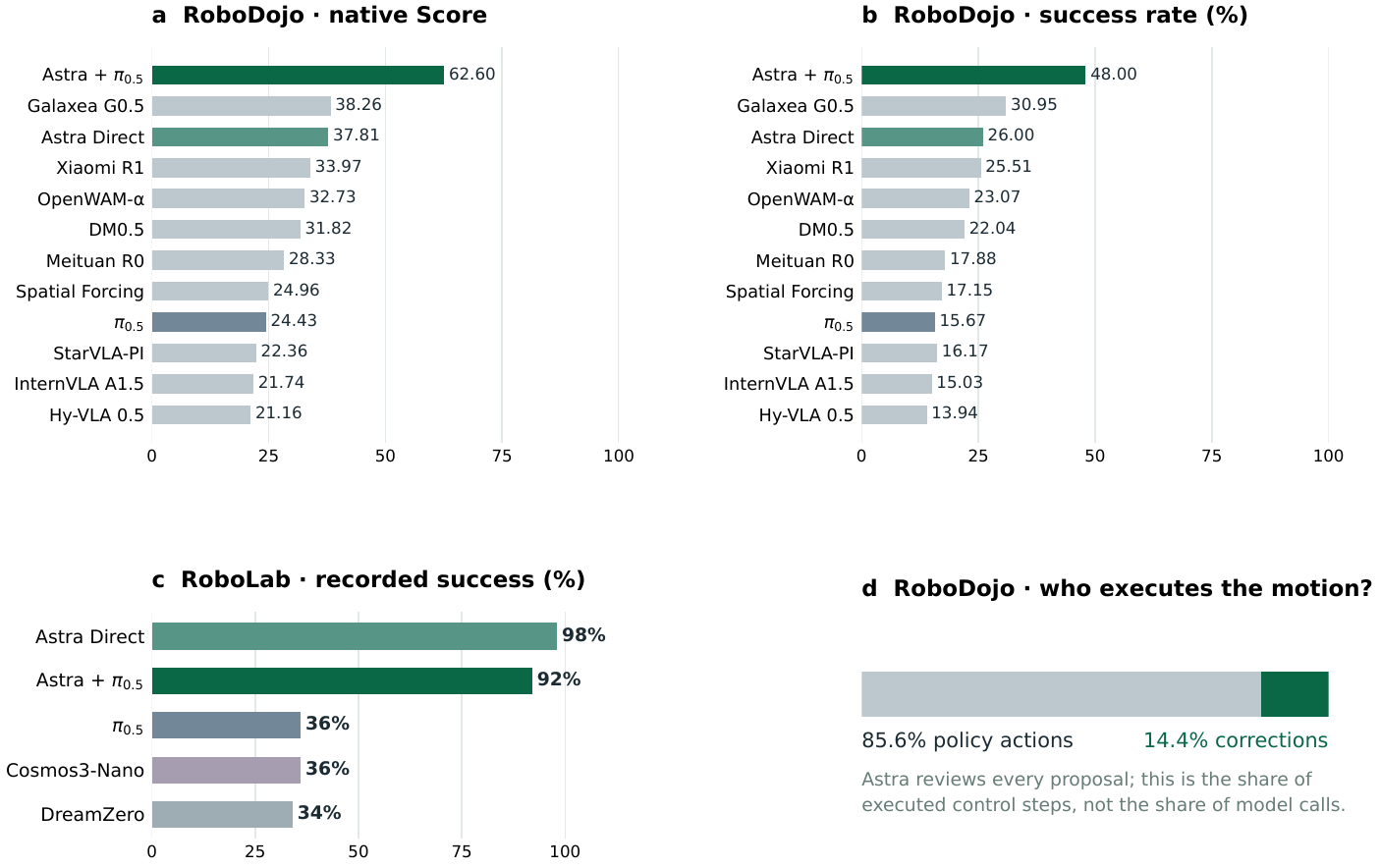}
  \caption{\textbf{Gripper manipulation results.} (a,b) RoboDojo Score and final success on ten tasks. Public references are reweighted to the same task and scene mixture, whereas Direct and Hybrid use paired starts. Direct has 48 valid Score entries; success is computed over all 50 task instances. (c) RoboLab success rates. (d) Of the Hybrid control steps, 85.6\% follow $\pifive$ and 14.4\% are generated or corrected by GPT 6 Astra.}
  \label{fig:manipulation}
\end{figure*}

\subsection{Control and evaluation setup}
\label{sec:manipulation-baselines}

The two architectures test complementary uses of GPT 6 Astra. Direct asks the model to construct the complete EEF action from the current state. Hybrid gives the model a candidate trajectory from $\pifive$ and asks it to preserve the trajectory or make a local correction. In Hybrid, GPT 6 Astra can therefore change the target, prepare the contact geometry, or continue after the simulator reports that the task is not yet complete, while $\pifive$ supplies most of the low-level motion. These architectures are evaluated separately on RoboDojo and RoboLab. 
Qualitative trajectories and the complete set of video cases are available in the accompanying \href{https://anonymous-report-421.github.io/public-website/?view=1}{web report}. 

\subsection{RoboDojo: paired bimanual evaluation}
\label{sec:robodojo}

\paragraph{Setting.}
RoboDojo~\cite{robodojo} is a bimanual manipulation benchmark covering semantic classification, sequence imitation, packing, construction, and deformable-object manipulation. We select ten tasks by stratifying the published $\pifive$ success rates: six tasks from the lowest interval, two from the next, and one from each of the two higher intervals. This selection emphasizes tasks with room for improvement while retaining diverse interaction requirements. Each task is evaluated five times for each architecture. Direct and Hybrid use the same task instances, scene configurations, and random seeds. Number arrangement, packing, and clothes folding use two standard and three randomized scenes; the other tasks use five standard layouts. We report RoboDojo's native final success rate and partial-completion Score. Hybrid uses the task-finetuned $\pifive$ weights released for RoboDojo.

\paragraph{Results.}
Hybrid succeeds on 24 of 50 instances (48\%), compared with 13 of 50 (26\%) for Direct. Its mean Score is 62.60, compared with 37.81 for Direct. The improvement is concentrated in tasks that require long-horizon coordination or contact-sensitive interaction: Hybrid scores 53 versus 0 on sequence imitation, 64 versus 12 on tower construction, 100 versus 40 on clothes folding, and 100 versus 36 on bottle disposal. Direct is higher on language-based classification (60 versus 38) and object classification (100 versus 71), showing that the Hybrid advantage is not uniform across tasks. In the 50 Hybrid trajectories, 36,576 of 42,750 executed control steps (85.6\%) follow $\pifive$, while 6,174 (14.4\%) are generated or corrected by GPT 6 Astra. The model therefore intervenes selectively rather than replacing the learned policy.

Reweighting the published RoboDojo results to the evaluated ten-task and scene mixture gives $\pifive$ a success rate of 15.67\% and a Score of 24.43. Table~\ref{tab:dojo-public} lists the published references alongside our paired Direct and Hybrid runs.

\begin{table*}[t]
\caption{\textbf{RoboDojo task-level Score.} Public references are reweighted to the evaluation task and scene mixture. Direct and Hybrid are our paired evaluations; the Direct Score uses 48 valid entries, while success uses all 50 task instances.}
\label{tab:dojo-public}
\centering\small
\setlength{\tabcolsep}{9.27pt}
\begin{tabular}{lrrrrrrr}
\toprule
\rowcolor{tablehead}
&  \multicolumn{5}{c}{\reporthead{Reweighted public references}}  &  \multicolumn{2}{c}{\reporthead{Our evaluation}} \\
\cmidrule(lr){2-6}\cmidrule(lr){7-8}
\rowcolor{tablehead}
\reporthead{Task} & \reporthead{DM0.5} & \reporthead{Galaxea} & \reporthead{Xiaomi R1} & \reporthead{OpenWAM} & \reporthead{\pifive} & \reporthead{Hybrid} & \reporthead{Direct}\\
\midrule
\input{tables/robodojo-public-score.tex}
\bottomrule
\end{tabular}
\end{table*}

\subsection{RoboLab: zero-shot evaluation}
\label{sec:robolab}

\paragraph{Setting.}
RoboLab~\cite{yang2026robolab} evaluates a single-arm Franka on ten tasks involving semantic pick-and-place, ordered block stacking, and mug reorientation. We report five trials per task and compare Direct, Hybrid, $\pifive$, Cosmos3-Nano-Policy~\cite{nvidia2026cosmos3policy}, and DreamZero~\cite{ye2026dreamzero}. This benchmark is evaluated with final task success only; RoboDojo's partial-completion Score is not used. RoboLab provides no training data for the test tasks. The policy baselines therefore use DROID-trained weights for zero-shot transfer, unlike RoboDojo, where Hybrid uses task-finetuned $\pifive$.

\paragraph{Results.}
Direct succeeds on 49/50 trials (98\%), with five successes on nine of the ten tasks. Hybrid succeeds on 46/50 (92\%), while $\pifive$, Cosmos3-Nano-Policy, and DreamZero succeed on 18/50 (36\%), 18/50 (36\%), and 17/50 (34\%), respectively. Hybrid therefore improves over $\pifive$ by 56 percentage points, but is three successful trajectories below Direct on this task subset.

\subsection{Discussion: why the benchmark rankings differ}
\label{sec:manipulation-discussion}

RoboDojo and RoboLab favor different capabilities. RoboDojo contains bimanual, long-horizon, deformable, and contact-sensitive tasks. On these tasks, the learned policy supplies interaction patterns and coordinated motion, while GPT 6 Astra can correct the target, local geometry, or task-progress interpretation. This division of labor is consistent with Hybrid outperforming Direct on RoboDojo.

The RoboLab tasks are predominantly single-arm semantic pick-and-place problems, with additional ordered stacking and mug reorientation. Direct can already plan these operations from the current observation and reaches 98\% success. In this setting, a candidate action from a zero-shot policy is not necessarily a useful prior: $\pifive$ alone succeeds on only 36\% of trials, so Hybrid may spend decisions correcting an action that is poorly adapted to the task. This suggests that the value of policy assistance depends on how well the learned motion prior fits the task.

%% file: tables/robodojo-public-score.tex
\rowcolor{white}
Organize the table & \cellcolor{galbot!14}44.00 & \cellcolor{galbot!15}46.33 & \cellcolor{galbot!17}57.67 & \cellcolor{galbot!19}\textbf{62.50} & \cellcolor{galbot!9}23.33 & \cellcolor{galbot!18}60.00 & \cellcolor{galbot!10}30.00 \\
\rowcolor{tablestripe}
Classify by language & \cellcolor{galbot!3}0.47 & \cellcolor{galbot!3}1.07 & \cellcolor{galbot!4}2.00 & \cellcolor{galbot!3}1.33 & \cellcolor{galbot!3}0.60 & \cellcolor{galbot!12}38.00 & \cellcolor{galbot!18}\textbf{60.00} \\
\rowcolor{white}
Imitate a sorting sequence & \cellcolor{galbot!3}1.80 & \cellcolor{galbot!3}1.67 & \cellcolor{galbot!4}2.50 & \cellcolor{galbot!4}2.90 & \cellcolor{galbot!3}1.60 & \cellcolor{galbot!16}\textbf{53.00} & \cellcolor{galbot!3}0.00 \\
\rowcolor{tablestripe}
Arrange the largest number & \cellcolor{galbot!5}7.85 & \cellcolor{galbot!4}4.11 & \cellcolor{galbot!5}8.56 & \cellcolor{galbot!4}4.36 & \cellcolor{galbot!4}2.29 & \cellcolor{galbot!16}50.00 & \cellcolor{galbot!17}\textbf{57.00} \\
\rowcolor{white}
Pack objects into a box & \cellcolor{galbot!7}14.72 & \cellcolor{galbot!7}17.12 & \cellcolor{galbot!8}18.69 & \cellcolor{galbot!8}20.83 & \cellcolor{galbot!8}18.36 & \cellcolor{galbot!16}\textbf{50.00} & \cellcolor{galbot!16}\textbf{50.00} \\
\rowcolor{tablestripe}
Classify objects & \cellcolor{galbot!10}26.83 & \cellcolor{galbot!6}10.33 & \cellcolor{galbot!7}17.20 & \cellcolor{galbot!4}5.53 & \cellcolor{galbot!9}24.67 & \cellcolor{galbot!21}71.00 & \cellcolor{galbot!28}\textbf{100.00} \\
\rowcolor{white}
Build a tower & \cellcolor{galbot!17}55.20 & \cellcolor{galbot!24}\textbf{82.93} & \cellcolor{galbot!16}52.60 & \cellcolor{galbot!16}52.53 & \cellcolor{galbot!12}37.73 & \cellcolor{galbot!19}64.00 & \cellcolor{galbot!6}12.00 \\
\rowcolor{tablestripe}
Make a Kong in Mahjong & \cellcolor{galbot!17}56.67 & \cellcolor{galbot!26}\textbf{90.00} & \cellcolor{galbot!13}41.33 & \cellcolor{galbot!11}32.00 & \cellcolor{galbot!10}26.67 & \cellcolor{galbot!13}40.00 & \cellcolor{galbot!3}0.00 \\
\rowcolor{white}
Fold clothes & \cellcolor{galbot!10}28.96 & \cellcolor{galbot!11}32.75 & \cellcolor{galbot!13}41.49 & \cellcolor{galbot!16}51.31 & \cellcolor{galbot!10}29.12 & \cellcolor{galbot!28}\textbf{100.00} & \cellcolor{galbot!13}40.00 \\
\rowcolor{tablestripe}
Put bottles in a bin & \cellcolor{galbot!23}81.70 & \cellcolor{galbot!27}96.30 & \cellcolor{galbot!27}97.70 & \cellcolor{galbot!27}94.03 & \cellcolor{galbot!23}79.93 & \cellcolor{galbot!28}\textbf{100.00} & \cellcolor{galbot!12}36.00 \\
\midrule
\rowcolor{tablesummary}
Aggregate & 31.82 & 38.26 & 33.97 & 32.73 & 24.43 & 62.60 & 37.81 \\

%% file: sec/4c_1_dexhand_fused.tex
\section{Dexterous Manipulation}
\label{sec:dexterous-manipulation}
\label{sec:dexterous-manipulation-fused}

Dexterous manipulation requires both task-level planning, such as selecting a suitable grasp and deciding how to use it, and precise coordination of finger contacts during execution. To assess these capabilities, we evaluate Astra on end-to-end manipulation tasks, with optional policy assistance, and on in-hand control from an established grasp. The former tests acquiring and using a grasp; the latter tests changing finger contacts while maintaining object support.

\subsection{Policy Assistance Across Ten Tasks}
The first benchmark covers grasping, retrieval, placement, insertion, stacking, and rearrangement with simulated Sharpa hands~\cite{sharpa_wave}. One multitask \pifive policy is finetuned on 100 demonstrations per task (1,000 in total). For evaluation, Direct, standalone \pifive, and Hybrid are each tested on the same five unseen cases per task, with simulation horizons of 20--60\,s. Direct outputs wrist and finger targets; Hybrid reviews or corrects policy actions, executing prefixes of 1--16 steps or corrections of 1--5 steps at approximately 30\,Hz. When the review or token budget is exhausted, the system falls back to \pifive-only control.

We evaluate performance using task completion scores on a 0--100 scale. The scores give partial credit for progress rather than measuring binary success, and we average them equally across tasks. Full credit requires task-specific verification of a maintained hold or a stable release. Appendix~\ref{app:dexterous} provides details on observations, training, actions, and scoring.

As shown in Table~\ref{tab:dexterous-mock-m1}, Hybrid achieves a mean Score of 61.6, compared with 44.2 for \pifive and 16.6 for Direct. Hybrid achieves a higher mean score than Direct on all ten tasks. Compared with \pifive, it scores higher on eight tasks and ties on the remaining two. Gains over the policy are largest for mahjong storage at 40 points, upright egg placement at 32, and bottle/can sorting at 28. On bread insertion and nesting-doll ordering, Hybrid improves only slightly over \pifive, with scores remaining at 28 and 24, respectively.

\input{tables/dexterous_s1}

\begin{figure*}[t]
  \centering
  \includegraphics[width=\textwidth]{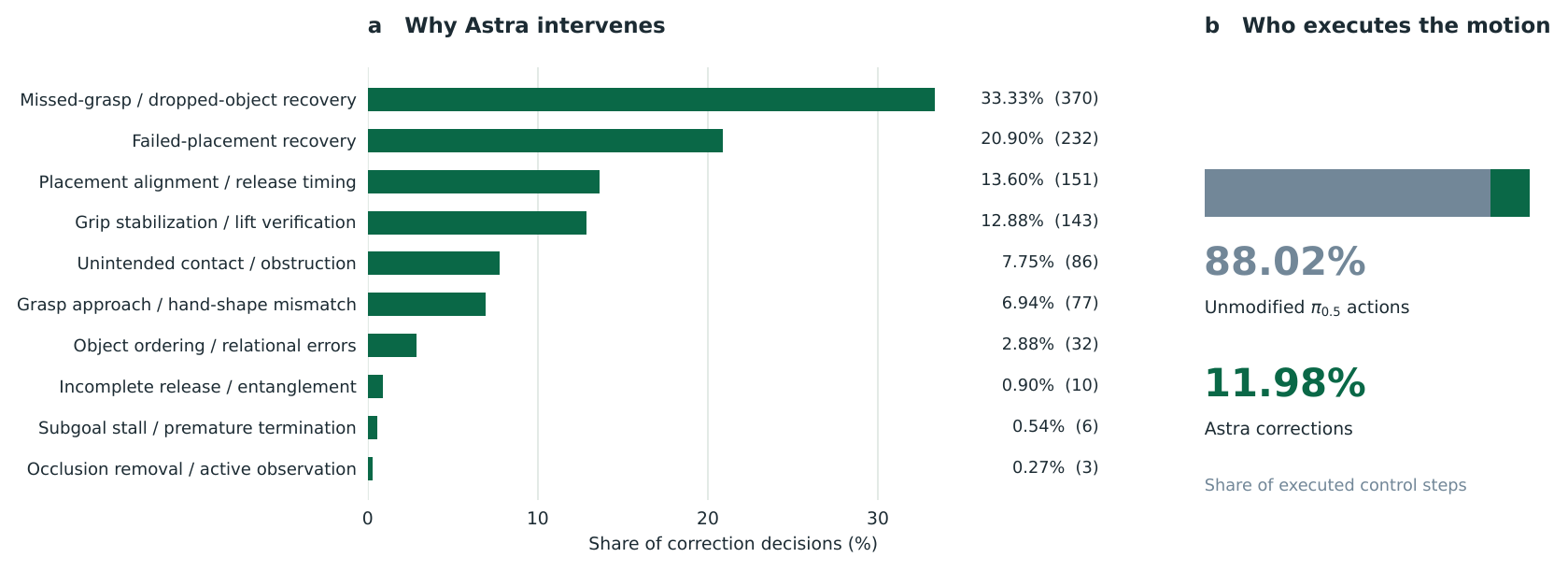}
  \caption{\textbf{Astra corrections in dexterous manipulation.} (a) Primary reasons for Astra correction decisions across 50 Hybrid cases. (b) Executed steps comprise 88.02\% unmodified $\pi_{0.5}$ actions and 11.98\% Astra corrections.}
  \label{fig:dexterous-interventions-fused}
\end{figure*}

Figure~\ref{fig:dexterous-interventions-fused}(a) categorizes Astra's interventions by failure type. The most common address missed grasps or dropped objects and failed placements, followed by placement alignment and grip stabilization. In individual trajectories, Astra redirects wrists toward displaced objects, clears occlusions, refines placements, and resumes unfinished subgoals. These interventions account for only 11.98\% of executed Hybrid steps (Figure~\ref{fig:dexterous-interventions-fused}(b)). Together with the task scores, this suggests that targeted changes to a small fraction of steps can yield substantial performance gains.

Even with these gains, grasp acquisition and completion checking remain the primary sources of failure. Direct can fail to establish a grasp when simultaneous finger closure pushes the object away or a weak two-finger lift leaves it poorly supported. Hybrid faces a similar difficulty when an object drops into an unfamiliar configuration: if the policy can no longer provide useful actions, Astra must generate a new grasp pose but may still fail to secure the object. These grasp failures suggest that Astra still relies heavily on a capable policy to generate viable action candidates for dexterous manipulation. In addition, Astra can misjudge task completion, as in an egg-placement case where it declares success while the egg remains on its side.

\subsection{Feedback-Guided Bimanual Manipulation in DexJoCo}
\label{sec:dexjoco-hanoi}
The Hanoi disk-stacking task in DexJoCo~\cite{wang2026dexjoco} tests two ordered transfers with dual Franka Panda arms and Allegro hands: the right hand moves the medium disk to the destination peg, then the left hand places the small disk above it. Astra reviews a fixed \pifive policy trained for this task, accepting action prefixes or correcting wrist and finger targets using visual feedback. Each trial allows up to 1,500 control steps at 50\,Hz, with at most 30 steps per action segment. Ten trials run in five successive pairs, starting from prior task experience. Written guidance is shared within each pair and refined between pairs, with model weights fixed. Image-history limits and compact references to older proposals are introduced during the run.

\begin{figure}[tb]
  \centering
  \includegraphics[width=\columnwidth]{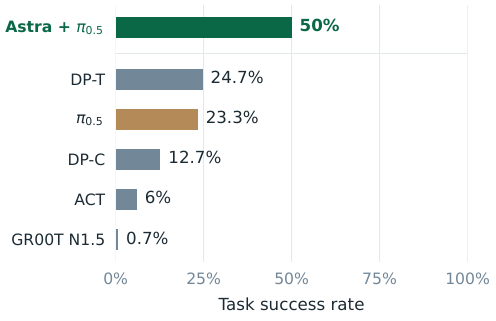}
  \caption{\textbf{Success on bimanual disk stacking.} Published baselines average three sets of 50 \texttt{rand-obj} episodes~\cite{wang2026dexjoco}; Astra + \pifive succeeds in 5/10 experience-guided trials.}
  \label{fig:dexjoco-hanoi-performance}
\end{figure}

\begin{figure*}[t]
  \centering
  \includegraphics[width=\textwidth,height=.36\textheight,keepaspectratio]{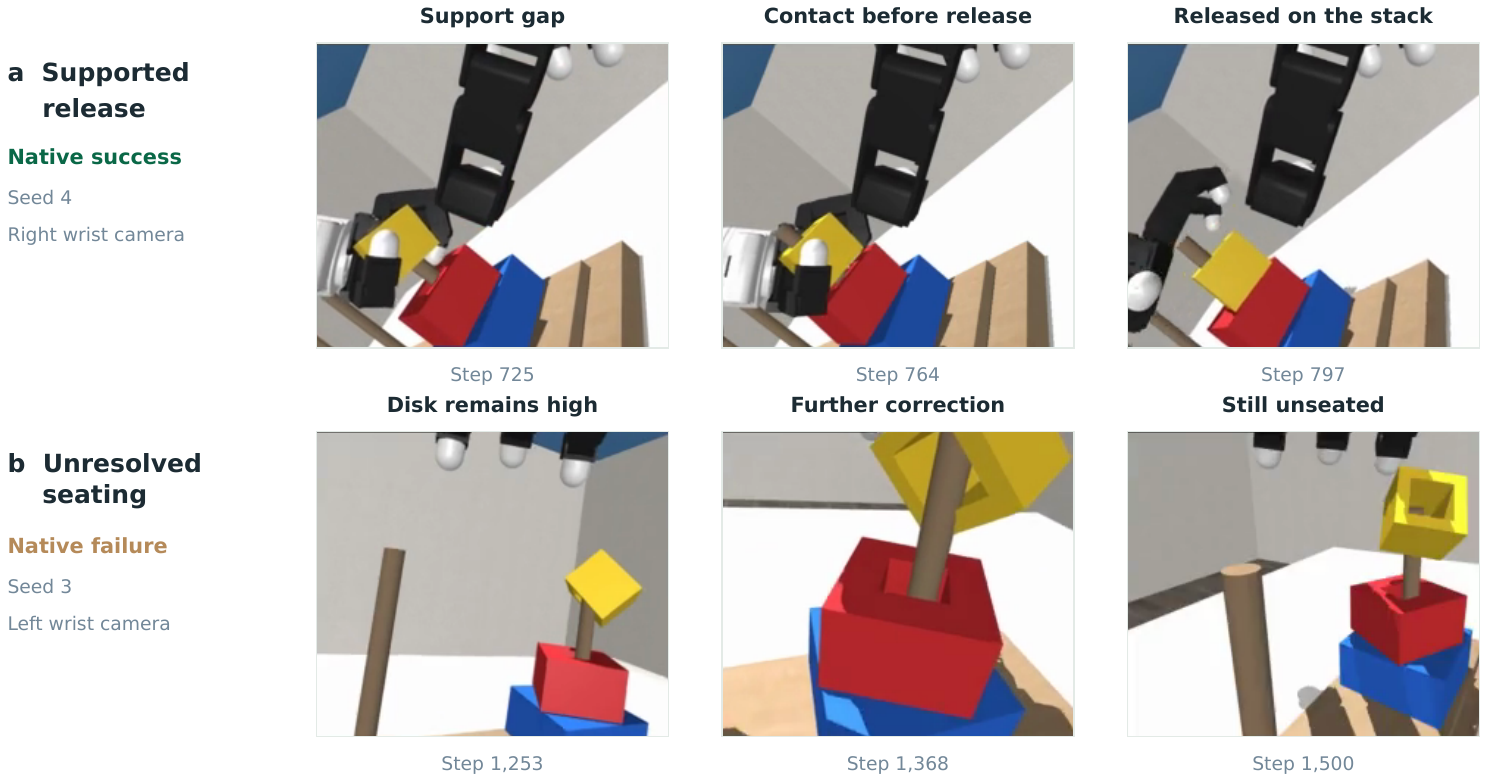}
  \caption{\textbf{Bimanual disk stacking: successful placement and failed seating.} Cropped wrist-camera observations show (a) Astra corrections followed by policy release and (b) persistent placement failure despite further corrections. Labels give control steps; outcomes follow the native success criterion.}
  \label{fig:dexjoco-hanoi}
\end{figure*}

The system completes 5/10 trials under the native success criterion. Figure~\ref{fig:dexjoco-hanoi-performance} places this result alongside published policy success rates on the same task. Of 11,320 executed steps, 73.6\% use unmodified policy actions and 26.4\% use Astra corrections. These results summarize the full sequence of experience-guided trials. The trajectories in Figure~\ref{fig:dexjoco-hanoi} show useful local interventions alongside persistent failures in grasp retention and final placement.

A successful placement requires more than reaching the correct peg. In one successful trial, Astra delays a proposed release because the small disk remains above its support. Short downward corrections bring the disk into contact with the stack, after which Astra returns control to \pifive for release and withdrawal. The resulting hand clearance allows a separate check that the disk remains supported. This sequence illustrates Astra's role in deciding when a physical precondition has been met, while the policy supplies the subsequent motion.

The same strategy is not consistently effective. In a failed trial, repeated seating corrections leave the small disk tilted high on the destination peg at the step limit. Other failures involve losing a grasp during transport or failing to lift the small disk from its source. Astra's own corrections can also hinder progress: in another successful trial, two wrist adjustments move the held disk away from the destination, whereas returning control to the policy restores transport. Effective cooperation therefore requires both timely intervention and recognition that the policy may offer the better recovery. The trials and five reviews consume 63.60 million recorded tokens, including cached input; only 0.136 million are used for review. Most token use arises during online control, even though the policy supplies most physical actions.

\subsection{Stable Grasping, Limited Rotation}
Once a grasp is established, success requires moving the object while maintaining support. We compare direct Astra joint commands with four frozen, task-specific RL policies to examine this demand. Sharpa tasks require continuous cylinder or cuboid rotation about world-frame $+Z$; Allegro tasks require cylinder translation, with or without an additional $-40^\circ$ long-axis rotation. Rotation tasks draw on ConTrack~\cite{liang2026contrack} and \texttt{mjlab\_hand}~\cite{qiao2026mjlabhand}; translation follows tactile in-hand manipulation~\cite{yin2025learning}.

Each task uses five paired saved hand/object states and targets. Astra receives images and state and commands 22 Sharpa or 16 Allegro joints for one to five 20-Hz steps. RL uses its training observations and acts every step. Thus physical starts are paired, but observations and decision rates differ. Four cuboid starts overlap the RL initialization bank. Training and per-case records appear in Appendix~\ref{sec:supp-dexterous}.

An established grasp does not remove Astra's difficulty with sustained object motion. RL tracks the rotation target within tolerance for 76.90\% of cylinder steps and 63.50\% of cuboid steps, versus 0.51\% and 4.40\% for Astra (Table~\ref{tab:dexterous-rotation-fused}). Astra reaches the horizon in three cylinder runs, drops the object once, and exhausts its decision budget once. The cuboid results isolate the distinction between retention and progress more clearly: both controllers retain the object through all five horizons, yet Astra rotates it too slowly. The gap therefore persists even when the grasp remains secure.

\begin{figure*}[t]
  \centering
  \includegraphics[width=\textwidth]{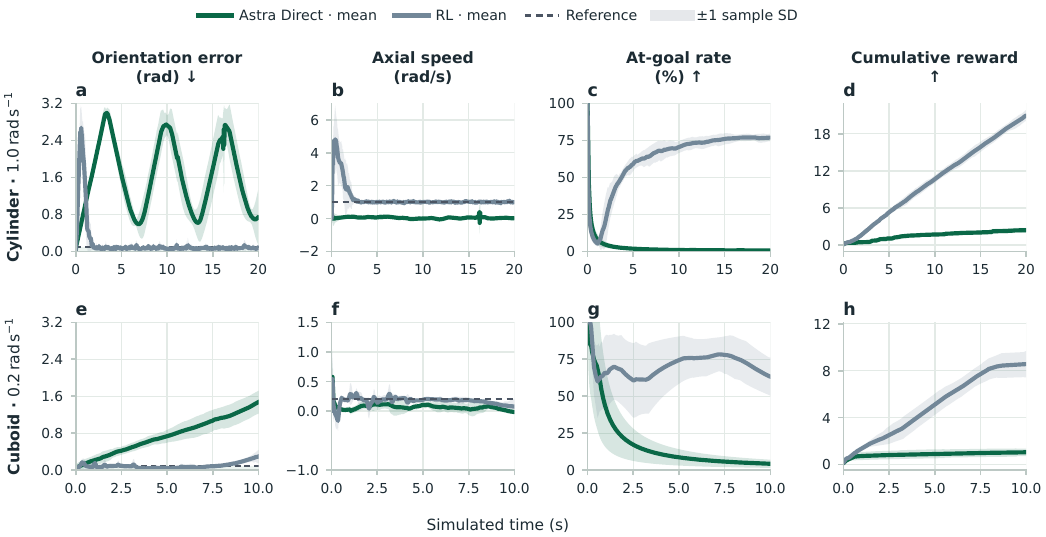}
  \caption{\textbf{Continuous rotation over five paired initial states.} Rows show cylinder results over 20\,s and cuboid results over 10\,s. Columns report orientation error, axial speed over the preceding second, cumulative at-goal rate, and cumulative reward. Curves and shading show means and one sample SD among active runs. Dashed lines mark target speeds and the $0.1$-rad goal threshold. Periodic orientation error requires interpretation alongside axial speed.}
  \label{fig:dexterous-rotation-trajectories-fused}
\end{figure*}

The motion traces show how this gap develops. Astra's cylinder speed stays near zero, while the cuboid turns gradually but falls behind the target (Figure~\ref{fig:dexterous-rotation-trajectories-fused}). Representative cuboid sequences show little contact reconfiguration under Astra, whereas RL continues reorientation (Appendix~\ref{sec:supp-dexterous}). This behavior suggests that preserving support is easier for Astra than reorganizing finger contacts to sustain motion. Periodic decreases in orientation error do not contradict this interpretation: the rotating target returns to an equivalent orientation after each revolution, so error must be read alongside axial speed.

\begin{table*}[t]
\caption{\textbf{Paired in-hand rotation results.} Full, Budget, and Drop denote runs reaching the horizon, exhausting Astra's decision limit, or losing the object, respectively. At-goal is the fraction of steps with quaternion geodesic error below $0.1\,\mathrm{rad}$. Error and Speed MAE are means over executed steps; terminated trajectories are not padded.}
\label{tab:dexterous-rotation-fused}

  \centering
  
  \small
  \setlength{\tabcolsep}{8.46pt}
  \begin{tabular}{llrrrrrrr}
    \toprule
    \rowcolor{tablehead}
\reporthead{Task} & \reporthead{Controller} & \reporthead{Steps} & \reporthead{Full} & \reporthead{Budget} & \reporthead{Drop} & \reporthead{At-goal $\uparrow$} & \reporthead{Error (rad) $\downarrow$} & \reporthead{Speed MAE $\downarrow$}\\
    \midrule
    \rowcolor{white}
Cylinder, 20 s & Astra Direct & 1,749 & 3/5 & 1/5 & 1/5 & 0.51\% & 1.626 & 0.960 \\
\rowcolor{white}
& RL           & 2,000 & 5/5 & -- & 0/5 & \textbf{76.90\%} & \textbf{0.173} & \textbf{0.276} \\
    \addlinespace
    \rowcolor{tablestripe}
Cuboid, 10 s   & Astra Direct & 1,000 & 5/5 & 0/5 & 0/5 & 4.40\% & 0.742 & 0.143 \\
\rowcolor{tablestripe}
& RL           & 1,000 & 5/5 & -- & 0/5 & \textbf{63.50\%} & \textbf{0.096} & \textbf{0.063} \\
    \bottomrule
  \end{tabular}
\end{table*}

\begin{table*}[t]
\caption{\textbf{Paired in-hand translation and reorientation results.} Errors are measured at the stated terminal endpoint. Joint success requires both position error below $22.4\,\mathrm{mm}$ and long-axis rotation error below $10^\circ$ at the same endpoint.}
\label{tab:dexterous-fixed-goal-fused}

  \centering
  
  \small
  \setlength{\tabcolsep}{10.11pt}
  \begin{tabular}{lllrrr}
    \toprule
    \rowcolor{tablehead}
\reporthead{Task} & \reporthead{Controller} & \reporthead{Endpoint} & \reporthead{Position (mm) $\downarrow$} & \reporthead{Rotation (deg) $\downarrow$} & \reporthead{Success}\\
    \midrule
    \rowcolor{white}
Translation & Astra Direct & 15 s & 59.05 & -- & 1/5 \\
\rowcolor{white}
& RL           & 15 s & \textbf{17.29} & -- & \textbf{4/5} \\
    \addlinespace
    \rowcolor{tablestripe}
Translation + rotation & Astra Direct & 120 decisions & 47.35 & 32.72 & 0/5 \\
\rowcolor{tablestripe}
& RL & matched steps & \textbf{16.22} & \textbf{8.31} & \textbf{4/5} \\
\rowcolor{tablestripe}
& RL & 15 s & \textbf{16.08} & \textbf{3.49} & \textbf{5/5} \\
    \bottomrule
  \end{tabular}
\end{table*}

Translation reveals some useful direct control, but limited consistency across starts. Astra succeeds in 1/5 cases and RL in 4/5, with mean terminal errors of 59.05 and 17.29\,mm (Table~\ref{tab:dexterous-fixed-goal-fused}). Astra achieves lower position error in one paired case, showing that it can produce an effective displacement. Yet it retains the object in every translation run and still misses most targets. As in rotation, failure to reach the goal cannot be explained by object drops alone.

\begin{figure*}[t]
  \centering
  \includegraphics[width=\textwidth]{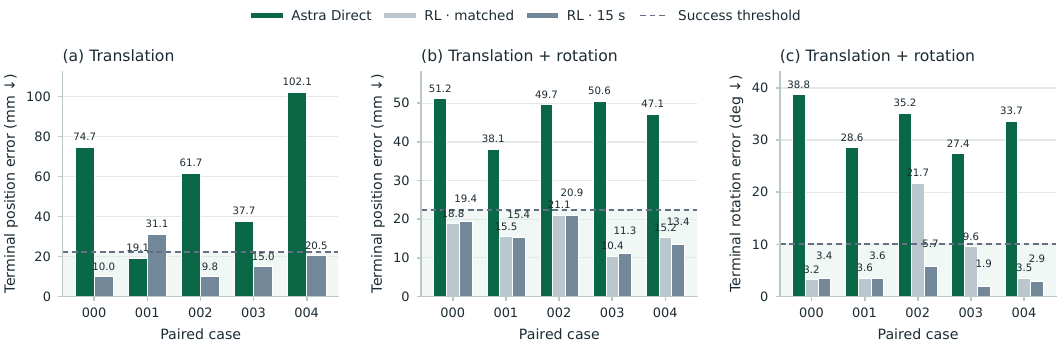}
  \caption{\textbf{Per-case terminal errors for Allegro tasks.} (a) Translation at 15\,s. (b,c) Translation and rotation: Astra at its 120-decision endpoint, RL at matched control steps, and RL at 15\,s. Shading and dashed lines mark the $22.4$-mm position and $10^\circ$ rotation success thresholds.}
  \label{fig:supp-dexterous-translation-results}
\end{figure*}

For combined translation and rotation, Astra succeeds in 0/5 cases, compared with 4/5 for RL at matched endpoints and 5/5 after 15\,s. Success requires position error below 22.4\,mm and rotation error below $10^\circ$ at the same endpoint. Astra's 120-decision budget ends after 6.00--7.65\,s, but RL already performs better over those same durations (Figure~\ref{fig:supp-dexterous-translation-results}). A longer action horizon alone therefore does not explain the gap. The results favor task-specific learned control for reliable object motion.

\subsection{Contact Changes and Object Support}
Together, the two studies distinguish useful task correction from reliable contact coordination. Astra can redirect a learned policy when its actions no longer match the scene or instruction, but direct control remains unreliable both when acquiring a grasp and when moving within one. A stable grasp is only an intermediate condition: continued manipulation requires some fingers to release and reposition while others preserve support. The central limitation is sustaining this coordination as the object moves. Hybrid benefits from retaining learned motion generation while allowing Astra to guide recovery, yet recovery and completion still require verification.

%% file: tables/dexterous_s1.tex
\begin{table*}[t]
\caption{\textbf{Dexterous manipulation scores on a custom simulation benchmark.} \label{tab:dexterous-mock-m1}
Scores measure subgoal completion on a 0--100 scale.
The evaluation protocol and scoring details are provided in
Appendix~\ref{app:dexterous-evaluation}.}

\centering
\small
\setlength{\tabcolsep}{20.88pt}

\begin{tabular}{lrrr}
\toprule
\rowcolor{tablehead}
\reporthead{Task} & \reporthead{$\pi_{0.5}$} & \reporthead{Direct} & \reporthead{Hybrid}\\
\midrule
\rowcolor{white}
Pot lift and hold & 40.0 & 44.0 & \textbf{62.0} \\
\rowcolor{tablestripe}
Headphones in box & \textbf{42.0} & 22.0 & \textbf{42.0} \\
\rowcolor{white}
Toy retrieval & \textbf{100.0} & 22.0 & \textbf{100.0} \\
\rowcolor{tablestripe}
Mug hanging & 62.0 & 4.0 & \textbf{82.0} \\
\rowcolor{white}
Mahjong tile storage & 48.0 & 18.0 & \textbf{88.0} \\
\rowcolor{tablestripe}
Bottles/cans sorting & 72.0 & 14.0 & \textbf{100.0} \\
\rowcolor{white}
Upright egg placement
 & 6.0 & 12.0 & \textbf{38.0} \\
\rowcolor{tablestripe}
Two-bowl stacking & 46.0 & 20.0 & \textbf{52.0} \\
\rowcolor{white}
Bread-slot insertion & 10.0 & 8.0 & \textbf{28.0} \\
\rowcolor{tablestripe}
Nesting-doll ordering & 16.0 & 2.0 & \textbf{24.0} \\

\midrule
\rowcolor{tablesummary}
\textbf{Overall mean} & 44.2 & 16.6 & \textbf{61.6} \\
\bottomrule
\end{tabular}
\end{table*}

%% file: sec/4g_mobile_manipulation.tex
\section{Mobile Manipulation}
\label{sec:mobile_manipulation}

Mobile manipulation couples base placement, arm reachability, and object interaction. Using RoboCasa365~\cite{nasiriany2026robocasa365}, we evaluate Astra Direct, Astra Hybrid with optional \pifive, and standalone \pifive.

\begin{figure*}[t]
\centering
\includegraphics[width=\textwidth]{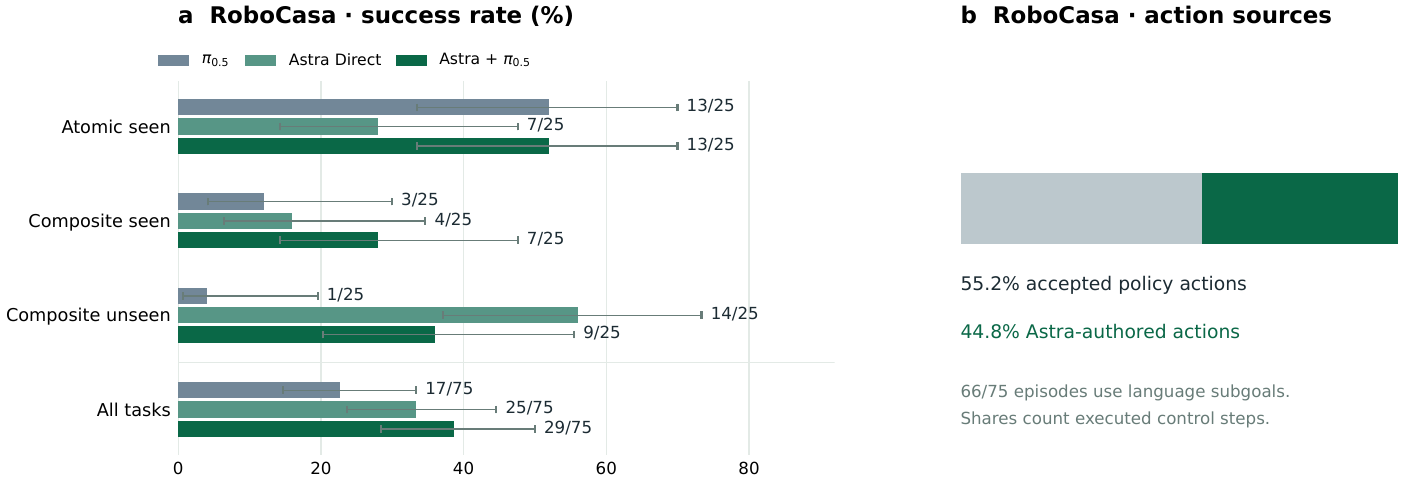}
\caption{\textbf{Cooperation changes the RoboCasa capability profile.} (a) Success on matched physical starts. Error bars show descriptive episode-level 95\% Wilson intervals~\cite{wilson1927}. (b) Step shares from accepted proposals and Astra actions.}
\label{fig:robocasa-results}
\end{figure*}

\subsection{Control Conditions and Shared Starts}
The fixed subset contains five atomic seen, five composite seen, and five composite unseen tasks, each with five seeds: 75 episodes per condition. Selection uses a seeded ordering without model outcomes. Conditions share initial-state fingerprints, the PandaOmron robot, controller, task horizon, and native success check. Seen/unseen refers to the policy's task-training split. All layouts use \texttt{split=pretrain}.

Astra receives three RGB views and end-effector, base, and gripper state, without object poses, contact truth, reference trajectories, or intermediate rewards. Direct supplies bounded arm, gripper, base, and torso commands through an operational-space controller. Segments normally execute 20 steps at 20\,Hz; horizons range from 450 to 4,350 steps.

Hybrid adds a frozen multitask \pifive checkpoint without separate finetuning for these tasks. Astra can accept a 20-step prefix, construct actions, or rewrite the policy instruction as a subgoal. The standalone baseline runs the same checkpoint locally on the same initial states, using the task instruction and the same 20-step cadence. Direct and Hybrid also differ in context management and feedback guards, including context compaction and required observations between action calls, as detailed in Appendix~\ref{app:robocasa-controls}.

\begin{figure*}[t]
\centering
\includegraphics[width=.82\textwidth]{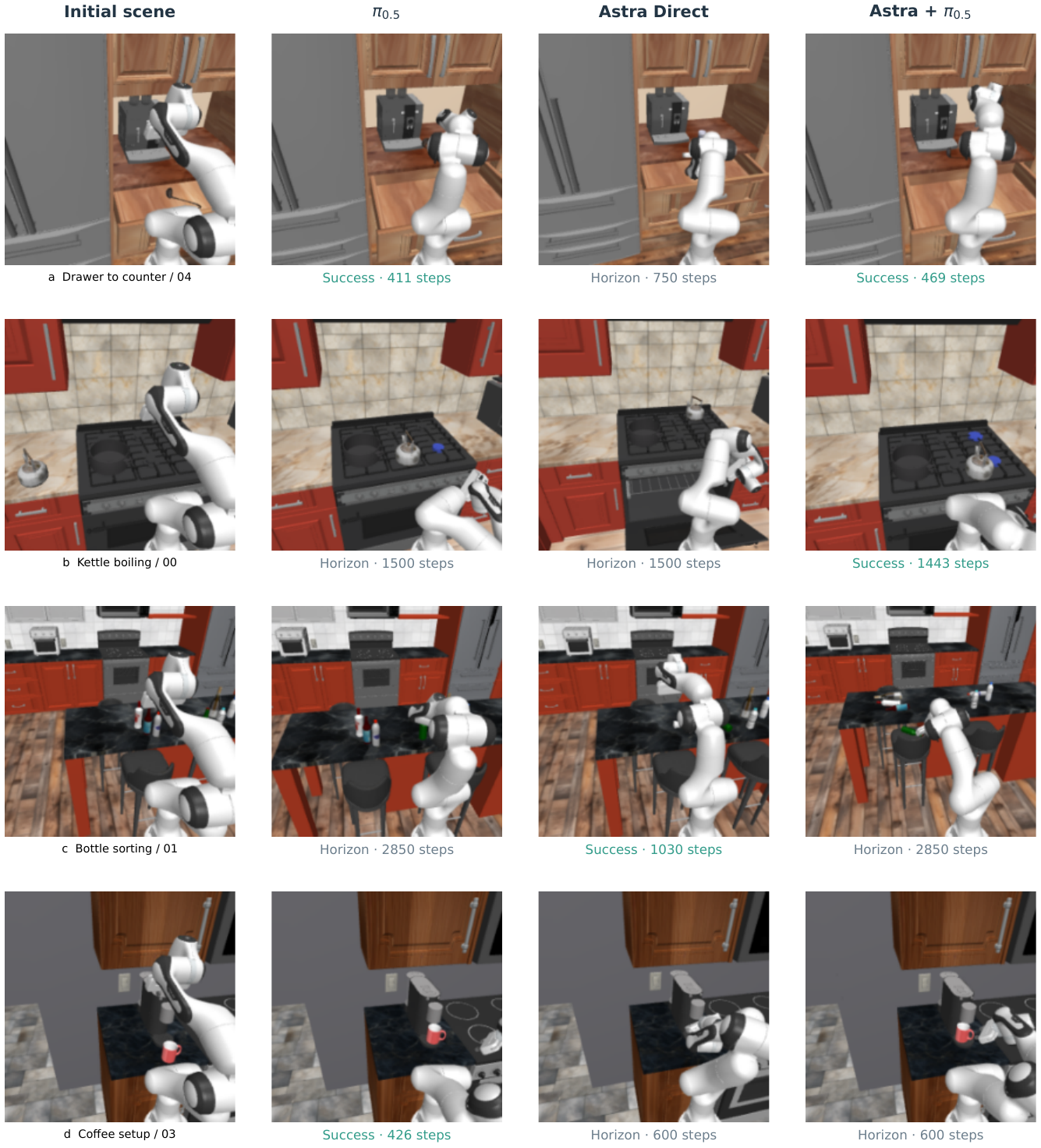}
\caption{\textbf{RoboCasa outcomes from matched physical starts.} Rows identify four task/seed pairs. Initial images come from the independent-policy run; subsequent columns show each condition\textquotesingle s terminal view. Labels report native outcomes and executed steps. All images are saved observations; the coffee case succeeds only with the standalone policy.}
\label{fig:robocasa-cases}
\end{figure*}

\begin{table*}[t]
\caption{\textbf{RoboCasa completion on matched initial states.} Each group contains five tasks with five episodes each; entries give successes/episodes and percentage. All conditions use the same 20-step feedback cadence and task horizons. Task groups describe the policy's training coverage.}
\label{tab:robocasa-results}

\centering\small
\setlength{\tabcolsep}{16.03pt}
\begin{tabular}{lrrr}
\toprule
\rowcolor{tablehead}
\reporthead{Task group} & \reporthead{\pifive alone} & \reporthead{Astra Direct} & \reporthead{Astra Hybrid}\\
\midrule
\input{tables/robocasa-results.tex}
\bottomrule
\end{tabular}

\end{table*}

\subsection{Uneven Gains from Policy Assistance}
Astra Hybrid completes 29/75 episodes, compared with 25/75 for Direct and 17/75 for standalone \pifive in Table~\ref{tab:robocasa-results}. Hybrid exceeds Direct on atomic seen tasks, 52\% versus 28\%, and composite seen tasks, 28\% versus 16\%. Direct is stronger on unseen compositions, 56\% versus 36\%. Hybrid and standalone \pifive both reach 13/25 on atomic tasks, but succeed on different instances.

Hybrid exceeds Direct on tasks seen during policy training but trails it on unseen compositions. These averages also conceal complementary successes. In Table~\ref{tab:robocasa-paired}, Hybrid alone succeeds on 13 paired starts and Direct alone on nine, with 16 shared successes. An exploratory exact McNemar test~\cite{fagerland2013mcnemar} gives $p=0.5235$.

Astra organizes grasping, transport, placement, and device operation through language decomposition and numerical control. In Hybrid, 66/75 episodes use rewritten instructions, which condition 72.6\% of proposals. Figure~\ref{fig:robocasa-results} attributes 55.2\% of executed steps to accepted policy actions and 44.8\% to Astra. Rewriting changes what the policy is asked to achieve; numerical correction changes the robot commands, while context and feedback guards structure proposal review. The reported results evaluate these components jointly. Delegation still requires proposal review; Section~\ref{sec:resources} examines its inference demands.

\subsection{Recovery and Lost Policy Successes}
Figure~\ref{fig:robocasa-cases} illustrates these differences on matched starts. Only Hybrid succeeds on \texttt{KettleBoiling/00}, and only Direct succeeds on \texttt{RecycleBottlesByType/01}. On \texttt{PickPlaceDrawerToCounter/04}, both Hybrid and standalone \pifive succeed.

Conversely, only standalone \pifive succeeds on \texttt{CoffeeSetupMug/03}. Across all five seeds, both Astra conditions fail CoffeeSetupMug and WashLettuce, while the policy completes three and two cases, respectively. Policy access need not preserve success.

Visual feedback lets Astra detect some action errors, but recognizing a problem does not ensure a successful correction. Failure trajectories expose weaknesses in precise grasping and sustained contact, base--arm coordination while carrying objects, and tracking preconditions and object states across task stages. \pifive helps on some tasks but does not ensure reliable state verification or physical control.

%% file: tables/robocasa-results.tex
\rowcolor{white}
Atomic seen & 13/25 (52.0\%) & 7/25 (28.0\%) & 13/25 (52.0\%) \\
\rowcolor{tablestripe}
Composite seen & 3/25 (12.0\%) & 4/25 (16.0\%) & 7/25 (28.0\%) \\
\rowcolor{white}
Composite unseen & 1/25 (4.0\%) & 14/25 (56.0\%) & 9/25 (36.0\%) \\
\rowcolor{tablesummary}
All tasks & 17/75 (22.7\%) & 25/75 (33.3\%) & 29/75 (38.7\%) \\

%% file: sec/4f_navigation.tex
\section{Navigation}
\label{sec:navigation}

We evaluate Astra on instruction following in continuous environments (VLN-CE) and category-directed object search (ObjectNav)~\cite{krantz2020vlnce,batra2020objectnav}, measuring both goal reaching and path efficiency. Astra chooses where to move and when to stop using RGB observations without pose or map input; the action interface applies discrete forward steps and turns.

\subsection{RGB Navigation on Four Fixed Subsets}
We select 50 validation-unseen episodes each from R2R and English-guide RxR, and 50 validation episodes each from MP3D ObjectNav v1 and HM3D ObjectNav v2~\cite{anderson2018r2r,ku2020rxr,chang2017matterport,yadav2022hm3dsem}. Sampling uses seed 0 and balances scenes and ObjectNav categories.

Astra receives one forward-facing $512\times384$ RGB image, the instruction or object category, a step counter, and action feedback. GPS, compass, depth, semantic labels, goal coordinates, reference routes, and evaluation distances are withheld. A persistent episode session selects discrete moves or local waypoints. The adapter converts these to 0.25\,m forward steps and $30^\circ$ turns without path planning, under a 500-primitive budget. Habitat~\cite{savva2019habitat} executes these primitives; VLN-CE enables simulator sliding, whereas ObjectNav disables it.

Success requires STOP within 3\,m of the VLN-CE goal or strictly within 1\,m of the ObjectNav annotated goal-viewpoint set. The latter is not distance to the object's surface and does not require final-image visibility; HM3D records are rescored at this tolerance. We report success rate (SR), success weighted by path length (SPL), and VLN-CE route agreement through nDTW and sDTW~\cite{anderson2018evaluation,ilharco2019dtw}.

\begin{table*}[t]
\caption{\textbf{Navigation on four fixed subsets, 50 episodes each.} Task, dataset, and split are listed separately. All metric values are percentages; RxR uses English guide instructions. ObjectNav has no instruction-specific reference trajectory.}
\label{tab:navigation}

\centering\small
\setlength{\tabcolsep}{11.63pt}
\begin{tabular}{lllrrrrr}
\toprule
\rowcolor{tablehead}
\reporthead{Task} & \reporthead{Dataset} & \reporthead{Split} & \reporthead{Successes} & \reporthead{SR $\uparrow$} & \reporthead{SPL $\uparrow$} & \reporthead{nDTW $\uparrow$} & \reporthead{sDTW $\uparrow$}\\
\midrule
\input{tables/navigation-results.tex}
\bottomrule
\end{tabular}

\end{table*}

\begin{figure*}[t]
\centering
\includegraphics[width=\textwidth]{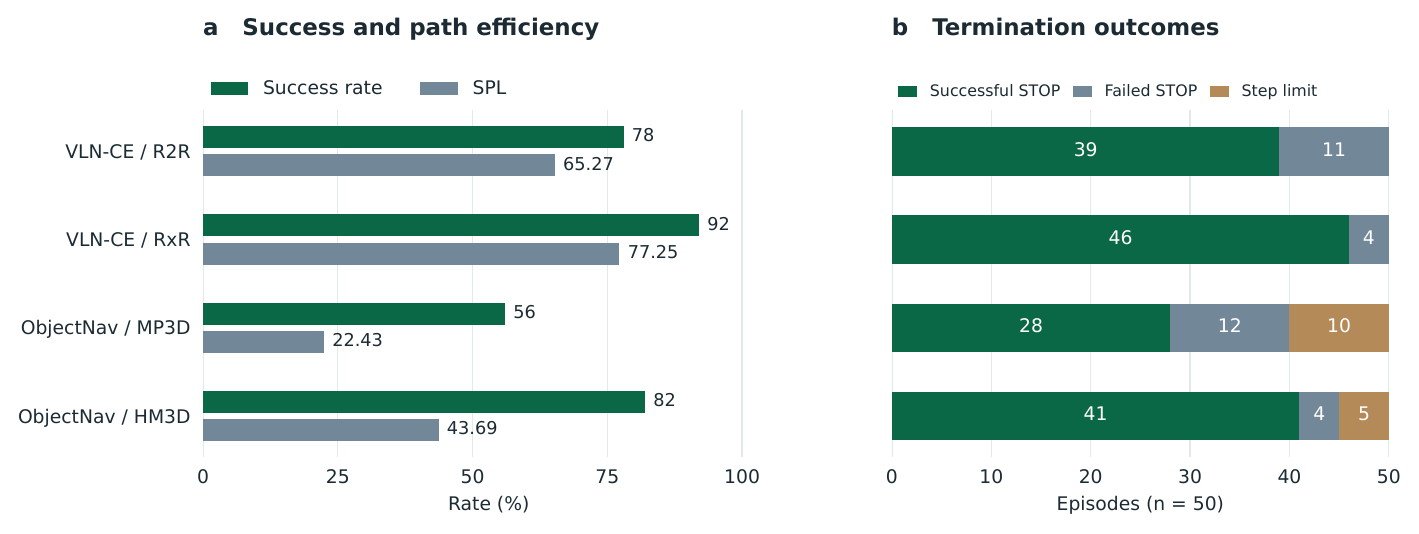}
\caption{\textbf{Navigation success, efficiency, and termination.} (a) SR and SPL. (b) Successful STOP, failed STOP, or exhaustion of the 500-step budget. Each subset contains 50 episodes.}
\label{fig:navigation-results}
\end{figure*}

\subsection{Success and Search Efficiency}
Astra completes 39/50 R2R and 46/50 RxR episodes, with nDTW of 72.20\% and 84.73\%, showing goal reaching and route agreement in Table~\ref{tab:navigation}.

ObjectNav yields 28/50 MP3D and 41/50 HM3D successes, but SPL is only 22.43\% and 43.69\%. SPL discounts excess travel even when the final endpoint succeeds; the gap therefore reveals inefficient search that SR alone conceals. Figure~\ref{fig:navigation-results} shows ten budget-limited MP3D episodes and five for HM3D; all evaluated VLN-CE episodes end with STOP.

A route instruction supplies spatial cues; an object category leaves the route to be discovered. Search thus requires deciding where to explore and when revisiting an area is no longer useful, in addition to grounding local observations. The subsets use different scenes, goals, instructions, and sliding settings.

\begin{table*}[t]
\caption{\textbf{Local evaluation of released navigation policies on the same episode lists.} Each dataset contains 50 episodes per system; values are percentages. The first three systems share the front-camera specification. OmniNav Flow uses three wider-angle views and is separated accordingly. All systems share task-specific scoring and primitive-action budgets, while their history processing, prompts, and action adapters differ.}
\label{tab:navigation-open-models}

\centering\small
\setlength{\tabcolsep}{10.96pt}
\begin{tabular}{lcrrrrrrrr}
\toprule
\rowcolor{tablehead}
&   &  \multicolumn{4}{c}{\reporthead{VLN-CE}}  &  \multicolumn{4}{c}{\reporthead{ObjectNav}} \\
\cmidrule(lr){3-6}\cmidrule(lr){7-10}
\rowcolor{tablehead}
&   &  \multicolumn{2}{c}{\reporthead{R2R}}  &  \multicolumn{2}{c}{\reporthead{RxR}}  &  \multicolumn{2}{c}{\reporthead{MP3D}}  &  \multicolumn{2}{c}{\reporthead{HM3D}} \\
\cmidrule(lr){3-4}\cmidrule(lr){5-6}\cmidrule(lr){7-8}\cmidrule(lr){9-10}
\rowcolor{tablehead}
\reporthead{System} & \reporthead{Views} & \reporthead{SR} & \reporthead{SPL} & \reporthead{SR} & \reporthead{SPL} & \reporthead{SR} & \reporthead{SPL} & \reporthead{SR} & \reporthead{SPL}\\
\midrule
\input{tables/navigation-open-models.tex}
\bottomrule
\end{tabular}
\end{table*}

\subsection{Local Comparisons and Failure Modes}
Released LightNav-0, Uni-NaVid 7B, OmniNav Flow, NavFoM and SPAN-Nav policies are evaluated on the same 200 episodes with shared scoring and primitive budgets~\cite{wang2026lightnav,zhang2025uninavid,xue2025omninav,zhang2026embodied,liu2026span}. LightNav-0 and Uni-NaVid use Astra's front-camera specification in place of their published camera settings; OmniNav Flow receives three wider-angle views. Local adapters convert LightNav-0 and OmniNav Flow outputs to the shared discrete actions. The OmniNav condition evaluates the released Flow policy without the full system's slow exploration planner.

Under this local protocol, Astra has higher SR and SPL than the evaluated systems on all subsets with reported results in Table~\ref{tab:navigation-open-models}. Relative to LightNav-0, SR gains are 20, 20, 36, and 16 percentage points. LightNav-0 has slightly higher R2R nDTW, 72.77\% versus 72.20\%, while Astra has higher sDTW, 59.35\% versus 48.89\%. Appendix~\ref{app:navigation-open-models} details the camera settings and action adapters.

\begin{figure*}[t]
\centering
\includegraphics[width=.9\textwidth]{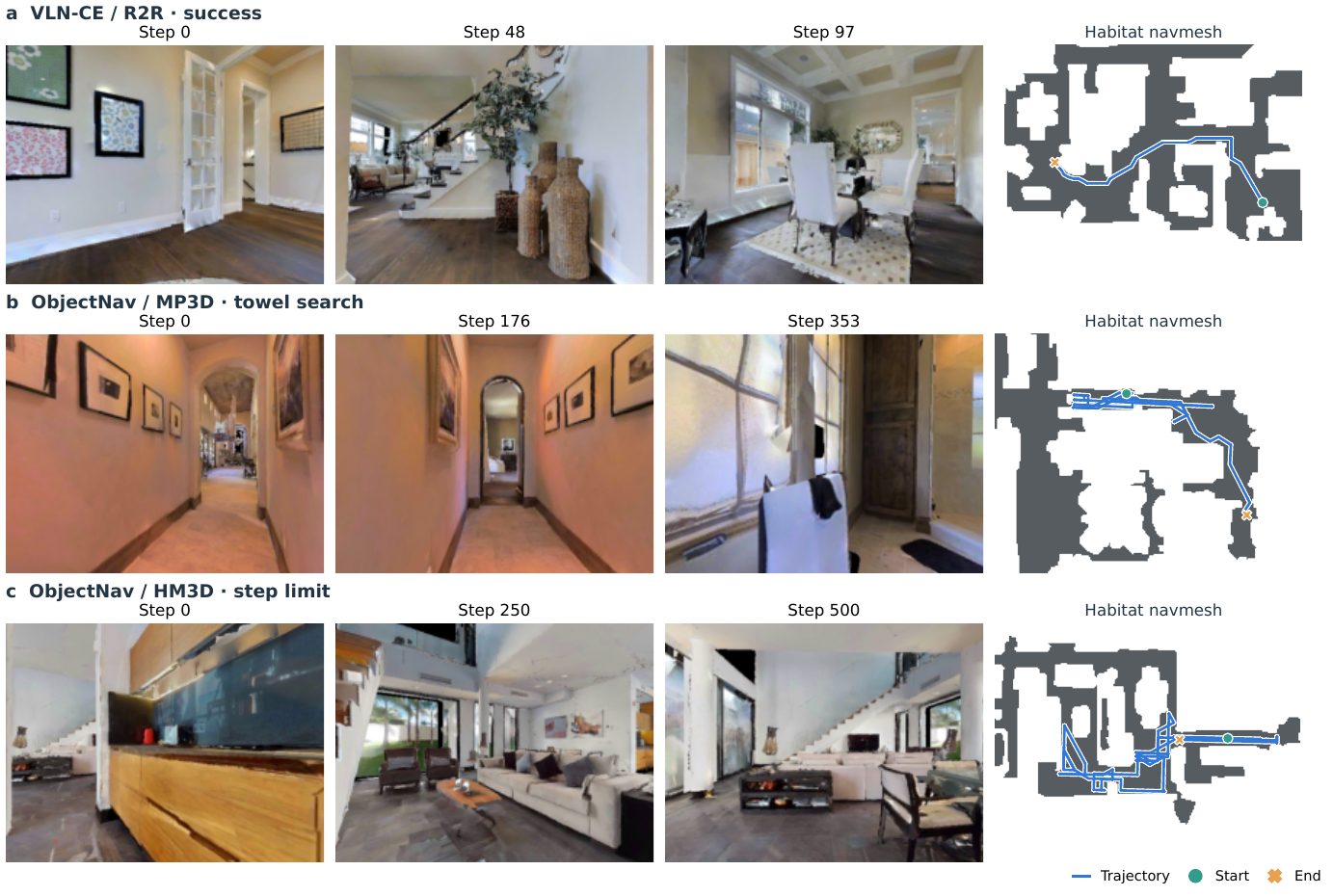}
\caption{\textbf{Two navigation successes and one budget-limited failure.} Rows show initial, midpoint, and final front views beside the executed route on the Habitat navigation mesh. These evaluator-only maps were not shown to Astra.}
\label{fig:navigation-cases}
\end{figure*}

Successful trajectories show Astra interpreting instructions, tracking landmark sequences, and recognizing target categories through multiple indoor stages. Figure~\ref{fig:navigation-cases} illustrates instruction following, search with detours, and budget exhaustion. Failure trajectories expose difficulties linking visual observations to specific locations, verifying goal completion, and using collision feedback or search history to change course. A plausible rationale for moving does not ensure correct goal judgment or spatial progress.

See Appendix~\ref{app:navigation-mobile} for protocols and termination counts.

%% file: tables/navigation-results.tex
\rowcolor{white}
VLN-CE & R2R & Val-Unseen & 39/50 & 78 & 65.27 & 72.20 & 59.35 \\
\rowcolor{tablestripe}
VLN-CE & RxR (English) & Val-Unseen & 46/50 & 92 & 77.25 & 84.73 & 80.42 \\
\midrule
\rowcolor{white}
ObjectNav & MP3D & Val & 28/50 & 56 & 22.43 & -- & -- \\
\rowcolor{tablestripe}
ObjectNav & HM3D & Val & 41/50 & 82 & 43.69 & -- & -- \\

%% file: tables/navigation-open-models.tex
\rowcolor{white}
Astra & 1 & 78.00 & 65.27 & 92.00 & 77.25 & 56.00 & 22.43 & 82.00 & 43.69 \\
\rowcolor{tablestripe}
LightNav-0 & 1 & 58.00 & 54.14 & 72.00 & 64.08 & 20.00 & 8.39 & 66.00 & 37.75 \\
\rowcolor{white}
Uni-NaVid 7B & 1 & 30.00 & 27.76 & 34.00 & 25.67 & 8.00 & 5.46 & 30.00 & 20.48 \\
\midrule
\rowcolor{tablegroup}
\multicolumn{10}{l}{\emph{Multi-view methods (separate observation regime)}} \\
\rowcolor{white}
OmniNav Flow & 3 & 56.00 & 53.89 & 72.00 & 56.29 & 4.00 & 2.96 & 6.00 & 2.90 \\
NavFoM & 3 & 44.00 & 40.04 & 58.00 & 51.79 & 18.00 & 13.01 & 42.00 & 30.38 \\
SPAN-Nav & 3 & 70.00 & 61.71 & 72.00 & 56.19 & \navdash & \navdash & \navdash & \navdash \\

%% file: sec/4b_1_humanoid_locomotion.tex
\section{Locomotion}
\label{sec:humanoidlocomotion}

Navigation primitives implement the motion Astra chooses. Locomotion exposes an additional demand: constructing the body motion itself. Here Astra generates dense motion references for a frozen tracker over repeated attempts on one obstacle course. Analytic tests examine compact interfaces without Astra.

\subsection{Dense Motion on a Matched Course}
\label{sec:astra-passage}
We replace PASSAGE's learned motion generator with Astra while retaining its scene-aligned $25\times65$ representation and frozen G1 ScaleTrack controller~\cite{ma2026passagescalingscenealignedmotion,zeng2026scalebfm}. Each proposal specifies 0.5\,s at 50\,Hz: root height, projected gravity, planar velocity, yaw rate, and positions and velocities for 29 joints. Each call requires 1,625 consistent values.

On one open-clutter course, Astra and PASSAGE share the initial state, goal, elevation observation, local route guidance, tracker, and termination criteria. Success requires final planar error at most 0.5\,m. Falls trigger resets; eight seconds with less than 0.10\,m root displacement triggers no-progress termination. The supplied route leaves motion construction to Astra.

Astra makes five sequential attempts. The first is zero-shot; later attempts retain textual summaries of progress, falls, and collisions. Before the fifth, Astra also receives a retargeted walking clip from another course. Experience and motion context therefore change without weight updates. This is one adaptation trajectory, not five independent trials.

\begin{table*}[t]
\caption{\textbf{PASSAGE and sequential Astra attempts on one matched
open-clutter course.}  Progress is the reduction in planar distance to the
goal, so a negative value denotes motion away from it.  The Astra rows are
successive stages of one adaptation trajectory, not independent trials.}
\label{tab:astra-passage-fixed-scene}
\centering\small
\setlength{\tabcolsep}{12.20pt}
\begin{tabular}{llrrrcc}
\toprule
\rowcolor{tablehead}
\reporthead{Planner / Attempt} & \reporthead{Outcome} & \reporthead{Time (s)} & \reporthead{Progress (m)} & \reporthead{Final error (m)} & \reporthead{Fall} & \reporthead{Reached}\\
\midrule
\rowcolor{white}
PASSAGE
& Goal reached
& 13.18 & 8.622 & 0.495 & No & Yes \\
\rowcolor{tablestripe}
Astra Attempt 1
& Fall
& 9.80 & 1.561 & 7.556 & Yes & No \\
\rowcolor{white}
Astra Attempt 2
& No progress
& 20.28 & -0.024 & 9.141 & No & No \\
\rowcolor{tablestripe}
Astra Attempt 3
& No progress
& 8.00 & 0.012 & 9.104 & No & No \\
\rowcolor{white}
Astra Attempt 4
& No progress
& 8.48 & 0.159 & 8.958 & No & No \\
\rowcolor{tablestripe}
Astra Attempt 5
& Time horizon
& 30.00 & 2.598 & 6.519 & No & No \\
\bottomrule
\end{tabular}
\end{table*}

\subsection{Persistent Locomotion Failures}
PASSAGE reaches the goal in 13.18\,s. Astra initially advances 1.561\,m before falling; the next three attempts remain upright with little useful progress. Table~\ref{tab:astra-passage-fixed-scene} shows that the fifth attempt stays upright for 30\,s, travels 3.833\,m, and reduces goal distance by 2.598\,m. It still ends 6.519\,m from the goal, before the principal obstacle field.

After five rounds of trial and error, Astra still does not achieve reliable locomotion through this dense-reference interface. Later attempts improve stability and produce some forward motion, but body and limb coordination is inconsistent. Across attempts, Astra uses motion feedback while the walking clip, gait timing, and amplitude are adjusted together.

The interface helps explain the difficulty. Joint limits alone do not make a reference coherent: root velocity, limb posture, and their evolution must describe compatible motion. A tracker can follow a reference yet produce in-place stepping, unwanted contacts, or a fall. A supplied route simplifies planning but does not supply coordinated body motion.

Timing is also limiting. The final 30\,s episode uses 250 synchronous calls averaging 39.86\,s of recorded latency, versus approximately 0.08\,s for PASSAGE planning. Physics pauses during inference. The inference budget permits five attempts.

\subsection{Compact-Reference Diagnostics without Astra}
\label{sec:humanoid-interface-ablation}
We additionally test two compact-reference interfaces with analytic generators, without Astra, to identify failures in converting references into robot motion. Both use six-frame link-pose sequences for the same 29-DoF robot. Five-point specifies the pelvis, wrists, and ankles; WholeBody-14 adds the torso and bilateral hips, knees, shoulders, and elbows.

The deterministic generators run on three open-clutter and three low-ceiling courses in two physics backends: 12 rollouts per interface. Strict success requires an upright robot, final error at most 0.2\,m, a five-second hold, tracked points within the course, and no obstacle force above 5\,N. Goal + stop requires only arrival and a hold.

\begin{table*}[t]
\caption{\textbf{Paired analytic interface feasibility study over six
courses and two physics backends.}  Each course--backend cell contains one
rollout.  Contact magnitudes are useful failure diagnostics, but are not
treated as calibrated measurements across simulators.  These rollouts do
not use Astra.}
\label{tab:interface-proxy}
\centering\small
\setlength{\tabcolsep}{4.18pt}
\begin{tabular}{lrrrrrrrrr}
\toprule
\rowcolor{tablehead}
\reporthead{Interface} & \reporthead{$N$} & \reporthead{\shortstack{Strict\\success}} & \reporthead{\shortstack{Goal\\+ stop}} & \reporthead{Falls} & \reporthead{\shortstack{Final\\error (m)}} & \reporthead{\shortstack{Travel\\(m)}} & \reporthead{\shortstack{Peak contact\\(N)}} & \reporthead{\shortstack{Non-foot contact\\(N)}} & \reporthead{\shortstack{Five-point\\RMSE (m)}}\\
\midrule
\rowcolor{white}
Five-point
& 12 & 0 & 4 (33.3\%) & 1 (8.3\%)
& 3.441 & 5.745 & 1853 & 644 & 0.161 \\
\rowcolor{tablestripe}
WholeBody-14
& 12 & 0 & 3 (25.0\%) & 4 (33.3\%)
& 5.334 & 3.866 & 2293 & 1562 & 0.197 \\
\bottomrule
\end{tabular}
\end{table*}

Neither interface achieves strict success in Table~\ref{tab:interface-proxy}. Five-point reaches and holds the endpoint in 4/12 rollouts with one fall; WholeBody-14 does so in 3/12 with four falls. Five-point has lower mean endpoint and tracking errors. Low ceilings remain difficult, with only 1/6 and 0/6 rollouts reaching the endpoint.

The comparison favors the tested Five-point generator--interface pair descriptively, without isolating dimensionality because the generators differ. Large obstacle contacts also show that trackable motion is not necessarily safe motion.

%% file: sec/4b_humanoid.tex
\section{Humanoid Loco-Manipulation}
\label{sec:humanoid}

We study Hybrid humanoid control, with Astra directing pretrained whole-body controllers. In HumanoidBench~\cite{sferrazza2024humanoidbench}, Astra commands frozen Humanoid-GPT~\cite{qi2026humanoidgpt} policies across a broad task suite. In SIMPLE~\cite{wei2026simple}, it uses a frozen ScaleBFM~\cite{zeng2026scalebfm} whole-body controller with task instructions refined through experience, examining manipulation and the retention of successful behavior. Astra selects task commands; whole-body controllers coordinate motion.

\subsection{HumanoidBench Evaluation Setup}
We evaluate the 30 executable HumanoidBench tasks in the pinned registry, excluding the two Highbar variants. Evaluation uses seeds 1 and 2 for each task. Torque replays independently verify episodes.

Astra receives RGB, privileged state, task geometry, and guidance. It selects walking velocities and yaw rate, or sparse pelvis and wrist references. Frozen whole-body controllers run at 50\,Hz over 500-Hz physics and joint control. Wrist targets couple to the body and depend on the control mode.

\begin{figure*}[t]
  \centering
  \includegraphics[width=\textwidth,height=.70\textheight,keepaspectratio]{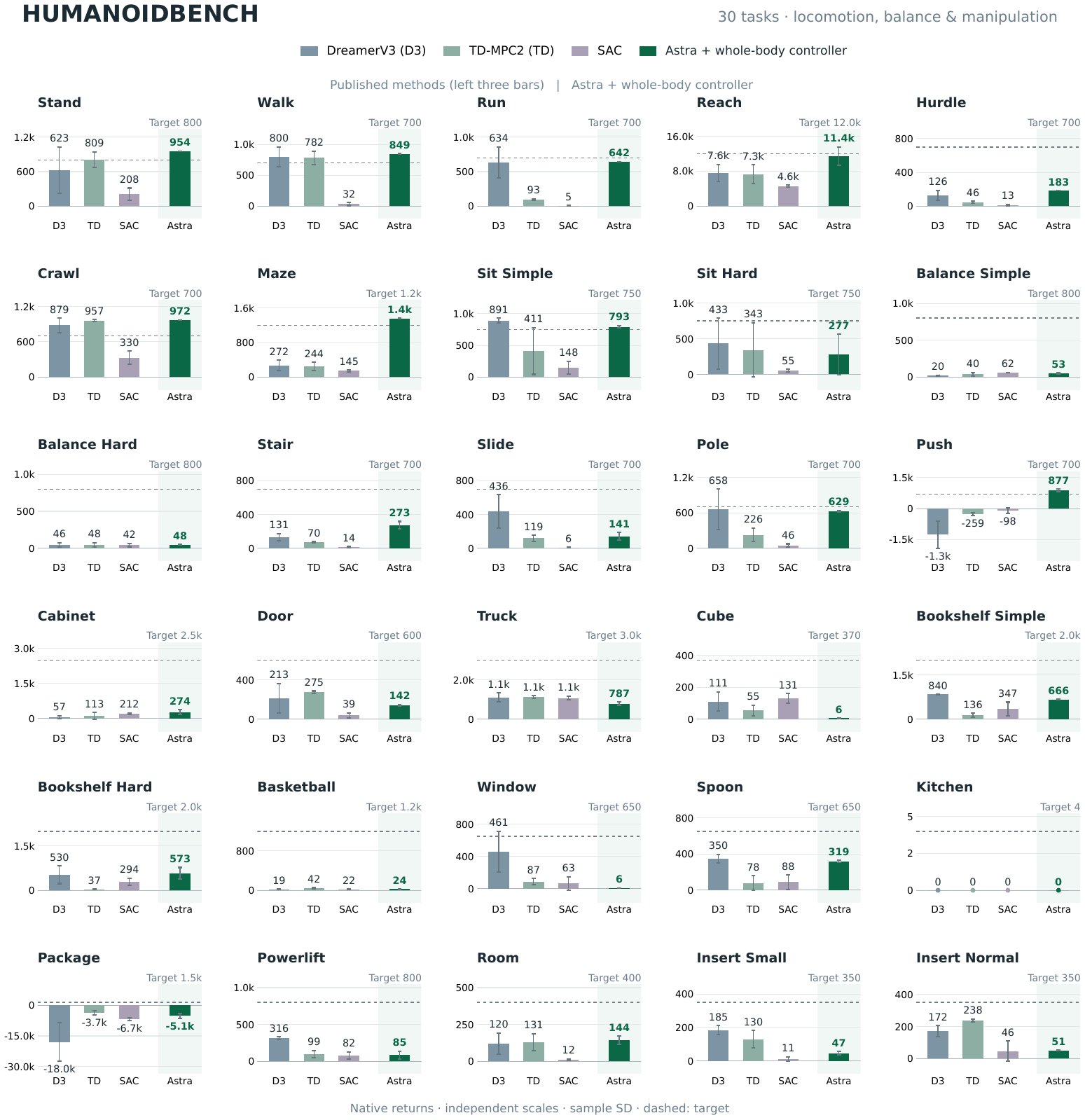}
  \caption{\textbf{Results across 30 HumanoidBench tasks.} Green bars show Astra mean returns and sample SDs. Left bars reproduce published method returns~\cite{sferrazza2024humanoidbench}. Axes are task-specific; dashed lines mark native thresholds. Appendix~\ref{app:g1} gives exact values and reference protocols.}
  \label{fig:humanoid}
\end{figure*}

Table~\ref{tab:g1-summary} compares Astra with published methods on eight tasks; Figure~\ref{fig:humanoid} covers all 30. Appendix~\ref{app:g1} gives complete Astra results and evaluation details.

\begin{table*}[t]
\caption{\textbf{HumanoidBench returns against published methods.} Entries are mean returns; bold marks the highest value per task. DreamerV3, TD-MPC2, and SAC results follow HumanoidBench~\cite{sferrazza2024humanoidbench}. Threshold denotes the native task criterion. Appendix~\ref{app:g1} documents result sources and evaluation protocols.}
\label{tab:g1-summary}

\centering\small
\setlength{\tabcolsep}{12pt}
\begin{tabular}{lrrrrr}
\toprule
\rowcolor{tablehead}
\reporthead{Task} & \reporthead{DreamerV3} & \reporthead{TD-MPC2} & \reporthead{SAC} & \reporthead{Astra} & \reporthead{Threshold}\\
\midrule
\rowcolor{white}
Maze & 272.3 & 244.3 & 144.8 & \textbf{1358.8} & 1200 \\
\rowcolor{tablestripe}
Reach & 7580.9 & 7316.1 & 4565.1 & \textbf{11430.2} & 12000 \\
\rowcolor{white}
Walk & 800.2 & 782.0 & 31.7 & \textbf{848.7} & 700 \\
\rowcolor{tablestripe}
Run & 633.8 & 93.3 & 5.0 & \textbf{642.1} & 700 \\
\rowcolor{white}
Crawl & 878.8 & 957.4 & 330.0 & \textbf{971.9} & 700 \\
\rowcolor{tablestripe}
Stair & 131.1 & 70.4 & 14.1 & \textbf{272.5} & 700 \\
\rowcolor{white}
Push & -1251.9 & -258.7 & -97.9 & \textbf{877.3} & 700 \\
\rowcolor{tablestripe}
Door & 213.0 & \textbf{274.7} & 39.4 & 142.4 & 600 \\
\bottomrule
\end{tabular}

\end{table*}

\subsection{Command Selection with Pretrained Whole-Body Controllers}
Maze provides a clear example of adapting commands to state. Astra slows near intersections, turns, and resumes forward motion using supplied geometry. Maze achieves a mean return of 1358.8, with the evaluated trajectories reaching native stage 4. Reach achieves 11430.2, below the 12000 threshold, with individual returns straddling it.

Walk combines posture calibration with startup and speed guidance, achieving a mean return of 848.7. Run uses a five-point gait and reaches 642.1; both runs stay upright but below threshold. Crawl uses a low, forward-pitched gait and reaches 971.9, with about 27.39\,m displacement per run. Astra selects speed and heading, while calibrated generators supply posture and limb coordination. Performance reflects Astra's decisions and whole-body controllers.

\begin{figure*}[t]
  \centering
  \includegraphics[width=\textwidth]{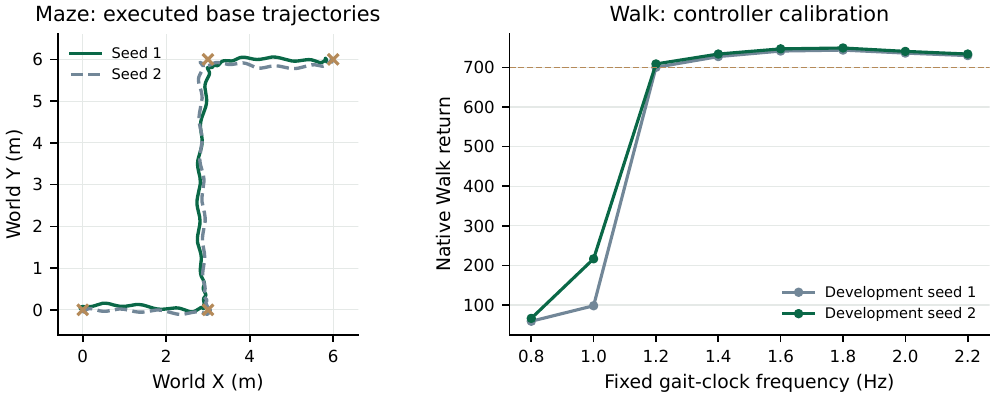}
  \par\smallskip
  \begin{minipage}{.24\textwidth}\centering
    \includegraphics[width=\linewidth]{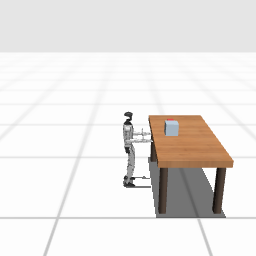}\\[-2pt]
    {\scriptsize Push: initial state (0)}
  \end{minipage}\hfill
  \begin{minipage}{.24\textwidth}\centering
    \includegraphics[width=\linewidth]{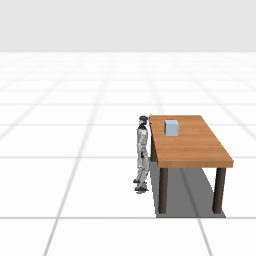}\\[-2pt]
    {\scriptsize Base staged (175)}
  \end{minipage}\hfill
  \begin{minipage}{.24\textwidth}\centering
    \includegraphics[width=\linewidth]{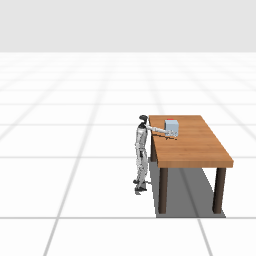}\\[-2pt]
    {\scriptsize Contact approach (267)}
  \end{minipage}\hfill
  \begin{minipage}{.24\textwidth}\centering
    \includegraphics[width=\linewidth]{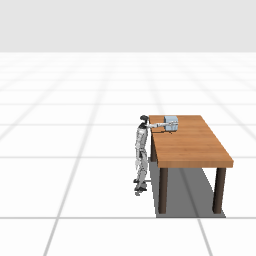}\\[-2pt]
    {\scriptsize Native success (323)}
  \end{minipage}
  \caption{\textbf{Decisions and physical response in three separate experiments.} Top left: Maze trajectories and supplied checkpoints; both runs reach stage 4. Top right: a scripted Walk frequency sweep with fixed weights and command speed; the dashed line marks the native threshold. Bottom: experience-conditioned Push, seed 7, ending at 4.81\,cm box error. Parentheses give control steps.}
  \label{fig:mechanisms}
\end{figure*}

Astra achieves the highest mean native return among the compared methods: 581.9, versus 338.2 for TD-MPC2, 42.7 for SAC, and $-41.9$ for DreamerV3, averaging task means equally across all 30 tasks in Figure~\ref{fig:humanoid}. Astra also exceeds the published returns of DreamerV3 on 16 tasks, TD-MPC2 on 19, and SAC on 25. They exceed the best of these three references on 13 tasks and tie on Kitchen~\cite{sferrazza2024humanoidbench}. Six Astra task means reach native thresholds: Stand, Walk, Maze, Crawl, Push, and Sit Simple.

Calibration also affects what a command accomplishes. In the scripted Walk sweep, changing gait-clock frequency from 1.2 to 1.8\,Hz raises mean return from 705.72 to 746.60 across eight paired seeds without changing weights, command speed, or feedback rule. Actual forward speed is about 1.48\,m/s for a 1.25\,m/s command. Return and measured speed together describe the effect of controller calibration.

\subsection{Contact Failures Between Task Stages}
\label{sec:focused}
Push illustrates useful preparation and limited transfer. A demonstration-guided configuration succeeds on three development seeds and one transfer seed fixed in advance with box errors below 5\,cm; two development trajectories reproduce supplied demonstrations. On the evaluation cases, staging and contact guidance support successful pushing in 277 and 357 controls, at 4.89 and 4.97\,cm error.

A separate scripted diagnostic clarifies the interaction: changing entry alone or contact references alone fails, while changing both reaches 4.97\,cm error. A hand offset depends on the body's configuration. The records in Figure~\ref{fig:mechanisms} and Appendix~\ref{app:evidence} show why base preparation and contact must be coordinated: an effective offset can fail from a different entry state.

Door exposes the next difficulty. Episodes rotate the hatch without opening a usable passage; scripted finger and body variants achieve partial opening without passage, and Astra also fails to complete the task. Reaching a mechanism is insufficient if contact fails under the loading and body movement needed to continue. Stair achieves a mean return of 272.5 with terrain-aware references, but one run falls after limited progress and the other retreats. Cube and Window lose objects early; Kitchen returns zero.

These failures occur between task stages, when changes in posture or contact disrupt the next action. Whole-body controllers alone do not resolve them. SIMPLE examines whether guidance drawn from experience helps Astra manage these transitions during grasping, transport, and handover.

\subsection{Refining Task Guidance through Experience in SIMPLE}
\label{sec:simple}
The SIMPLE study examines whether experience can help Astra coordinate grasping, transport, and handover. Using visual and proprioceptive feedback, Astra selects actions for a G1 humanoid, while ScaleBFM supplies whole-body motion. After each trial, Astra reviews the outcome and revises the instructions and notes used in subsequent attempts. This adaptation changes the written guidance while keeping both models' weights fixed.

We evaluate Astra using the official SIMPLE codebase and L2 evaluation protocol, covering six tasks with ten scenes each. Astra succeeds in 50 of 60 trials, achieving an overall success rate of 83.3\% (Table~\ref{tab:simple-l2}). Performance is strongest in tabletop and picking tasks, while mobile pick/place remains more difficult. Appendix~\ref{app:simple} details the protocol and separate retention checks.

\begin{table}[!tb]
\caption{\textbf{Astra with ScaleBFM on SIMPLE L2.} Success under the official evaluation protocol, with ten scenes per task.}
\label{tab:simple-l2}
\centering\small
\setlength{\tabcolsep}{6pt}
\begin{tabular}{>{\raggedright\arraybackslash}p{\dimexpr\columnwidth-98pt-6\tabcolsep\relax}rr}
\toprule
\rowcolor{tablehead}
\reporthead{Task} & \reporthead{Successes} & \reporthead{Rate}\\
\midrule
\rowcolor{white}
Handover & 8/10 & 80\% \\
\rowcolor{tablestripe}
Mobile pick/place & 7/10 & 70\% \\
\rowcolor{white}
Tabletop & 9/10 & 90\% \\
\rowcolor{tablestripe}
XMove bend pick & 9/10 & 90\% \\
\rowcolor{white}
XMove pick & 9/10 & 90\% \\
\rowcolor{tablestripe}
Bend & 8/10 & 80\% \\
\midrule
\rowcolor{tablesummary}
\textbf{Total} & \textbf{50/60} & \textbf{83.3\%} \\
\bottomrule
\end{tabular}
\end{table}

\subsection{Why Handover Remains Fragile}
Handover reveals a limitation hidden by the aggregate score: native success does not require the carton to remain stable after complete release. In the observed failures, finger closure or carton tilt can be mistaken for a secure grasp while the table or the other hand still carries the load. Moving the receiving hand can then shift the supplying wrist as the body adjusts. A grasp that appears adequate at one stage may therefore fail during the next.

The grasp procedure addresses this problem by checking support before advancing. It instructs Astra to separate grasping from lifting, withdraw the assisting hand, and check that the entire carton rises and moves with the supplying hand. If it merely rotates or the body keeps leaning forward, the procedure calls for restoring support and trying again. These instructions specify physical checks; their reliability still depends on Astra's observations and actions.

The handover guidance does not consistently preserve successful behavior across repeated attempts. In a sequence of instruction-adaptation trials, a successful handover is followed by a failed attempt after further reflection. A fixed procedure tested on three previously successful scenes passes native success and an additional hold check on only two. The recovery trials in Appendix~\ref{app:simple} also fail on all three valid attempts despite longer budgets. A diagnosis need not yield correction or repeatable success.

\begin{figure*}[t]
\centering
\includegraphics[width=\textwidth]{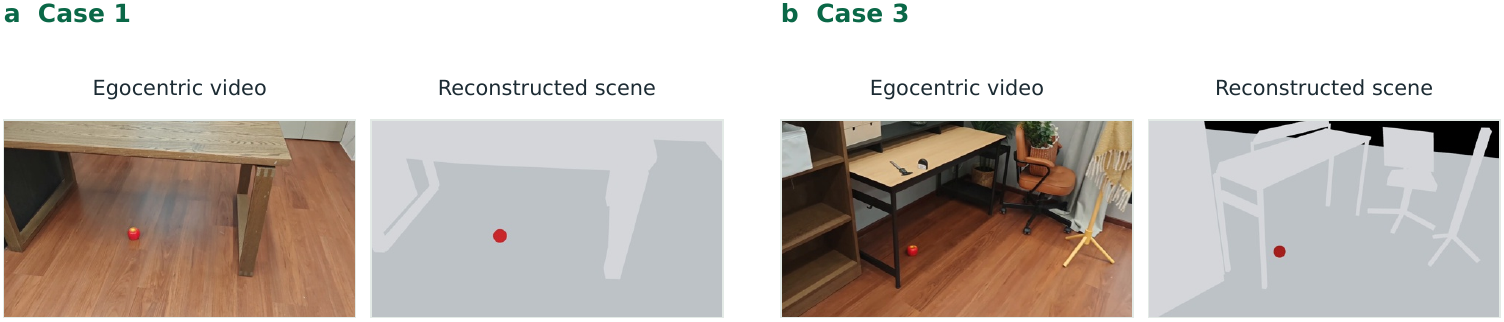}
\caption{\textbf{Interactive scenes from egocentric video.} Source frames and reconstructed geometry for two apple-transfer cases. Object and support layout is recovered, while table edges and furniture remain approximate.}
\label{fig:egocentric-scenes}
\end{figure*}

\subsection{Scene and Motion from Egocentric Video}
\label{sec:egocentric}
Astra also builds and refines a pipeline that converts egocentric videos into interactive scenes and humanoid motion references. It assembles perception and retargeting tools, inspects their outputs, and corrects geometric and temporal inconsistencies. Camera estimates, object roles, and support relations define the scene; declared camera or support heights provide metric scale. Motion references specify pelvis, wrist, and ankle poses together with hand closure. They are generated from video-derived geometry or through EgoAllo body estimation~\cite{yi2024egoallo}, SMPL-X fitting~\cite{pavlakos2019smplx}, and motion retargeting~\cite{araujo2025gmr}.

In a kitchen with motion-capture-calibrated cameras, image-based refinement reduces scene Chamfer-L1 error from 35.93 to 24.73\,mm; incorporating CAD dimensions reduces it further to 12.14\,mm. Transferring to other recordings exposes the importance of object placement: reusing a bottle position displaced by a few centimeters causes grasp failures. Figure~\ref{fig:egocentric-scenes} shows reconstructions from separate everyday videos. Across three apple-transfer cases, complete-video references support one complete interaction: opening a bin by its pedal, releasing the apple inside, and allowing the lid to close. The other two cases still lose the apple during lifting despite local scene adjustments.

The central challenge is preserving interaction structure through reconstruction and retargeting. Support surfaces, container interiors, and movable mechanisms must remain usable, while motion references must connect approach, grasp, transport, and release in the correct order. The successful bin interaction preserves the approach and pedal contact before release; local grasp corrections alone leave the task incomplete. Astra can identify and revise these dependencies, but geometric plausibility does not ensure a successful physical interaction. The in-the-wild study consumed 217\,million tokens across development, evaluation, and associated documentation, highlighting the cost of per-video refinement.

%% file: sec/4_analysis.tex
\section{Discussion}
\label{sec:analysis}

Across the six domains, Astra shows strong navigation results and useful manipulation capabilities, alongside persistent difficulties in contact coordination, locomotion, and completion verification. To interpret this uneven performance, we examine how control is divided and why feedback does not consistently turn useful local decisions into successful, efficient task behavior.

\begin{table*}[t]
\caption{\textbf{Inference demands across studies.} M denotes million tokens; rounded totals may not sum exactly. Units and recording scope are listed for each study. Full manipulation accounting appears in Tables~\ref{tab:robodojo-resources} and~\ref{tab:robocasa-cost}.}
\label{tab:resource-summary}

\centering\small
\setlength{\tabcolsep}{5.00pt}
\begin{tabular}{>{\raggedright\arraybackslash}p{\dimexpr0.190\textwidth-2\tabcolsep\relax}>{\raggedright\arraybackslash}p{\dimexpr0.450\textwidth-2\tabcolsep\relax}>{\raggedright\arraybackslash}p{\dimexpr0.360\textwidth-2\tabcolsep\relax}}
\toprule
\rowcolor{tablehead}
\reporthead{Setting} & \reporthead{Recorded resource use} & \reporthead{Scope and interpretation}\\
\midrule
\rowcolor{white}
RoboDojo, 50 task instances per condition & Hybrid / Direct: 624.8M / 1,132.3M total tokens; 607.6M / 1,107.3M cached input; 15.90M / 22.91M uncached input; 1.31M / 2.09M output & Earlier attempts excluded. Cached input dominates the totals. \\
\addlinespace
\rowcolor{tablestripe}
RoboCasa, 75 episodes per condition & Direct / Hybrid: 7,910 / 8,941 recorded model requests; median 85 / 108 per episode & Counts cover saved request logs, including recovery history, rather than all inference calls. \\
\addlinespace
\rowcolor{white}
Dense locomotion, final attempt & 250 synchronous calls for 30\,s of simulated robot motion; mean model latency 39.86\,s per call & Physics pauses during inference; PASSAGE planning takes approximately 0.08\,s per call. \\
\bottomrule
\end{tabular}

\end{table*}

\subsection{Capability Depends on the Complete Control Path}
An action acquires physical meaning through its interface and controller. A navigation move invokes a discrete motion primitive; a dense motion reference requires compatible body and joint targets; a whole-body controller command invokes learned coordination. Evaluating Astra therefore requires specifying both the commands it constructs and the motion other components supply. The varied tasks and controllers provide complementary evidence about the complete control paths.

The in-hand results reveal a tension between stability and progress: retaining the current grasp can prevent the release and repositioning needed for continued rotation. Navigation presents a different problem, since a recognizable landmark need not establish the correct location or justify stopping. Control responsibility thus includes deciding which state changes matter and checking that they occur. These failures require different diagnostics and explanations.

\subsection{Intervention and Verified Progress}
Useful cooperation requires both repairing failures and preserving behavior that already works. In the manipulation cases, Astra can improve the conditions under which a policy acts by redirecting a wrist or preparing contact, without constructing the subsequent trajectory. This division succeeds unevenly: RoboCasa Hybrid completes some instances that defeat the standalone policy and fails on others the policy completes. The trajectories illustrate useful roles for reasoning, while the aggregate comparisons measure the combined effects of policy access, action timing, and feedback management.

Detecting an error, proposing a correction, and verifying its effect are separate demands. A new grasp still needs clearance and support; an adjusted placement still needs a correct completion check. Action-source fractions describe who supplies commands, but neither the number of corrections nor the plausibility of their explanations measures their value. That value rests on the ensuing physical outcome.

\subsection{Retaining Success across Trials}
Feedback can improve a procedure on a known case without making it reliable across attempts. The Push development records include a successful trial followed by a failed repeat. Diagnosis, written guidance, and sustained control thus need separate assessment: a sound explanation can identify what went wrong while leaving the next action inadequate.

This distinction matters when interpreting improvement with fixed model weights. Changes to guidance, interfaces, or whole-body controller calibration can change system performance. Repeated trials reveal whether the resulting behavior persists; transfer to new states tests whether the guidance extends beyond the cases that produced it.

\subsection{Resource Costs and Capability Limits}
\label{sec:resources}
A capability is demonstrated with a particular budget. Long input histories and repeated review increase resource use, while paused physics separates simulated motion time from inference time. Table~\ref{tab:resource-summary} reports recorded resource use.

The two manipulation studies illustrate different effects of cooperation. RoboDojo Hybrid reduces recorded tokens by 44.8\%, yet retains substantial uncached input and output. RoboCasa Hybrid reduces Astra-authored action segments but records more model requests than Direct. Delegating motion therefore does not necessarily delegate the work of reviewing it. Inference efficiency depends on review frequency as well as action generation.

Dense locomotion exposes a more immediate limit: a decision takes far longer than the motion executed before the next observation. Pausing physics makes that delay tolerable in the experiment; a physical scene cannot be assumed to wait. Practical control therefore requires assessing inference delays alongside the rate at which physical feedback changes.

%% file: sec/5_conclusion.tex
\section{Conclusion}
\label{sec:conclusion}

We systematically evaluate GPT-6 Astra as an embodied policy across six domains. Astra shows strengths in navigation and task-oriented manipulation, correcting targets and contact conditions while cooperating with learned policies; pretrained whole-body controllers also support humanoid loco-manipulation. However, direct in-hand control and dense-reference locomotion remain unreliable, and policy assistance can fail on tasks that the standalone policy solves. These findings expose a gap between generating useful action decisions and reliably producing and verifying their physical effects. Learned policies and whole-body controllers help bridge this gap, but effective cooperation still requires Astra to recognize when to intervene and when to return control to the policy. Substantial token use and inference latency further constrain practical control, even when learned policies supply most physical actions.

%% file: sec/6_contributors.tex
\section*{Contributors}
\phantomsection
\label{sec:contributors}

{\small\raggedright Each group lists names alphabetically by surname.\par}

\begin{description}[style=nextline,leftmargin=0pt,labelindent=0pt,
  font=\sffamily\bfseries,itemsep=2pt,parsep=0pt,topsep=0pt]
  \raggedright
  \item[Manipulation]
  Xiaoqian~Cheng, Jiayi~Su, Mi~Yan, Yixin~Zheng

  \item[Navigation]
  Jiahang~Liu, Ruochen~Xu, Tianyu~Xu

  \item[Dexterous Manipulation]
  Yu~Deng, Lihe~Ding, Shaocong~Dong, Xiangjun~Gao, Sikai~Liang, Qingtao~Liu, Zhe~Xu, Siming~Yan

  \item[Locomotion]
  Haozhe~Jia, Zhoujian~Li, Yunrui~Lian, Chenghuai~Lin, Yuxuan~Ma, Xudong~Xu, Zhikai~Zhang, Weiyi~Zhu

  \item[Humanoid Loco-Manipulation]
  Xuchuan~Chen, Zekai~Li, Dairu~Liu, Zekun~Qi, Ruixi~Yu, Jinlu~Zhang, Yintianrun~Zhang

  \item[Report Organization]
  Yu~Deng, Jiahang~Liu, Qingtao~Liu, Yuxuan~Ma, Zekun~Qi, Jiayi~Su, Ruochen~Xu, Siming~Yan

  \item[Supervision]
  He~Wang, Li~Yi, Zhizheng~Zhang
\end{description}

%% file: sec/3_method.tex
\onecolumn
\section{Additional Interface Details}
\label{app:interfaces}
\label{sec:harness}
\label{sec:development}

\begin{figure}[htb]
  \centering
  \includegraphics[width=\textwidth]{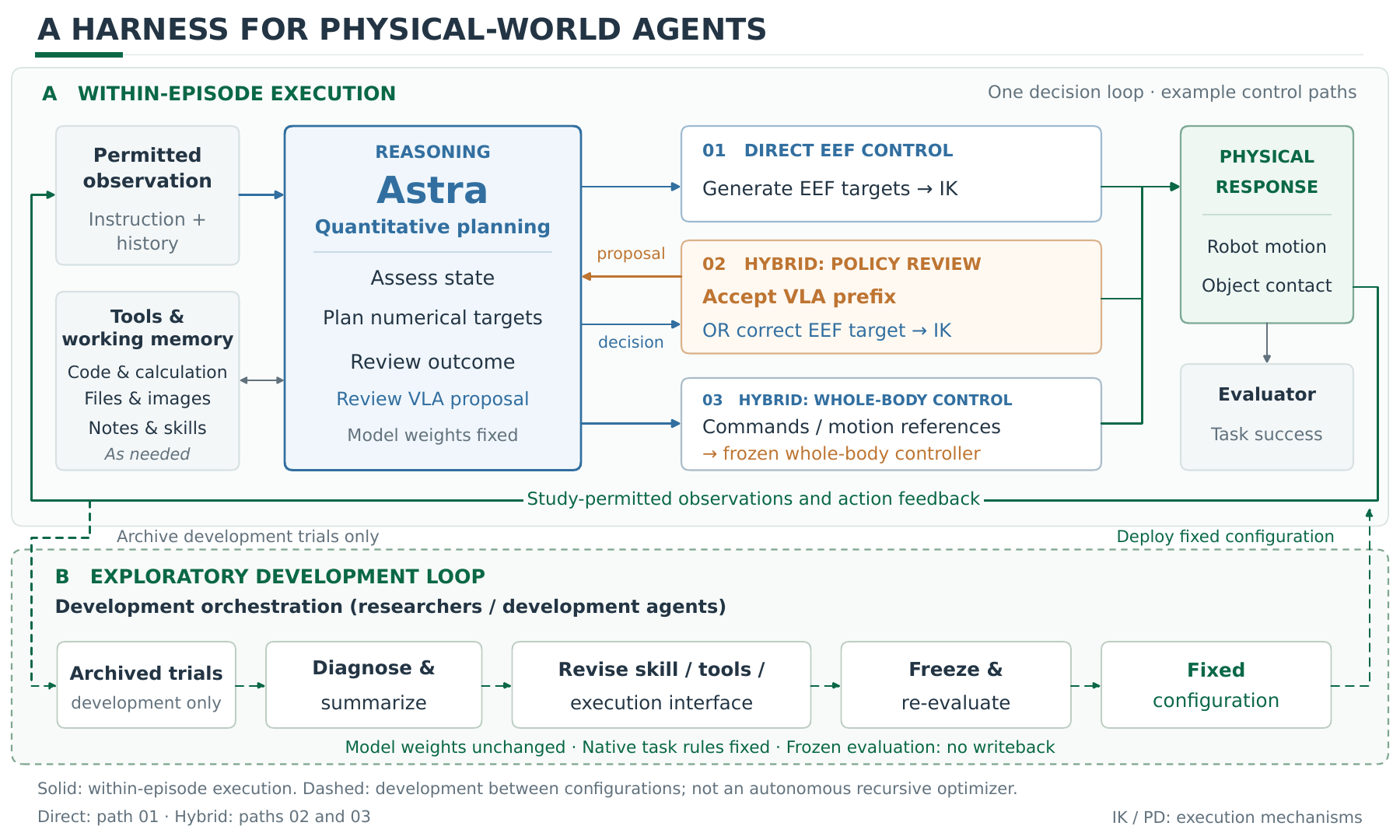}
  \caption{\textbf{Action and development loops.} Path 01 illustrates Direct control; paths 02 and 03 illustrate Hybrid with a task policy or whole-body controller. The action loop returns only study-permitted observations and feedback; task evaluation is a separate branch. The outer loop represents researcher and development-agent changes to guidance, tools, or interfaces before evaluation.}
  \label{fig:capability}
\end{figure}

This appendix specifies the gripper and humanoid interfaces summarized in Section~\ref{sec:method}. Figure~\ref{fig:capability} illustrates how these systems connect action proposals, whole-body controllers, and permitted feedback.

\subsection{Gripper Interfaces}

RoboDojo supplies head and bilateral wrist images, a 14-dimensional proprioceptive state, measured end-effector poses, the instruction, execution history, progress, step budget, and completion rules. Hidden object states, future trajectories, rollback, and hidden planners are excluded. Policy images remain $640\times480$; Astra generally views previews with a 480-pixel long side, with full-resolution images accessible.

Direct specifies bimanual position, orientation, and gripper targets. Bounded damped least-squares inverse kinematics~\cite{wampler1986ik} converts them to joint commands, with 5\,cm translation and 0.35\,rad rotation safeguards. Segments last one to five 25-Hz control steps. Hybrid exposes a $50\times14$ task-finetuned \pifive proposal through forward kinematics. Astra accepts a prefix of one to fifteen steps or replaces it with a one-to-five-step correction. Both use \texttt{xhigh} reasoning and the same native goals and safeguards. Many tasks require both arms home.

RoboLab uses a single-arm Franka and the same division between direct targets and policy assistance. Its \pifive weights are DROID-trained~\cite{khazatsky2024droid}, without finetuning on demonstrations from the test tasks. Appendix~\ref{app:manipulation} gives the evaluation procedure and decision budgets.

\subsection{Humanoid Interfaces}

The HumanoidBench interface uses Humanoid-GPT whole-body controllers~\cite{qi2026humanoidgpt}. It supplies body-frame forward/lateral velocity and yaw rate to a frozen walking policy, or pelvis-height and wrist offsets to a frozen sparse five-point tracker. Feet and keypoint orientations keep nominal references. Walking and tracking policies operate at 50\,Hz over 500-Hz native physics and joint proportional--derivative control. IK and PD implement the commanded motion.

Sparse targets are coupled whole-body references, re-entering sparse mode resets the anchor, and walking changes arm posture. Observations include privileged state and task geometry. The task suite also includes task-specific access such as bounded finger closure for Door.

Run, Crawl, and Stair use continuous five-point gait generators. Astra chooses speed, heading, and observation interval; calibrated generators supply posture, phase, and limb references. Stair additionally uses measured support heights and lateral foot-placement feedback. Walk combines cadence and leg-extension calibration with startup guidance. Push uses staging, contact guidance, and bounded hand control. The dense ScaleTrack locomotion study uses a separate robot--tracker protocol.

\subsection{Experience and Evaluation Scope}

Completion and timing conventions follow Section~\ref{sec:method}; replay verification is documented in Appendix~\ref{app:evidence}. The interfaces explicitly supply the state variables listed above to Astra.

Astra's weights remain fixed; guidance may include calibrated examples, feedback rules, or motion clips.

%% file: sec/X_suppl.tex
\section{Detailed Manipulation Results}
\label{app:manipulation}

\subsection{RoboDojo}

Table~\ref{tab:dojo} expands the aggregate results in Section~\ref{sec:robodojo}. Direct Score averages the 48 episodes with valid Scores; its success rate uses all 50 instances. Published policy scores are reweighted to the same task and scene mixture. Direct and Hybrid share scene configurations and seeds.

\begin{table}[htb]
\caption{\textbf{RoboDojo results.} Each architecture has five evaluation instances per task. Direct and Hybrid are paired on scene configuration and recorded seeds.}
\label{tab:dojo}

\centering\small
\setlength{\tabcolsep}{18.08pt}
\begin{tabular}{lrrrr}
\toprule
\rowcolor{tablehead}
&  \multicolumn{2}{c}{\reporthead{Native Score}}  &  \multicolumn{2}{c}{\reporthead{Success rate (\%)}} \\
\cmidrule(lr){2-3}\cmidrule(lr){4-5}
\rowcolor{tablehead}
\reporthead{Task} & \reporthead{Direct} & \reporthead{Hybrid} & \reporthead{Direct} & \reporthead{Hybrid}\\
\midrule
\rowcolor{white}
\input{tables/robodojo-tasks.tex}
\midrule
Aggregate & 37.81 & 62.60 & 26 & 48 \\
\bottomrule
\end{tabular}

\end{table}

\subsection{RoboLab}

\begin{table}[htb]
\caption{\textbf{RoboLab task success.} Each method contributes five outcomes per task. Cosmos denotes Cosmos3-Nano-Policy~\cite{nvidia2026cosmos3policy}; DreamZero follows~\citet{ye2026dreamzero}.}
\label{tab:robolab}

\centering\small
\setlength{\tabcolsep}{14.59pt}
\begin{tabular}{lrrrrr}
\toprule
\rowcolor{tablehead}
\reporthead{Task} & \reporthead{Direct} & \reporthead{Hybrid} & \reporthead{\pifive} & \reporthead{Cosmos} & \reporthead{DreamZero}\\
\midrule
\rowcolor{white}
\input{tables/robolab-tasks.tex}
\midrule
Total & 49/50 & 46/50 & 18/50 & 18/50 & 17/50 \\
\bottomrule
\end{tabular}

\end{table}

RoboLab permits repeat attempts for Astra. Two Direct Blocks-in-Bin attempts use a 500-decision budget; the standard budget is 180. Simulation horizons and native success criteria remain fixed. Each baseline contributes its first five runs per task, ordered by run, episode, and environment ID, with the initial states, task versions, and control settings of those runs.

\section{Focused Experiments, Resource Use, and Reproducibility}
\label{app:evidence}

\subsection{Focused Push and Walk Records}

These focused studies use separately frozen configurations. The table distinguishes development seeds from transfer seeds fixed in advance; these runs are separate from the task-suite results in Section~\ref{sec:humanoid}.

\begin{table}[htb]
\caption{\textbf{Focused online Astra runs under separately frozen configurations.} Push uses calibrated examples, with seeds 3 and 5 reproducing supplied physical trajectories; its physical outcome gives the final box-to-goal error and native success. Walk uses a gait frequency calibrated offline and a supplied heading-feedback rule. These outcomes are not substituted into the 30-task suite.}

\centering\small
\setlength{\tabcolsep}{14.01pt}
\begin{tabular}{llrrrr}
\toprule
\rowcolor{tablehead}
\reporthead{Configuration} & \reporthead{Seed role} & \reporthead{Seed} & \reporthead{Native return} & \reporthead{Controls} & \reporthead{Physical outcome}\\
\midrule
\rowcolor{white}
Focused Push & Development & 0 & 872.58 & 396 & 4.66\,cm; success \\
\rowcolor{tablestripe}
Focused Push & Development & 3 & 866.75 & 423 & 4.78\,cm; success \\
\rowcolor{white}
Focused Push & Development & 5 & 848.01 & 457 & 4.85\,cm; success \\
\rowcolor{tablestripe}
Focused Push & Transfer & 7 & 900.37 & 323 & 4.81\,cm; success \\
\midrule
\rowcolor{white}
Walk, 1.8\,Hz & Development & 1 & 743.54 & 1000 & No fall \\
\rowcolor{tablestripe}
Walk, 1.8\,Hz & Development & 2 & 748.92 & 1000 & No fall \\
\rowcolor{white}
Walk, 1.8\,Hz & Transfer & 8 & 743.18 & 1000 & No fall \\
\rowcolor{tablestripe}
Walk, 1.8\,Hz & Transfer & 9 & 748.54 & 1000 & No fall \\
\bottomrule
\end{tabular}

\end{table}

A separate scripted Push diagnostic tests whether contact must be calibrated jointly with the state at which the robot enters the contact phase. On seed 5, the final box-to-goal error is 27.02\,cm with unchanged entry and references, 36.81\,cm with changed entry only, 23.11\,cm with changed contact references only, and 4.97\,cm when both are changed. Heading and velocity also vary slightly between conditions.

\subsection{Independent Reach Records}

Two Reach runs from a separate Walk/Reach/Door batch measure target dwell time. Seeds 0 and 2 finish all 1000 controls with native returns of 15675.19 and 15380.91, respectively, both above the 12000 threshold. The hand is within the 5\,cm target radius for 11.36\,s and 12.78\,s. These records use separate archived instructions and execution settings; they are not part of the task suite's Reach sample.

\subsection{RoboDojo Execution and Token Accounting}

Table~\ref{tab:robodojo-resources} provides exact counts for Section~\ref{sec:resources}. Simulated duration excludes inference latency and can be shortened by early failure.

\begin{table}[htb]
\caption{\textbf{RoboDojo action and token accounting.} Mean simulated duration is the total control-step count multiplied by 0.04\,s and divided by 50 slots. Token totals include cached input and exclude earlier attempts.}
\label{tab:robodojo-resources}

\centering\small
\setlength{\tabcolsep}{21.25pt}
\begin{tabular}{lrr}
\toprule
\rowcolor{tablehead}
\reporthead{Metric} & \reporthead{Hybrid} & \reporthead{Direct}\\
\midrule
\rowcolor{white}
Executed control steps & 42,750 & 38,221 \\
\rowcolor{tablestripe}
Executed action segments & 3,776 & 7,729 \\
\rowcolor{white}
Mean simulated duration per slot (s) & 34.20 & 30.58 \\
\rowcolor{tablesummary}
Total tokens, including cached input & 624,762,828 & 1,132,343,772 \\
\rowcolor{white}
Cached input tokens & 607,555,840 & 1,107,349,760 \\
\rowcolor{tablestripe}
Uncached input tokens & 15,901,963 & 22,907,448 \\
\rowcolor{white}
Output tokens & 1,305,025 & 2,086,564 \\
\bottomrule
\end{tabular}

\end{table}

\subsection{Source Snapshots and Verification}

The audit reports 270 checked hashes across 90 online HumanoidBench episodes and checks on ten further runs. Reproduction requires the archived manifests, prompts, execution traces, and robot runtime. Replay verifies recorded dynamics and rewards.

\section{Manipulation Tasks and Behavioral Cases}
\label{app:visual-cases}

\subsection{Task Coverage}

\begin{figure}[H]
\centering
\begin{minipage}{.19\textwidth}\centering
\includegraphics[width=\linewidth]{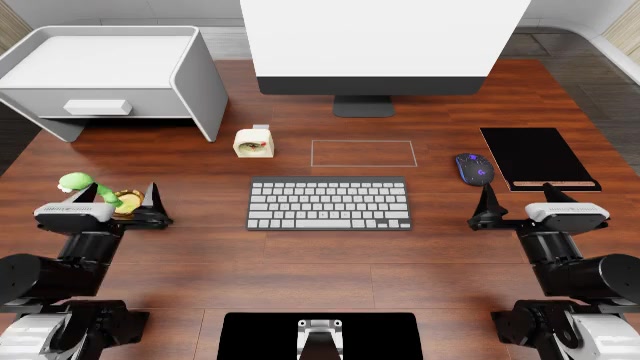}\\
{\scriptsize Table organization}\end{minipage}
\hfill
\begin{minipage}{.19\textwidth}\centering
\includegraphics[width=\linewidth]{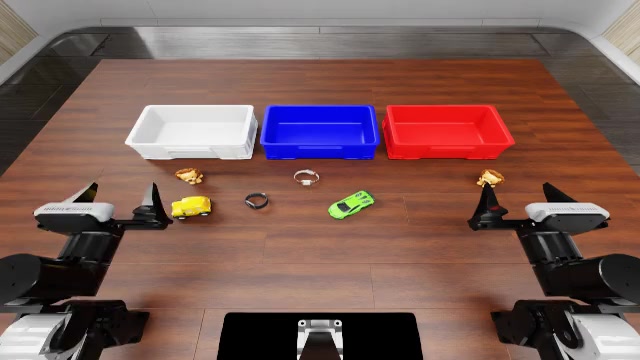}\\
{\scriptsize Language classification}\end{minipage}
\hfill
\begin{minipage}{.19\textwidth}\centering
\includegraphics[width=\linewidth]{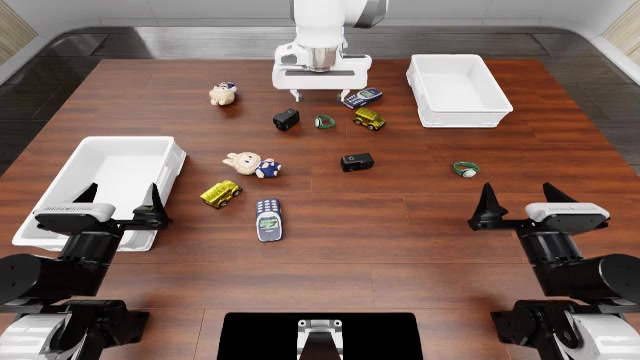}\\
{\scriptsize Sequence imitation}\end{minipage}
\hfill
\begin{minipage}{.19\textwidth}\centering
\includegraphics[width=\linewidth]{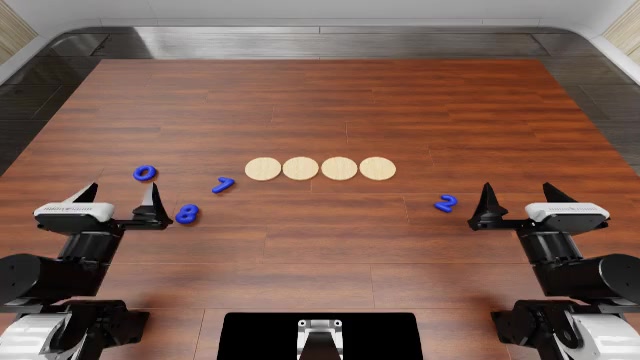}\\
{\scriptsize Number arrangement}\end{minipage}
\hfill
\begin{minipage}{.19\textwidth}\centering
\includegraphics[width=\linewidth]{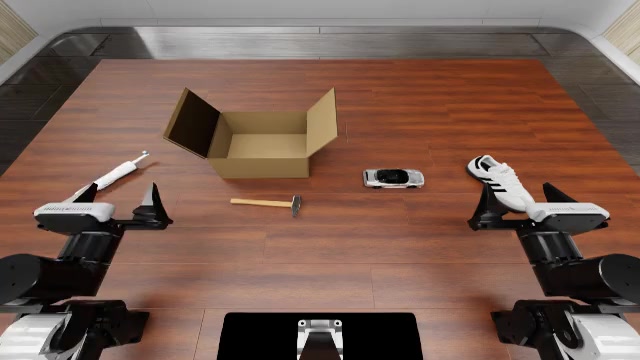}\\
{\scriptsize Packing}\end{minipage}
\par\medskip
\begin{minipage}{.19\textwidth}\centering
\includegraphics[width=\linewidth]{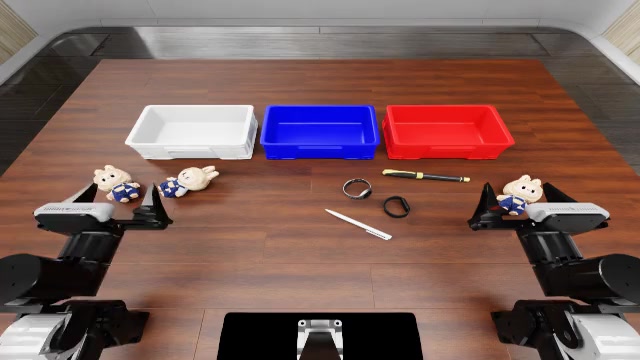}\\
{\scriptsize Object classification}\end{minipage}
\hfill
\begin{minipage}{.19\textwidth}\centering
\includegraphics[width=\linewidth]{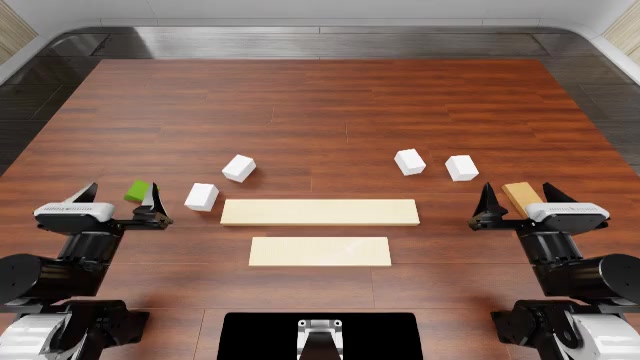}\\
{\scriptsize Tower construction}\end{minipage}
\hfill
\begin{minipage}{.19\textwidth}\centering
\includegraphics[width=\linewidth]{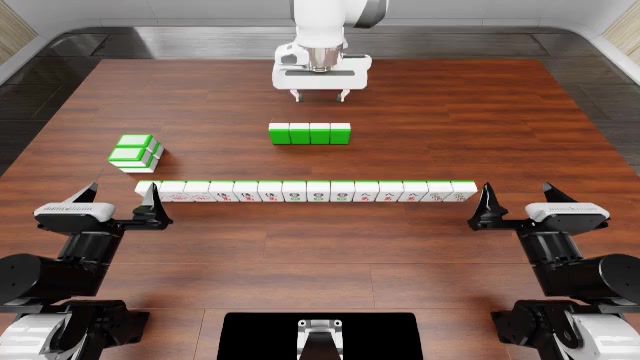}\\
{\scriptsize Mahjong}\end{minipage}
\hfill
\begin{minipage}{.19\textwidth}\centering
\includegraphics[width=\linewidth]{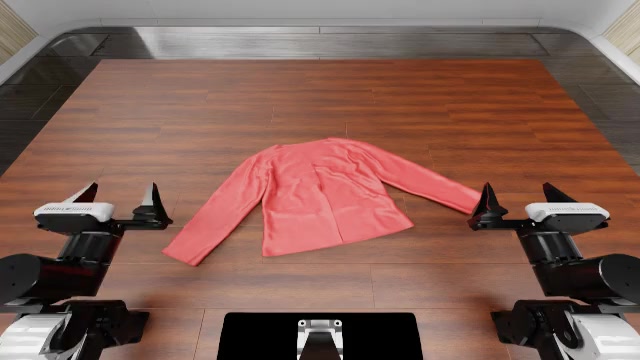}\\
{\scriptsize Clothes folding}\end{minipage}
\hfill
\begin{minipage}{.19\textwidth}\centering
\includegraphics[width=\linewidth]{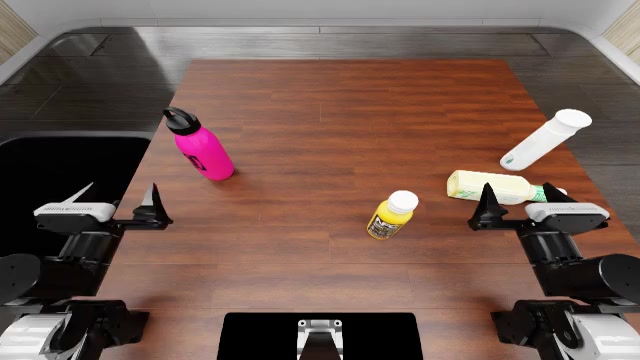}\\
{\scriptsize Bottle disposal}\end{minipage}
\caption{\textbf{The ten RoboDojo manipulation tasks.} Task images from our evaluation website are reproduced without alteration. They illustrate the environments rather than evaluation outcomes; the Direct and Hybrid results in Table~\ref{tab:dojo-public} use the paired episodes.}
\label{fig:dojo-scenes}
\end{figure}

\subsection{Behavioral Case Index}
The 21 recorded cases cover five aspects of manipulation: cases 1--4 concern goal alignment; 5--8 contact preparation; 9--12 terminal-state checking; 13--16 Hybrid contact and feedback failures; and 17--21 Direct strategies and local recovery. The cases illustrate behaviors observed in the recorded trajectories. Both Direct and Hybrid receive robot state, and the interpretations refer to complete trajectories rather than still images.

%% file: tables/robodojo-tasks.tex
\rowcolor{white}
Organize Table & 30.0 & 60.0 & 0 & 0 \\
\rowcolor{tablestripe}
Classify Objects By Language & 60.0 & 38.0 & 40 & 20 \\
\rowcolor{white}
Imitate Sorting Sequence & 0.0 & 53.0 & 0 & 40 \\
\rowcolor{tablestripe}
Arrange Largest Number & 57.0 & 50.0 & 40 & 40 \\
\rowcolor{white}
Pack Objects Into Box & 50.0 & 50.0 & 20 & 20 \\
\rowcolor{tablestripe}
Classify Objects & 100.0 & 71.0 & 100 & 60 \\
\rowcolor{white}
Build Tower & 12.0 & 64.0 & 0 & 60 \\
\rowcolor{tablestripe}
Make Kong & 0.0 & 40.0 & 0 & 40 \\
\rowcolor{white}
Fold Clothes & 40.0 & 100.0 & 40 & 100 \\
\rowcolor{tablestripe}
Put Bottles Into Dustbin & 36.0 & 100.0 & 20 & 100 \\

%% file: tables/robolab-tasks.tex
\rowcolor{white}
Blocks into bin & 5/5 & 5/5 & 0/5 & 2/5 & 0/5 \\
\rowcolor{tablestripe}
Pumpkins in clutter & 5/5 & 4/5 & 0/5 & 0/5 & 0/5 \\
\rowcolor{white}
Butter on raisin box & 5/5 & 5/5 & 0/5 & 1/5 & 2/5 \\
\rowcolor{tablestripe}
Stack blocks in order & 5/5 & 4/5 & 0/5 & 0/5 & 0/5 \\
\rowcolor{white}
Reorient red mug & 5/5 & 4/5 & 2/5 & 1/5 & 1/5 \\
\rowcolor{tablestripe}
Larger raisin box into bin & 4/5 & 4/5 & 3/5 & 0/5 & 0/5 \\
\rowcolor{white}
Sauce bottle into crate & 5/5 & 5/5 & 2/5 & 5/5 & 5/5 \\
\rowcolor{tablestripe}
Canned food into bin & 5/5 & 5/5 & 4/5 & 4/5 & 4/5 \\
\rowcolor{white}
Yogurt into bowl & 5/5 & 5/5 & 2/5 & 0/5 & 0/5 \\
\rowcolor{tablestripe}
Rubik's cube into bowl & 5/5 & 5/5 & 5/5 & 5/5 & 5/5 \\

%% file: sec/Z_humanoid_suppl.tex
\section{Humanoid Loco-Manipulation: Supporting Results}
\label{app:g1}

Table~\ref{tab:g1-all} expands the HumanoidBench study in Section~\ref{sec:humanoid}, reporting Astra returns across all 30 tasks. Task-specific return thresholds and observed physical outcomes are reported separately. The separate SIMPLE development records follow below.

Figure~\ref{fig:humanoid} reproduces published returns from HumanoidBench Table V~\cite{sferrazza2024humanoidbench}: DreamerV3~\cite{hafner2023dreamerv3} and SAC~\cite{haarnoja2018sac} use 10M training steps, and TD-MPC2~\cite{hansen2023tdmpc2} uses 2M. We use the average returns in Table V, reported over three seeds, rather than the maximum returns in Table VI. The comparison counts a task only when Astra's reported mean is strictly higher than the reference value; Kitchen is a tie at zero.

\begin{table}[htb]
\caption{\textbf{All 30 evaluated HumanoidBench tasks.} Entries are means $\pm$ sample SD over two seeds. Bold values meet or exceed the listed return threshold.}
\label{tab:g1-all}

\centering\small
\setlength{\tabcolsep}{40pt}
\begin{tabular}{lrr}
\toprule
\rowcolor{tablehead}
\reporthead{Task} & \reporthead{Threshold} & \reporthead{Astra}\\
\midrule
\input{tables/humanoid-all.tex}
\bottomrule
\end{tabular}

\end{table}

\Needspace*{5\baselineskip}
\subsection{Additional Physical Diagnostics}

\paragraph{Walk.}
The four 1.8\,Hz runs are listed in Appendix~\ref{app:evidence}. The task-suite results use posture calibration together with startup and speed guidance.

\paragraph{Run and Crawl.}
Run finishes 1,000 controls without falls, with mean forward center-of-mass speeds of 3.85 and 3.83\,m/s and displacements of 77.05 and 76.50\,m. Crawl maintains mean forward speed near 1.37\,m/s. Both Crawl runs stay within the lateral band, and neither records head height below 0.2\,m. A standing-based diagnostic flags the intended pitched posture; native task terms and progress are more informative.

\paragraph{Stair.}
Seed 1 reaches root X of 1.55\,m before falling at control 970; seed 2 reaches 1.06\,m, retreats, and finishes behind its starting X position.

\paragraph{Push.}
The Push runs are described in Section~\ref{sec:focused}; Appendix~\ref{app:evidence} gives the separate focused runs and exact entry/contact diagnostic outcomes.

\Needspace*{5\baselineskip}
\subsection{SIMPLE Development and Retention Checks}
\label{app:simple}
\paragraph{Control and observations.}
The SIMPLE implementation uses G1 with Dex3 hands, MuJoCo physics and native scoring, and Isaac Sim rendering. Astra runs with \texttt{xhigh} reasoning and commands both wrists and both ankles, together with independent finger actions. Frozen ScaleBFM executes the whole-body targets. Astra receives external, head, and wrist RGB, joint encoders, forward kinematics, and a base-state estimate. The depth-assisted variants also use relative depth inferred from RGB. Object ground truth, simulator contact diagnostics, and scoring signals are excluded from both action selection and reflection.

\paragraph{Experience representation.}
Two text documents support adaptation: \emph{Skill} contains task instructions, and \emph{Memory} contains cross-episode notes. Both are separate from shared control instructions and the robot interface. After each episode, Astra uses its observations and actions to revise these documents. Astra and ScaleBFM remain frozen.

\paragraph{Single-scene development.}
Across ten tasks, 58 attempts are recorded, including interruptions. The task-development endpoints meet native success on nine tasks; eight also meet execution, audit, and additional hold requirements. Handover triggers native success but times out before the hold check. Basket carrying receives a favorable qualitative assessment of its action sequence but fails native scoring. Neither is counted as a complete success. These outcomes summarize development rather than 58 independent trials.

\paragraph{Separate Handover studies.}
Table~\ref{tab:simple-development} separates grasp stabilization, fallen-carton recovery, and instruction adaptation. The stabilization variants pass native success and the additional hold check in 2/5 and 3/5 trials, respectively. The latter changes both instructions and relative-depth input. Recovery starts with the carton already fallen and allows a longer budget; its results are not pooled with upright-carton trials. In the retention check, one fixed procedure fails on scene00 and passes both criteria on scene02 and scene08. The instruction-adaptation trials use changing guidance.

\begin{table}[htb]
\caption{\textbf{SIMPLE Handover development records.} Native success and success with the additional hold check have identical counts in these trials and are reported together. Rows describe separate development conditions and are excluded from the L2 total in Table~\ref{tab:simple-l2}.}
\label{tab:simple-development}
\centering\small
\setlength{\tabcolsep}{7pt}
\begin{tabular}{>{\raggedright\arraybackslash}p{\dimexpr.27\textwidth-2\tabcolsep\relax}>{\raggedright\arraybackslash}p{\dimexpr.49\textwidth-2\tabcolsep\relax}>{\centering\arraybackslash}p{\dimexpr.11\textwidth-2\tabcolsep\relax}>{\centering\arraybackslash}p{\dimexpr.13\textwidth-2\tabcolsep\relax}}
\toprule
\rowcolor{tablehead}
\reporthead{Stage} & \reporthead{Condition} & \reporthead{Valid trials} & \reporthead{Success + hold}\\
\midrule
\rowcolor{white}
Grasp stabilization & Grasp entry, enclosure, and support transfer & 5 & 2/5 \\
\rowcolor{tablestripe}
Depth-assisted grasp & Relative depth and side-entry grasp guidance & 5 & 3/5 \\
\rowcolor{white}
Fallen-carton recovery & Table-edge support, regrasping, and handover & 1 & 0/1 \\
\rowcolor{tablestripe}
Depth-assisted recovery & Relative depth and recovery notes & 2 & 0/2 \\
\midrule
\rowcolor{white}
Trajectory-informed guidance & Scene 1; instructions derived from contrasting trajectories & 1 & 1/1 \\
\rowcolor{tablestripe}
Post-trial reflection & Scene 1; guidance informed by the preceding trial & 1 & 0/1 \\
\rowcolor{white}
Difficult-scene check & Scene 9; guidance informed by trial feedback & 1 & 0/1 \\
\rowcolor{tablestripe}
Frozen retention check & Scenes 00, 02, 08; one fixed candidate & 3 & 2/3 \\
\bottomrule
\end{tabular}
\end{table}

\paragraph{L2 evaluation protocol.}
We follow the official SIMPLE codebase and evaluation rules, using the same L2 protocol as the benchmark baselines. The evaluation comprises six tasks with ten scenes each; 50 of 60 trials meet the official success criteria, yielding an overall success rate of 83.3\%. The separate Handover development and retention studies are excluded from this total.

%% file: tables/humanoid-all.tex
\rowcolor{white}
Stand & 800 & \textbf{954.2} $\pm$ 0.1 \\
\rowcolor{tablestripe}
Walk & 700 & \textbf{848.7} $\pm$ 5.9 \\
\rowcolor{white}
Run & 700 & 642.1 $\pm$ 2.8 \\
\rowcolor{tablestripe}
Kitchen & 4 & 0.0 $\pm$ 0.0 \\
\rowcolor{white}
Maze & 1200 & \textbf{1358.8} $\pm$ 7.4 \\
\rowcolor{tablestripe}
Hurdle & 700 & 183.0 $\pm$ 0.3 \\
\rowcolor{white}
Cube & 370 & 6.3 $\pm$ 0.3 \\
\rowcolor{tablestripe}
Bookshelf Simple & 2000 & 666.3 $\pm$ 7.6 \\
\rowcolor{white}
Bookshelf Hard & 2000 & 572.8 $\pm$ 201.0 \\
\rowcolor{tablestripe}
Crawl & 700 & \textbf{971.9} $\pm$ 0.2 \\
\rowcolor{white}
Window & 650 & 6.1 $\pm$ 0.7 \\
\rowcolor{tablestripe}
Spoon & 650 & 319.1 $\pm$ 8.6 \\
\rowcolor{white}
Door & 600 & 142.4 $\pm$ 7.8 \\
\rowcolor{tablestripe}
Push & 700 & \textbf{877.3} $\pm$ 59.6 \\
\rowcolor{white}
Reach & 12000 & 11430.2 $\pm$ 2089.0 \\
\rowcolor{tablestripe}
Basketball & 1200 & 24.0 $\pm$ 2.9 \\
\rowcolor{white}
Truck & 3000 & 787.3 $\pm$ 87.1 \\
\rowcolor{tablestripe}
Package & 1500 & -5148.0 $\pm$ 1288.1 \\
\rowcolor{white}
Cabinet & 2500 & 274.1 $\pm$ 110.0 \\
\rowcolor{tablestripe}
Sit Simple & 750 & \textbf{793.0} $\pm$ 19.1 \\
\rowcolor{white}
Sit Hard & 750 & 277.5 $\pm$ 283.5 \\
\rowcolor{tablestripe}
Balance Simple & 800 & 53.1 $\pm$ 6.7 \\
\rowcolor{white}
Balance Hard & 800 & 48.3 $\pm$ 1.8 \\
\rowcolor{tablestripe}
Stair & 700 & 272.5 $\pm$ 44.0 \\
\rowcolor{white}
Slide & 700 & 141.1 $\pm$ 45.9 \\
\rowcolor{tablestripe}
Pole & 700 & 628.7 $\pm$ 10.5 \\
\rowcolor{white}
Room & 400 & 143.9 $\pm$ 27.6 \\
\rowcolor{tablestripe}
Insert Normal & 350 & 50.6 $\pm$ 1.5 \\
\rowcolor{white}
Insert Small & 350 & 46.7 $\pm$ 8.8 \\
\rowcolor{tablestripe}
Powerlift & 800 & 84.6 $\pm$ 53.7 \\

%% file: sec/Y_harness_suppl.tex
\section{Harness Development and Repeatability}
\label{app:harness}
\label{sec:harness_study}

These development cases supplement Section~\ref{sec:focused}. Each retains its own configuration and reasoning setting; none belongs to the frozen evaluation suite.

\subsection{Contact-Conditioned Guidance}
Contact-conditioned Push guidance reduces the prompt from 129 to 90 lines, replacing unconditional replay of a calibrated example with checks on contact, measured box displacement, and remaining error. Controller code, observations, action limits, native rules, and the 1.0\,s walking stop delay remain unchanged. On development seed 3, return changes from $-148.56$ to 863.90 and final box error from 29.45 to 4.88\,cm, with success after 428 controls. Seed 0 also succeeds; the additional seed 5 fails at 500 controls with 27.02\,cm error.

The archived replay reports record no state, reward, or native-information discrepancies. Each condition has one online attempt on known development seeds, with the prompt changes described above.

A separate Push request selects ten controls while noting rising lateral box speed and the need to observe before overshoot. Its explicit body and wrist references show state-informed action timing. The record captures local anticipation from observed state changes.

\subsection{A Saved Diagnosis and a Failed Repeat}
After a seed-2 Push failure, an online tool event records Astra writing \texttt{NOTES.md}: repeated lifting leaves the hand below the tabletop, without useful pushing contact. The subsequent guidance calls for outward lifting on the robot side and checks clearance before extension.

This procedure succeeds once in 381 controls with return 803.86, then fails on a repeat after 500 controls with return $-259.83$. Task and composed-skill hashes match; full transmitted-prompt hashes differ because the workspace path changes. All three workers use \texttt{high} reasoning. The archive attributes the written diagnosis to Astra; it does not identify the author of the subsequent guidance change.

\subsection{Finger Control Does Not Solve Door}
Developers extend Door with bounded finger closure and observations of finger positions and contact geometry. One online Astra episode issues five full-closure commands and completes 1,000 controls, but returns 144.38 against a threshold of 600, with no passage and maximum door angle approximately 0.0195\,rad. Guidance, observations, and action access change together in this condition.

\subsection{Provenance and Evidence Availability}
The development archive is organized under \texttt{sources/harness-study/}, with provenance, hashes, source excerpts, prompt versions, and replay reports. Its collection-time commit, \texttt{db82157bf9a64143433ec75e824a8cb7f89e2b8f}, identifies the inspected checkout. Verification requirements are given in Appendix~\ref{app:evidence}.

%% file: sec/Z_dexterous_suppl.tex
\section{Dexterous Manipulation: Data, Interfaces, and Evaluation}
\label{app:dexterous}

This appendix specifies the task suite, demonstration data, control interfaces, and evaluation protocol for Section~\ref{sec:dexterous-manipulation}. Aggregate scores are reported in Table~\ref{tab:dexterous-mock-m1}.

\subsection{Task Suite and Demonstration Data}
\label{app:dexterous-data}

Table~\ref{tab:dexterous-task-protocol} lists instructions and simulation budgets for the ten tasks; Figure~\ref{fig:dexterous-task-scenes} shows their initial configurations.

\input{figures/dexterous-initial-states}
\input{tables/dexterous-task-protocol}

Demonstrations use the instructions in Table~\ref{tab:dexterous-task-protocol}; their counts and training configuration appear in Table~\ref{tab:dexterous-implementation}.

\paragraph{Initial configurations.}
The task configurations vary object locations and, where applicable, object yaw and initial hand posture. Initial hand postures are drawn from the first frames of the corresponding task's demonstrations. The evaluation consequently measures execution across specified scene configurations with the same task objectives and assets. The fixed evaluation subset is described in Section~\ref{app:dexterous-evaluation}.

\subsection{Multitask Policy Finetuning}
\label{app:dexterous-training}

Standalone \pifive and Hybrid share one policy finetuned on the pooled demonstrations from all ten tasks. Table~\ref{tab:dexterous-implementation} specifies initialization, AdamW optimization~\cite{loshchilov2017adamw}, and execution settings for this shared policy.

\Needspace*{5\baselineskip}
\subsection{Observations, Numerical Actions, and Execution}
\label{app:dexterous-interfaces}

\paragraph{Policy observations.}
We use Sharpa dexterous hands. The simulation uses MuJoCo~3.11.0~\cite{todorov2012mujoco}. The \pifive policy receives RGB images from the head and both wrist cameras together with the task instruction and a 62-dimensional state vector comprising the two wrist poses and finger-joint angles. Each image is resized from $640\times360$ to $224\times224$ with aspect-ratio-preserving padding.

\paragraph{Policy action representation.}
Each \pifive action comprises two wrist poses, each encoded by 3 position values and a 6D rotation representation~\cite{zhou2019rotation}, and 22 joint angles per hand. The resulting action dimension is $2\times(3+6)+2\times22=62$.

\paragraph{Execution and model budgets.}
The baseline discards each unexecuted prediction suffix before observing and replanning. Table~\ref{tab:dexterous-implementation} gives execution cadences, Hybrid corrections, and budget-exhaustion behavior. Task limits in Table~\ref{tab:dexterous-task-protocol} count simulated time; model calls and inference wall time are separate.

\input{tables/dexterous-implementation}

\Needspace*{5\baselineskip}
\subsection{Evaluation Protocol and Scoring}
\label{app:dexterous-evaluation}
\label{app:dexterous-results}

\paragraph{Fixed evaluation cases.}
We evaluate each task on five fixed cases, yielding 50 evaluation cases across the ten tasks. All three methods are evaluated on the same case set.

\paragraph{Continuous execution and verification.}
Each task has a simulation-time limit $t_{\max}$, chosen to cover at least 1.5 times the longest demonstration and twice the median demonstration duration. Verification is performed during continuous execution. For the nine placement and rearrangement tasks, full completion requires all task subgoals to hold, the manipulated objects to be released, and their positions to remain stable for 2\,s. Release is verified from contact data showing zero contacts between an object and the hand collision geometries. Pot lifting instead requires the pot to remain at least 10\,cm above its initial height, in contact with a hand and out of contact with the table, for a 3\,s stable-holding window.

The positional stability threshold is 1\,cm over the verification window. The documented window definition measures each object's maximum root-position displacement from the start of that window; it is not a sum of frame-to-frame displacements. An interrupted verification condition restarts the window. Successful verification permits early completion; otherwise the episode is scored at termination. There is no separate stage in which the robot is frozen to validate an earlier placement.

\paragraph{Task subgoals.}
Table~\ref{tab:dexterous-scoring-rules} defines each predicate, including geometric tolerances, release requirements, and dependencies between subgoals.

\input{tables/dexterous-scoring-rules}

\paragraph{Partial-credit scores.}
Let $K_t$ be the number of subgoals for task $t$, and let $k$ be the number satisfied at termination. The documented score tiers use an internal scale $s\in[0,1]$:
\begin{itemize}[leftmargin=1.5em,itemsep=2pt,topsep=3pt]
    \item $s=1$ when all subgoals and the full continuous verification window are satisfied;
    \item $s=0.8$ when all subgoals hold at termination but full verification has not been completed;
    \item $s=0.8k/K_t$ for partial completion, where $0<k<K_t$;
    \item $s=0.1$ when no subgoal holds at termination but a manipulated object was lifted by 5\,cm during execution;
    \item $s=0$ for no effective progress or an applicable invalidating termination.
\end{itemize}
Thus, completing one of two subgoals ordinarily gives 40 points on the reported scale, whereas satisfying all subgoals without verification gives 80 points. Partial credit reflects the terminal configuration, not accumulated subgoals. The 5\,cm lift event separately records earlier progress.

\paragraph{Safety and termination.}
The task configurations specify a 200\,N hand--table threshold, a 60\,N wrist-camera contact threshold, and self-collision checks. Model-call exhaustion must be distinguished from task timeout and physical termination.

\paragraph{Aggregation.}
For method $m$, let $s_{t,i,m}$ denote the internal score on case $i$ of task $t$. Task and overall scores are
\begin{equation}
S_{t,m}=\frac{100}{5}\sum_{i=1}^{5}s_{t,i,m},
\qquad
S_m=\frac{1}{10}\sum_{t=1}^{10}S_{t,m}.
\label{eq:dexterous-mean-score}
\end{equation}
Each task contributes equally to the 0--100 partial-credit score reported in Table~\ref{tab:dexterous-mock-m1}; this is not a success rate.

%% file: figures/dexterous-initial-states.tex
\begin{figure}[!htbp]
\centering
\begin{minipage}[t]{.24\textwidth}\vspace{0pt}\centering
\includegraphics[width=\linewidth]{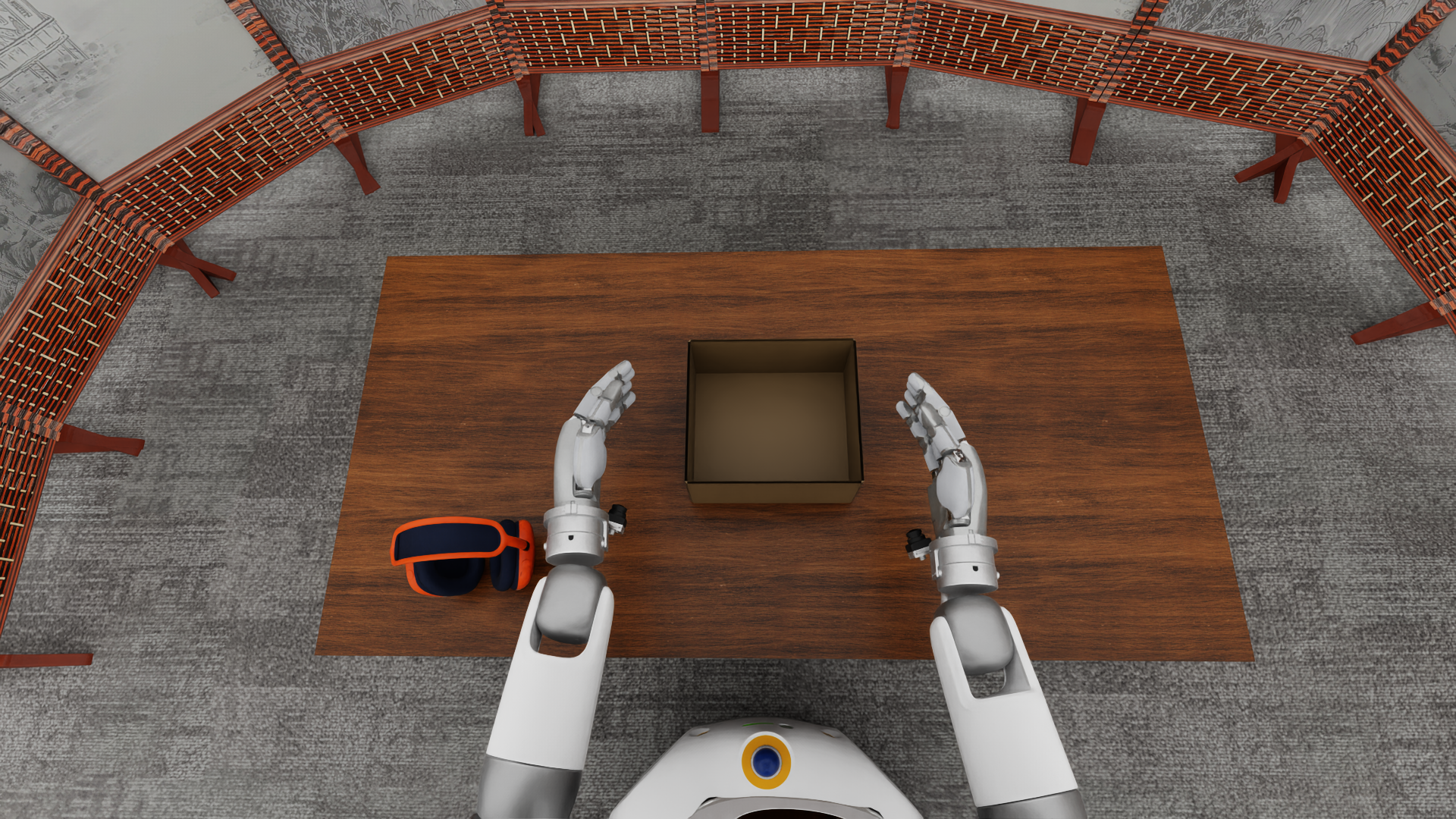}\\[3pt]
{\small (a) Headphones in box}\end{minipage}
\hfill
\begin{minipage}[t]{.24\textwidth}\vspace{0pt}\centering
\includegraphics[width=\linewidth]{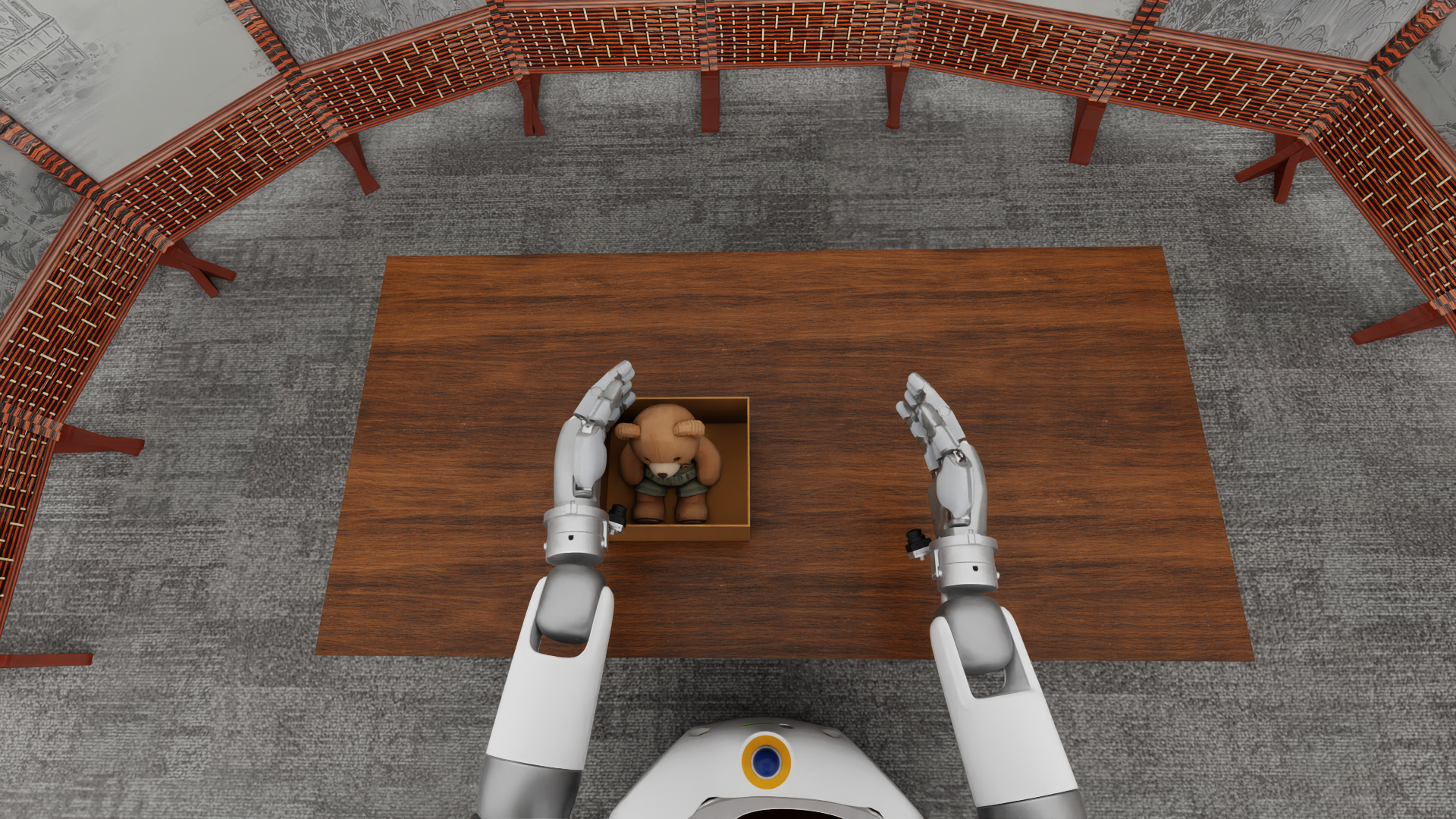}\\[3pt]
{\small (b) Toy retrieval}\end{minipage}
\hfill
\begin{minipage}[t]{.24\textwidth}\vspace{0pt}\centering
\includegraphics[width=\linewidth]{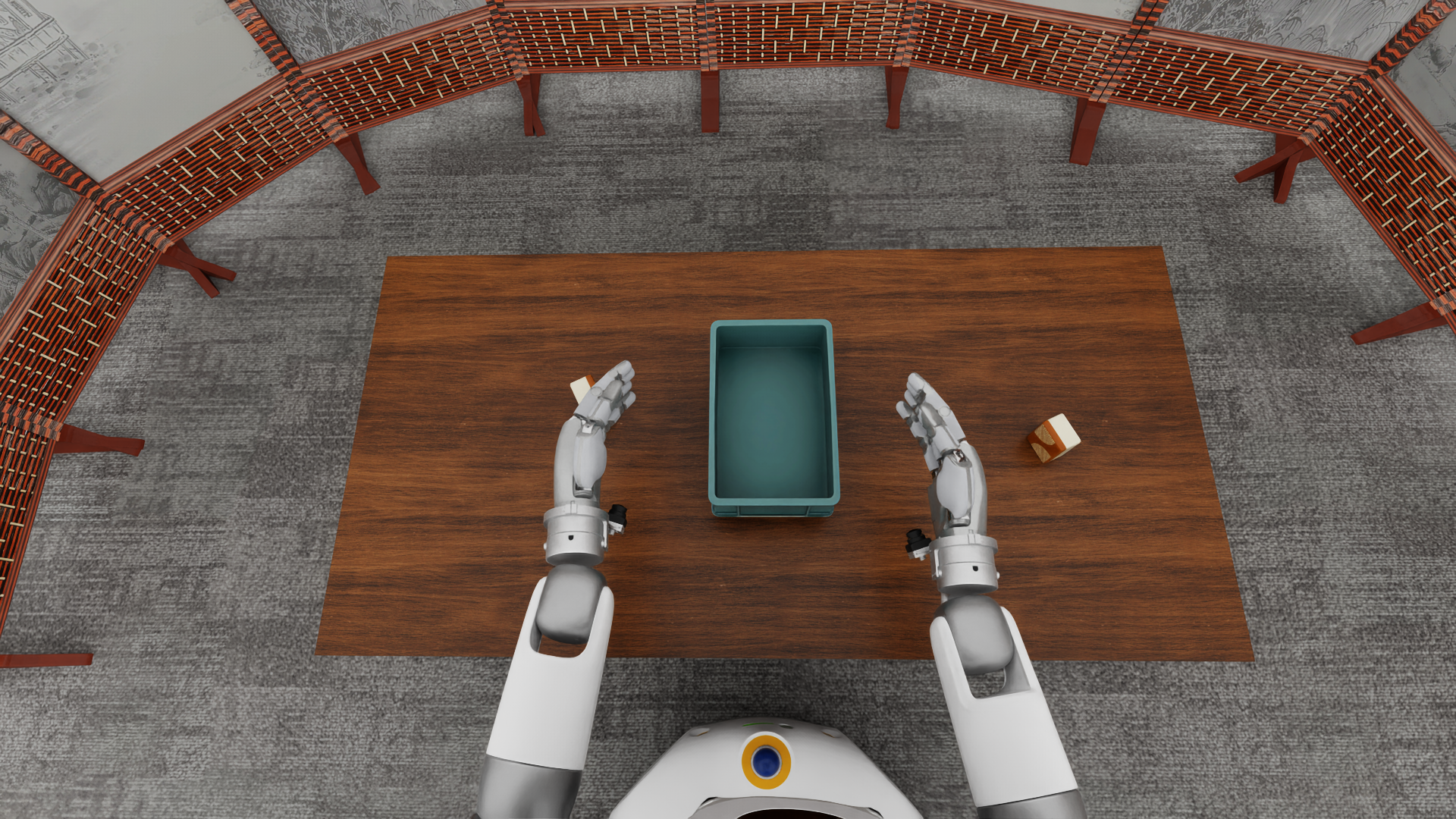}\\[3pt]
{\small (c) Mahjong placement}\end{minipage}
\hfill
\begin{minipage}[t]{.24\textwidth}\vspace{0pt}\centering
\includegraphics[width=\linewidth]{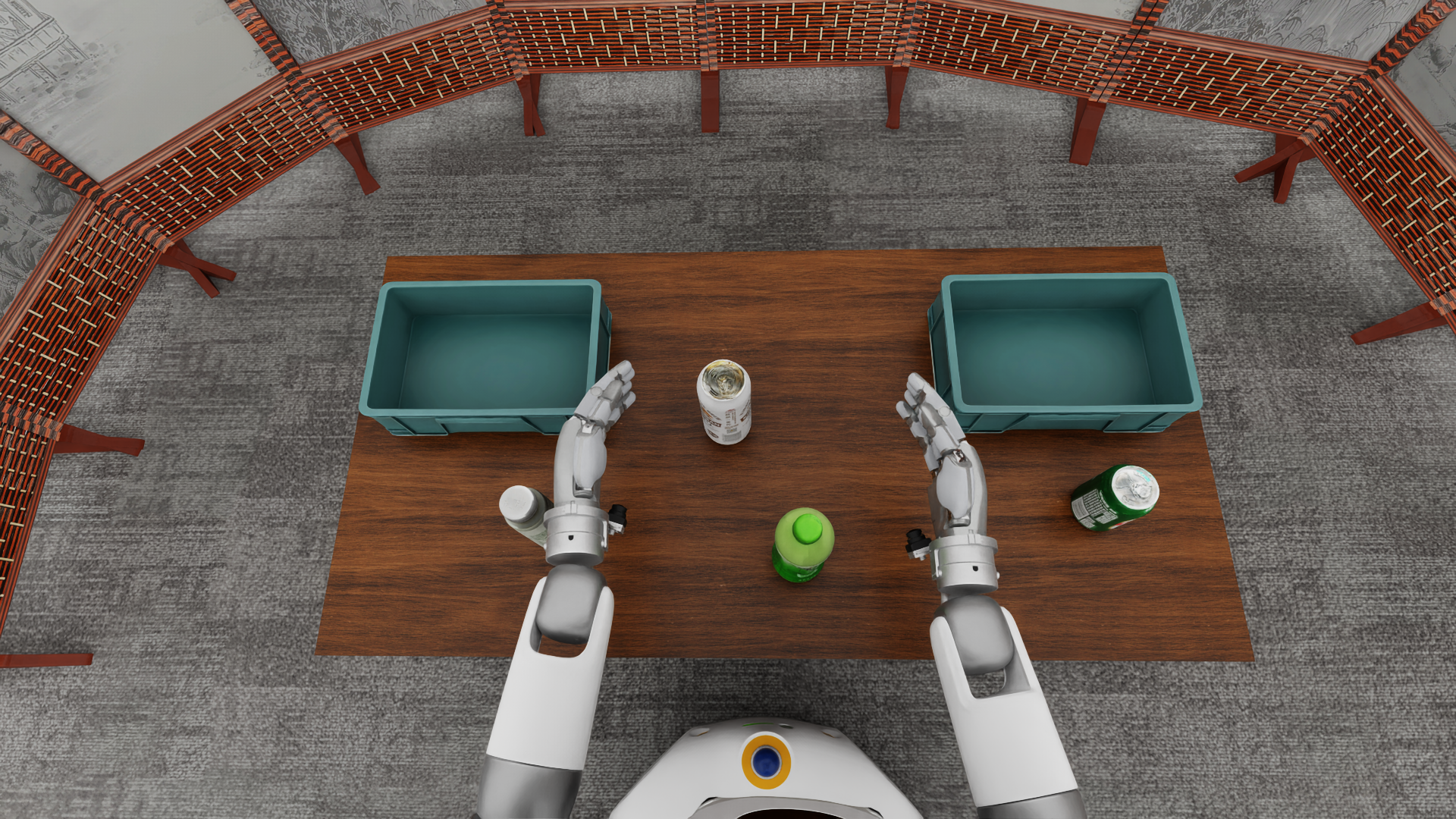}\\[3pt]
{\small (d) Bottle/can sorting}\end{minipage}
\par\medskip
\begin{minipage}[t]{.24\textwidth}\vspace{0pt}\centering
\includegraphics[width=\linewidth]{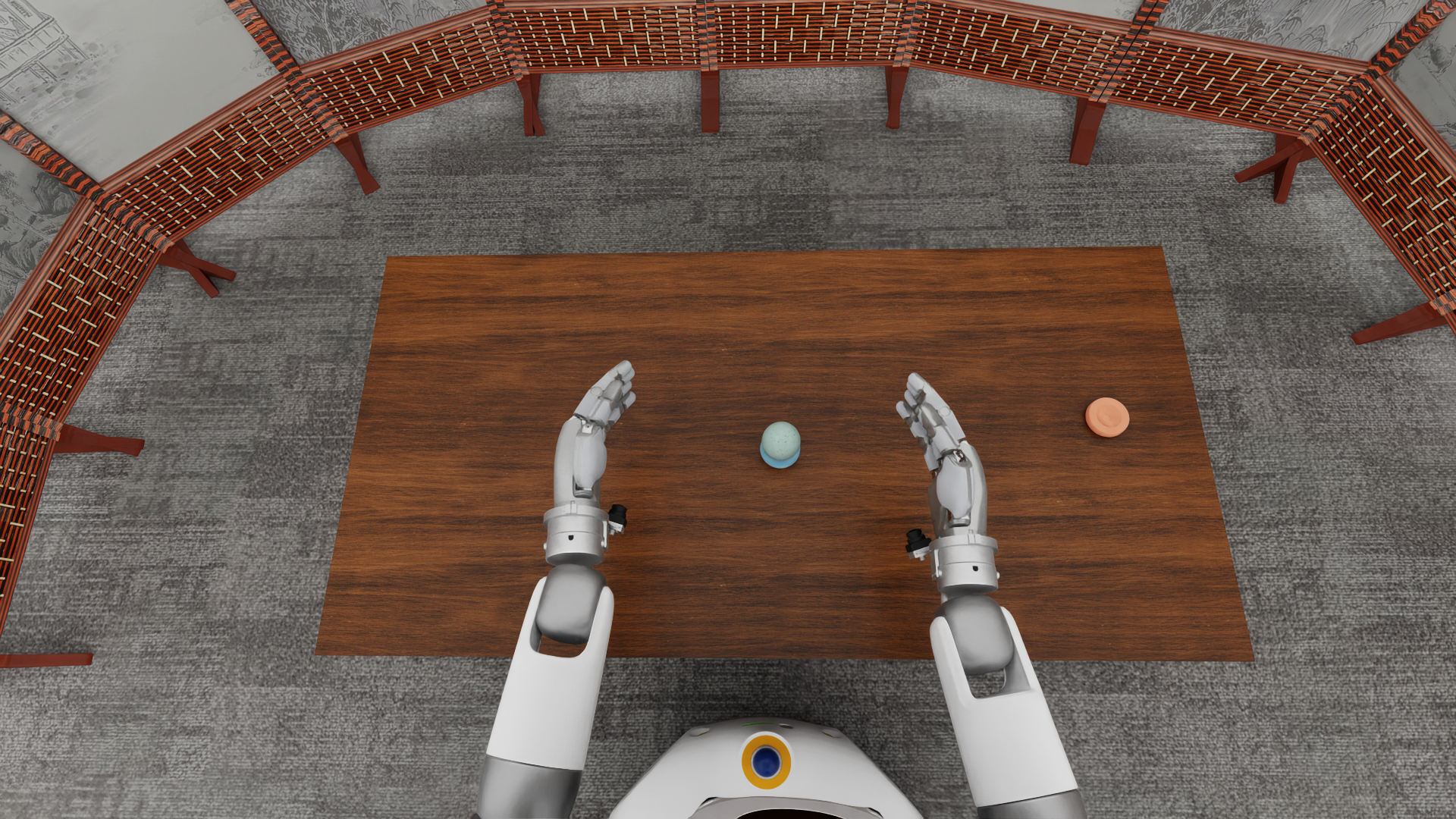}\\[3pt]
{\small (e) Upright egg placement}\end{minipage}
\hfill
\begin{minipage}[t]{.24\textwidth}\vspace{0pt}\centering
\includegraphics[width=\linewidth]{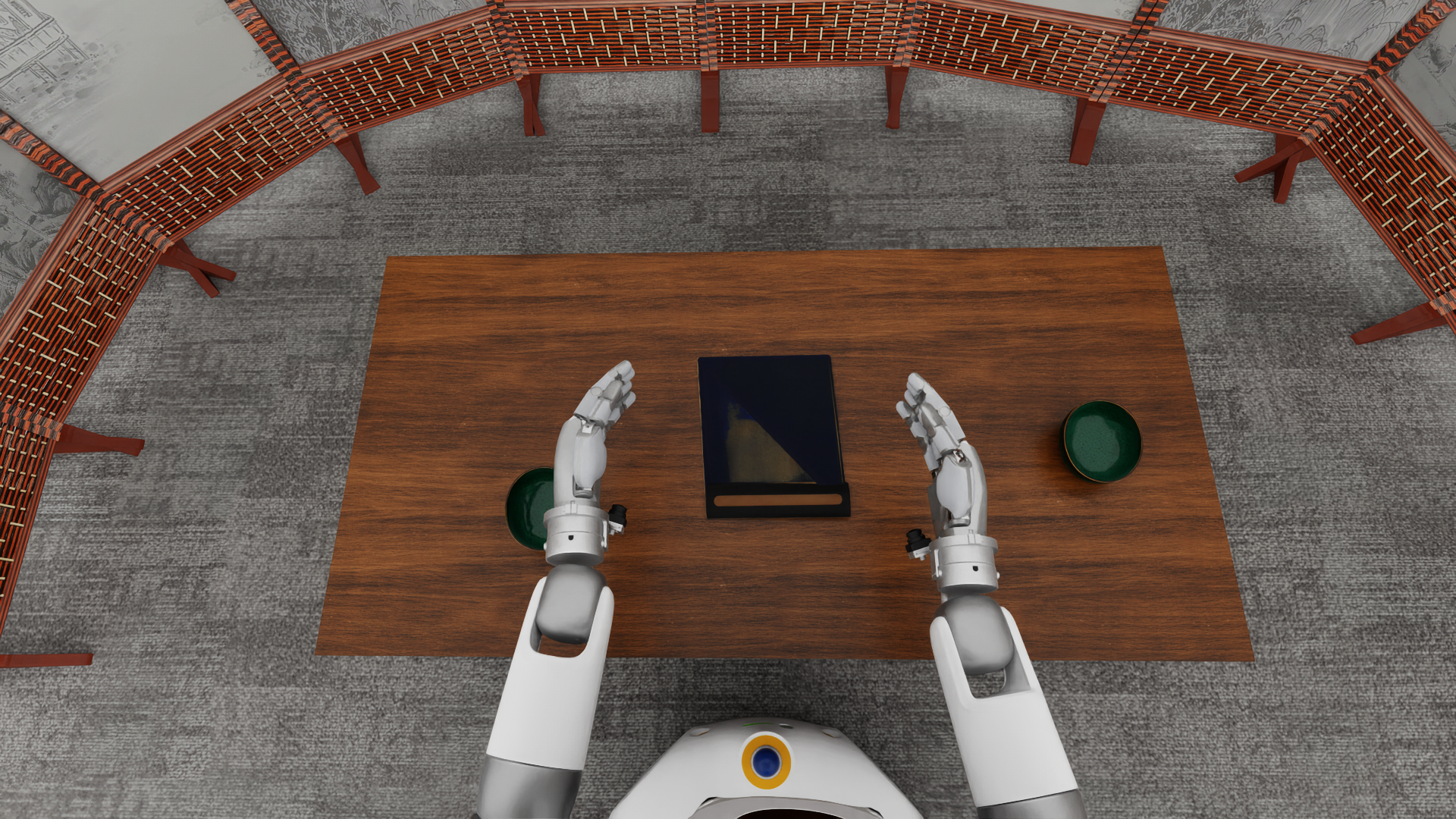}\\[3pt]
{\small (f) Two-bowl stacking}\end{minipage}
\hfill
\begin{minipage}[t]{.24\textwidth}\vspace{0pt}\centering
\includegraphics[width=\linewidth]{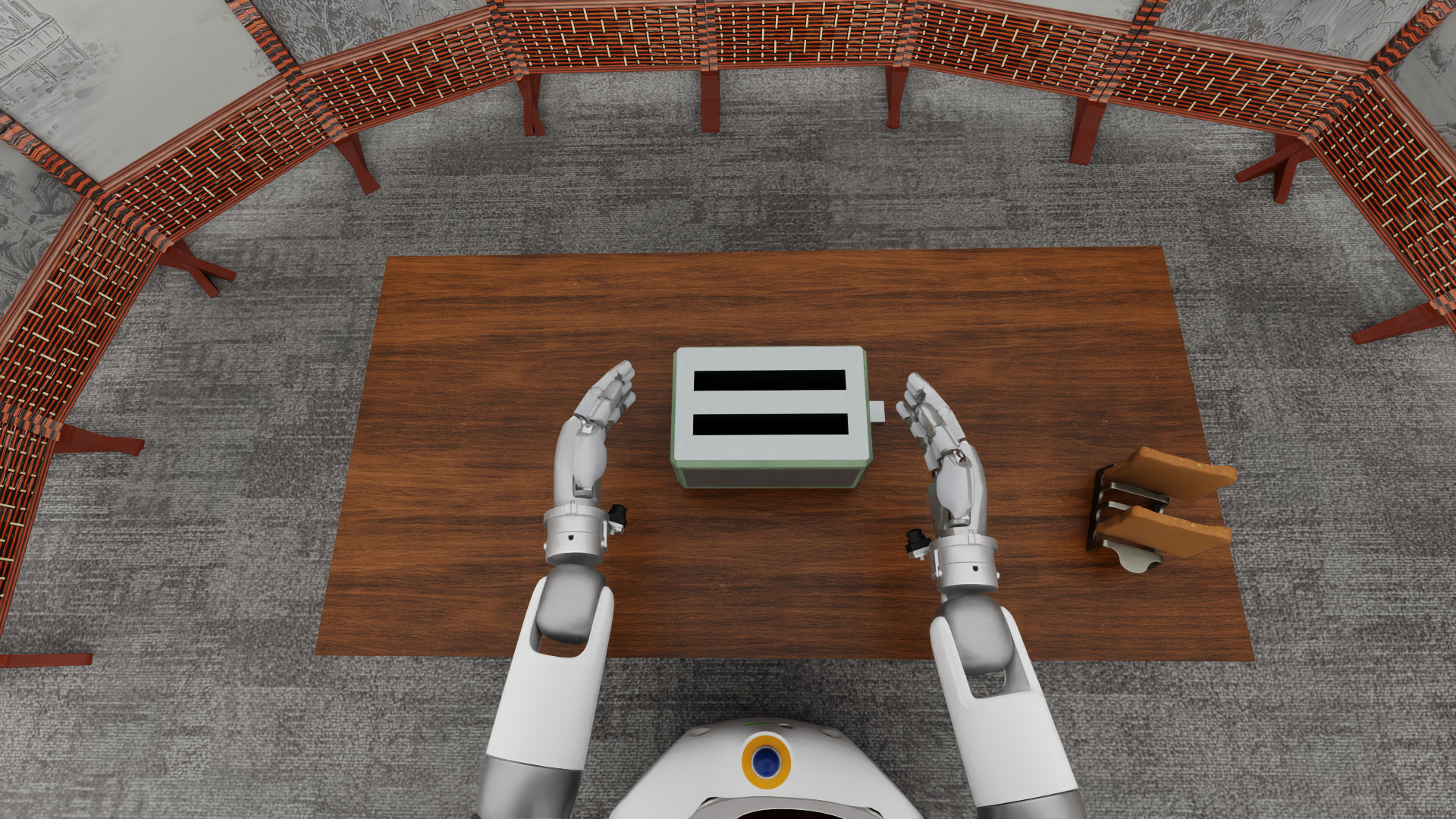}\\[3pt]
{\small (g) Bread insertion}\end{minipage}
\hfill
\begin{minipage}[t]{.24\textwidth}\vspace{0pt}\centering
\includegraphics[width=\linewidth]{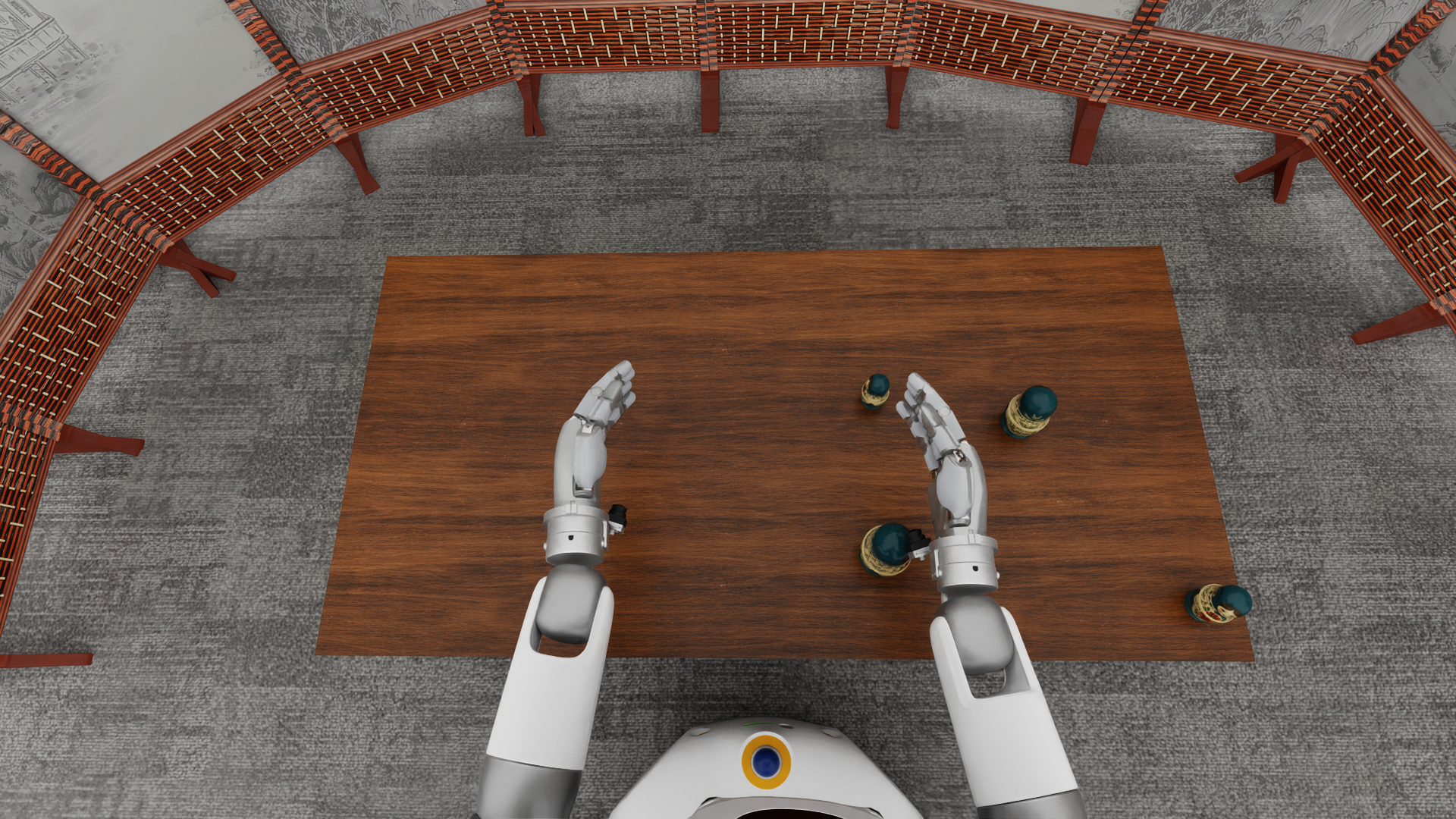}\\[3pt]
{\small (h) Nesting-doll ordering}\end{minipage}
\caption{\textbf{Example initial states for dexterous manipulation.} Panels show eight benchmark tasks, with one initial state each.}
\label{fig:dexterous-task-scenes}
\end{figure}

%% file: tables/dexterous-task-protocol.tex
\begin{table*}[!htbp]
\caption{\textbf{Dexterous manipulation tasks and evaluation horizons.}
Task instructions reproduce the prompts used for training and evaluation.
$K$ is the number of subgoals used for partial-credit scoring.
$t_{\max}$ and the continuous verification window are measured in simulation time.
Subgoal predicates are specified in Table~\ref{tab:dexterous-scoring-rules}.}
\label{tab:dexterous-task-protocol}

\centering
\small
\setlength{\tabcolsep}{5.00pt}
\begin{tabular}{>{\raggedright\arraybackslash}p{\dimexpr0.200\textwidth-2\tabcolsep\relax}>{\raggedright\arraybackslash}p{\dimexpr0.525\textwidth-2\tabcolsep\relax}>{\centering\arraybackslash}p{\dimexpr0.055\textwidth-2\tabcolsep\relax}>{\centering\arraybackslash}p{\dimexpr0.085\textwidth-2\tabcolsep\relax}>{\centering\arraybackslash}p{\dimexpr0.135\textwidth-2\tabcolsep\relax}}
\toprule
\rowcolor{tablehead}
\reporthead{Task} & \reporthead{Task instruction} & \reporthead{$K$} & \reporthead{\makecell{$t_{\max}$\\(s)}} & \reporthead{\makecell{Verification\\(s)}}\\
\midrule
\rowcolor{white}
Pot lift and hold & Lift the pot at least 10 centimeters above the table and hold it steadily without dropping it. & 1 & 30 & 3 \\
\addlinespace[3pt]
\rowcolor{tablestripe}
Headphones in box & Pick the headset and place it into the box. & 1 & 20 & 2 \\
\addlinespace[3pt]
\rowcolor{white}
Toy retrieval & Take the toy out of the bin and place it on the table. & 2 & 20 & 2 \\
\addlinespace[3pt]
\rowcolor{tablestripe}
Mug hanging & Hang the mug on the mug rack\newline without requiring handle-specific alignment. & 1 & 25 & 2 \\
\addlinespace[3pt]
\rowcolor{white}
Two mahjong tiles & Place both mahjongs into the basket. & 2 & 30 & 2 \\
\addlinespace[3pt]
\rowcolor{tablestripe}
Bottles/cans sorting & Place bottles into the front-left basket\newline and cans into the front-right basket. & 4 & 60 & 2 \\
\addlinespace[3pt]
\rowcolor{white}
Upright egg placement & Move the egg from the blue platform to the orange platform and leave it upright in the holder. & 2 & 40 & 2 \\
\addlinespace[3pt]
\rowcolor{tablestripe}
Two-bowl stacking & Stack the two bowls nested upright on the black mousemat. & 2 & 30 & 2 \\
\addlinespace[3pt]
\rowcolor{white}
Bread-slot insertion & Place both bread slices into the toaster,\newline one slice in each slot. & 2 & 45 & 2 \\
\addlinespace[3pt]
\rowcolor{tablestripe}
Nesting-doll ordering & Arrange the four nesting dolls in a row from left to right, from smallest to largest. & 2 & 60 & 2 \\
\bottomrule
\end{tabular}

\end{table*}

%% file: tables/dexterous-implementation.tex
\begin{table}[!htbp]
\caption{\textbf{Dexterous manipulation implementation details.}}
\label{tab:dexterous-implementation}

\centering
\small
\setlength{\tabcolsep}{5.00pt}
\begin{tabular}{>{\raggedright\arraybackslash}p{\dimexpr0.270\textwidth-2\tabcolsep\relax}>{\raggedright\arraybackslash}p{\dimexpr0.730\textwidth-2\tabcolsep\relax}}
\toprule
\rowcolor{tablehead}
\reporthead{Configuration} & \reporthead{Specification}\\
\midrule
\rowcolor{white}
Demonstration budget & 100 trajectories per task; ten tasks; 1,000 trajectories in total. \\
\rowcolor{tablestripe}
S1 adaptation & One multi-task \pifive policy, shared by the standalone baseline and Hybrid. \\
\rowcolor{white}
Direct action responsibility & Astra specifies wrist/arm targets and finger-joint targets. \\
\rowcolor{tablestripe}
Hybrid action responsibility & \pifive supplies manipulation actions; Astra can edit wrist and finger targets. \\
\midrule
\rowcolor{white}
S1 initialization and training & \pifive base initialization; full-parameter finetuning; 50,000 optimization steps; global batch size 256. \\
\rowcolor{tablestripe}
S1 optimizer & AdamW with $\beta=(0.9,0.95)$, $\epsilon=10^{-8}$, weight decay $10^{-10}$, and gradient-norm clipping at 1.0. \\
\rowcolor{white}
S1 learning rate & Peak $2.5\times10^{-5}$; cosine decay to $2.5\times10^{-6}$. \\
\rowcolor{tablestripe}
S1 numerical actions & $50\times62$ output; world-frame wrist poses with 6D rotations and absolute finger-joint targets; wrist increments restored to absolute poses before execution. \\
\rowcolor{white}
\pifive baseline execution & 30\,Hz; execute the first 16 steps of each 50-step prediction before replanning; physics paused during inference. \\
\midrule
\rowcolor{tablestripe}
Hybrid observations and actions & Astra receives the task instruction, synchronized head and wrist RGB views, measured wrist poses, 22 joint angles per hand, hand keypoints, and the remaining simulation budget. A persistent episode conversation retains observation and action history; each decision also includes the previous execution record, a 50-step \pifive wrist preview, and hand-motion summaries. Astra selects an unchanged \pifive action prefix or specifies bounded wrist and finger corrections, which are converted into 62-dimensional control targets. \\
\addlinespace[4pt]
\rowcolor{white}
Hybrid execution and prompting & Control runs at approximately 30\,Hz, with one reset-settling step; \pifive action prefixes execute for 1--16 steps and Astra corrections for 1--5 steps before re-observation. Astra uses \texttt{xhigh} reasoning, with prompts and model configuration archived per run. Takeover requires observed execution failure or misaligned \pifive intent; control returns to \pifive once recovery permits. Each run can set a maximum number of reviewed action chunks and a token budget. Under the default fallback, either limit ends Astra's involvement, and \pifive continues alone until the simulator terminates the episode. Simulation time pauses while Astra deliberates. \\
\bottomrule
\end{tabular}

\end{table}

%% file: tables/dexterous-scoring-rules.tex
\begin{table}[!htbp]
\caption{\textbf{Task-specific completion criteria.}
Positions refer to object roots unless stated otherwise.
Full credit additionally requires continuous stability within 1\,cm and the verification windows in Table~\ref{tab:dexterous-task-protocol}; all tasks require release except pot holding.
Headset and mahjong release conditions also apply to their subgoal predicates.}
\label{tab:dexterous-scoring-rules}

\centering
\small
\setlength{\tabcolsep}{5.00pt}
\begin{tabular}{>{\raggedright\arraybackslash}p{\dimexpr0.210\textwidth-2\tabcolsep\relax}>{\raggedright\arraybackslash}p{\dimexpr0.790\textwidth-2\tabcolsep\relax}}
\toprule
\rowcolor{tablehead}
\reporthead{Task} & \reporthead{Subgoal predicates and task-specific verification conditions}\\
\midrule
\rowcolor{white}
Pot lift and hold & The pot root is at least 10\,cm above its initial height. The 0.8 tier requires this lift condition at termination. Full-score verification requires continued hand contact, no pot--table contact, and stability for 3\,s. \\
\addlinespace[4pt]
\rowcolor{tablestripe}
Headphones in box & The root projects inside the box-bottom polygon; at least 50\% of the vertically constrained minimum bounding-box volume lies in the vertical prism extending upward from that polygon; and hand contact is absent. The prism has no upper height limit, so a released headset resting across the rim can qualify. \\
\addlinespace[4pt]
\rowcolor{white}
Toy retrieval & Two independent subgoals: (i) the root projects outside the box-bottom polygon; (ii) the toy contacts the table and its root height is at most the table height plus its reference resting height plus 3\,cm. Reference resting height is the initial root height above the box-bottom reference plane. \\
\addlinespace[4pt]
\rowcolor{tablestripe}
Mug hanging & Conjunctive subgoal: lift from the initial root height $>4.5$\,cm, root height above the table $>13.5$\,cm, and root-to-rack horizontal distance $<9$\,cm. Handle alignment is unnecessary. \\
\addlinespace[4pt]
\rowcolor{white}
Two mahjong tiles & One subgoal per tile: the root projects inside the basket interior (half-widths 10.85 and 6.67\,cm), and hand contact is absent. No constraint is imposed on the tiles\textquotesingle{} vertical positions. Full credit requires both released tiles to remain stable for 2\,s. \\
\addlinespace[4pt]
\rowcolor{tablestripe}
Bottles/cans sorting & One subgoal per object: the root projects inside its assigned basket interior and its height relative to the basket reference plane is in $[-6,10]$\,cm. Bottles belong in the front-left basket and cans in the front-right basket; placement in the other basket receives no credit. \\
\addlinespace[4pt]
\rowcolor{white}
Upright egg placement & Two nested subgoals: (i) the root is within 2\,cm horizontally and 1.5\,cm vertically of the target seat; (ii) condition (i) holds and the egg's local $z$ axis is within $30^\circ$ of world vertical. Upright orientation at the source provides no subgoal credit. \\
\addlinespace[4pt]
\rowcolor{tablestripe}
Two-bowl stacking & Two independent subgoals: (i) the bowls' horizontal separation is at most 2\,cm and vertical separation at least 0.5\,cm, each bowl tilts at most $45^\circ$, and the lower bowl at most $7^\circ$; (ii) the lower bowl's center lies inside the coaster-local rectangle $x\in[-12.005,12.005]$\,cm, $y\in[-9.601,9.601]$\,cm, with no edge tolerance. \\
\addlinespace[4pt]
\rowcolor{white}
Bread-slot insertion & One subgoal per distinct occupied slot: a bread root projects within the slot polygon, its height above the slot reference plane is in $[2,8]$\,cm, and at least one of its local $\pm x$ or $\pm y$ axes lies within $30^\circ$ of world upward. Two slices in one slot count once; either end may face upward. Full-score verification requires both slices to be released. \\
\addlinespace[4pt]
\rowcolor{tablestripe}
Nesting-doll ordering & Two subgoals: (i) the four roots span at most 6\,cm along world $x$; (ii) in increasing size order, every adjacent pair has a signed world-$y$ separation of at least 4\,cm. \\
\bottomrule
\end{tabular}

\end{table}

%% file: sec/Z_dexterous_suppl1.tex
\section{In-Hand Manipulation: Controllers and Complete Results}
\label{sec:supp-dexterous}

This appendix details the in-hand experiments in Section~\ref{sec:dexterous-manipulation}.

\subsection{Benchmark Definition}
\label{sec:supp-dexterous-benchmark}

\begin{figure*}[htb]
  \centering
  \includegraphics[width=\textwidth]{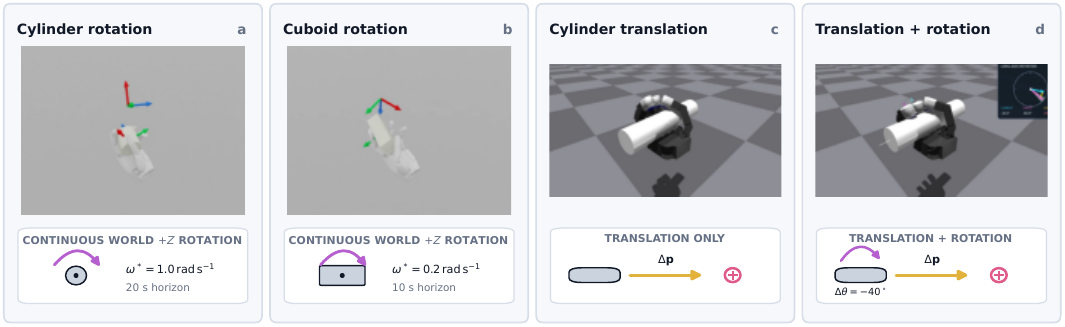}
  \caption{The four dexterous-manipulation tasks. (a--b) The Sharpa hand rotates a cylinder or cuboid continuously about world-frame $+Z$ at the stated target speed. (c) The Allegro Hand translates the cylinder center to a Cartesian target. (d) The second Allegro task additionally requires a $-40^\circ$ long-axis rotation. The images and lower schematics define the tasks rather than illustrate performance.}
  \label{fig:supp-dexterous-benchmark}
\end{figure*}

Sharpa Wave~\cite{sharpa_wave} rotation runs in Isaac Lab~\cite{nvidia2025isaaclab}, with task formulations drawn from ConTrack~\cite{liang2026contrack} and the \texttt{mjlab\_hand} codebase~\cite{qiao2026mjlabhand}, built on mjlab~\cite{zakka2026mjlab}. Allegro tasks use the Isaac Gym~\cite{makoviychuk2021isaacgym} tactile-IHT implementation: translation follows Yin et al.~\cite{yin2025learning}, and translation with long-axis rotation is our extension. Both simulators run at 20 control steps per simulated second. Table~\ref{tab:supp-dexterous-suite} and Figure~\ref{fig:supp-dexterous-benchmark} define the goals and endpoints.

\begin{table*}[htb]
\caption{\textbf{In-hand manipulation benchmark.} Each task contains five paired initial states evaluated by Astra Direct and the corresponding frozen RL policy.}
\label{tab:supp-dexterous-suite}

  \centering
  
  \small
  \setlength{\tabcolsep}{5.00pt}
  \begin{tabular}{>{\raggedright\arraybackslash}p{\dimexpr0.180\textwidth-2\tabcolsep\relax}>{\raggedright\arraybackslash}p{\dimexpr0.160\textwidth-2\tabcolsep\relax}>{\raggedright\arraybackslash}p{\dimexpr0.240\textwidth-2\tabcolsep\relax}>{\raggedright\arraybackslash}p{\dimexpr0.190\textwidth-2\tabcolsep\relax}>{\raggedright\arraybackslash}p{\dimexpr0.230\textwidth-2\tabcolsep\relax}}
    \toprule
    \rowcolor{tablehead}
\reporthead{Task} & \reporthead{Platform} & \reporthead{Goal} & \reporthead{Endpoint} & \reporthead{Reported measures}\\
    \midrule
    \rowcolor{white}
Cylinder rotation
      & Sharpa / Isaac Lab
      & Rotate about world $+Z$ at $1.0\,\mathrm{rad\,s^{-1}}$
      & 20 s (400 steps)
      & Orientation error, axial-speed MAE, At-goal, completion, drop \\
\rowcolor{tablestripe}
Cuboid rotation
      & Sharpa / Isaac Lab
      & Rotate about world $+Z$ at $0.2\,\mathrm{rad\,s^{-1}}$
      & 10 s (200 steps)
      & Orientation error, axial-speed MAE, At-goal, completion, drop \\
    \addlinespace[2pt]
    \rowcolor{white}
Cylinder translation
      & Allegro / Isaac Gym
      & Reach a Cartesian target
      & 15 s (300 steps)
      & Terminal position error, success, drop \\
\rowcolor{tablestripe}
Translation + rotation
      & Allegro / Isaac Gym
      & Reach the Cartesian target and rotate the long axis by $-40^\circ$
      & Astra at 120 decisions; RL at matched steps and 15 s
      & Terminal position and rotation errors, joint success, drop \\
    \bottomrule
  \end{tabular}
\end{table*}

\paragraph{Case selection and pairing.}
For every reported pair, the two controllers load the same saved hand joint state, object pose and velocity, and task target before control begins. Simulator randomization is disabled after loading. The five cylinder-rotation cases are drawn from the held-out grasp bank. The cuboid grasp source yielded only six validated stable starts; we fixed the first five case identifiers before running the formal Astra evaluation. Four of these starts overlap the RL initialization bank and one is held out. The two Allegro tasks use five distinct saved physical initial states each. All translation-and-rotation states come from the stabilized rerun; the earlier unstable pilot is excluded from every reported value.

\subsection{Task-Specific RL Controllers and Training}
\label{sec:supp-dexterous-rl-training}

Table~\ref{tab:supp-rl-provenance} summarizes the four frozen task-specific policies. Evaluation uses no adaptation or action tuning.

\begin{table*}[htb]
\caption{\textbf{Training provenance of the evaluated RL policies.} $N$ is the number of parallel training environments and $H$ is the rollout length per PPO update. The translation baseline uses the released checkpoint without retraining; training counts are listed for the three policies trained in this study.}
\label{tab:supp-rl-provenance}

  \centering
  
  \small
  \setlength{\tabcolsep}{5.00pt}
  \begin{tabular}{>{\raggedright\arraybackslash}p{\dimexpr0.230\textwidth-2\tabcolsep\relax}>{\raggedright\arraybackslash}p{\dimexpr0.380\textwidth-2\tabcolsep\relax}>{\raggedright\arraybackslash}p{\dimexpr0.390\textwidth-2\tabcolsep\relax}}
    \toprule
    \rowcolor{tablehead}
\reporthead{Task} & \reporthead{Policy construction} & \reporthead{Training setup}\\
    \midrule
    \rowcolor{white}
Cylinder rotation & PPO from scratch; four distributed ranks; speed curriculum & $N=4\!\times\!8192$, $H=16$, 1,000 updates; $5.24\!\times\!10^8$ transitions \\
\rowcolor{tablestripe}
Cuboid rotation & PPO initialized from the cylinder actor and actor normalizer & $N=8192$, $H=16$, 4,000 updates; $5.24\!\times\!10^8$ transitions \\
\rowcolor{white}
Cylinder translation & Released privileged-teacher/tactile-student policy & Released checkpoint; no retraining \\
\rowcolor{tablestripe}
Translation + rotation & Privileged PPO teacher initialized from the released oracle & $N=4096$, $H=16$, 20,000 updates; $1.31\!\times\!10^9$ transitions \\
    \bottomrule
  \end{tabular}
\end{table*}

\paragraph{Sharpa continuous-rotation policies.}
Both Sharpa policies use PPO~\cite{schulman2017ppo} in RSL-RL~\cite{schwarke2025rslrl} and a 22-dimensional action at 20 Hz. Their actor and critic are three-layer ELU MLPs with widths $(512,256,128)$ and separate running observation normalization. PPO uses discount $\gamma=0.99$, generalized advantage estimation~\cite{schulman2015gae} with $\lambda=0.95$, clipping parameter $0.2$, five optimization epochs, an adaptive learning rate initialized at $5\times10^{-4}$, target KL $0.01$, value coefficient $2.0$, zero entropy coefficient, and gradient-norm clipping at $1.0$. The policy observation contains finger joint positions and velocities, object pose and velocity relative to the palm, the moving rotation target, goal-relative orientation, and the previous action. Both tasks share the reward
\begin{equation}
  r_t = \Delta t\left[
    \frac{0.02}{e_t+0.1}
    + \mathbb{I}(e_t\leq 0.1)
    - 10^{-6}\lVert\mathbf{a}_t-\mathbf{a}_{t-1}\rVert_2^2
    + 0.02 r_t^{\mathrm{contact}}
  \right],
  \label{eq:supp-sharpa-reward}
\end{equation}
where $\Delta t=0.05\,\mathrm{s}$, $e_t=2\arccos\!\left(\left|\langle q_t,q_t^\star\rangle\right|\right)$ is the geodesic error between the object orientation and the moving target, and $r_t^{\mathrm{contact}}$ is the ContactExplorer novelty reward~\cite{liu2026contactexplorer} over fingertip proximity and contacted surface regions. Thus angular-speed tracking is encouraged indirectly by following the continuously advancing orientation target rather than by a separate velocity-reward term.

The cylinder policy is trained with seed 42 on a group-disjoint 80\% grasp-bank split. Training advances through target-speed stages of $0.25$, $0.5$, $0.75$, and $1.0\,\mathrm{rad\,s^{-1}}$, samples the training grasp bank with probability 0.5, and randomizes wrist position by $\pm5\,\mathrm{mm}$ and wrist orientation by $\pm5^\circ$. The policy is frozen after 1,000 updates and evaluated on five starts from the held-out 20\% grasp-bank split.

The cuboid policy is trained only on the cuboid training split. Its actor and actor-observation normalizer are initialized from the cylinder policy; the critic, critic normalizer, action variance, optimizer, update counter, and curriculum state are reinitialized. We evaluate the final seed-45 policy deterministically at $0.2\,\mathrm{rad\,s^{-1}}$. Table~\ref{tab:supp-cuboid-cases} marks each start's training-set overlap.

\paragraph{Allegro translation policy.}
The translation baseline is the unmodified released signed-three-axis tactile policy of Yin et al.~\cite{yin2025learning}, obtained from their public tactile-IHT codebase.\footnote{\url{https://github.com/jessicayin/tactile_skin_model}} Its source method first trains a privileged PPO teacher with simulator object and contact information. The teacher's environment reward is
\begin{equation}
\begin{aligned}
  \tilde r_t^{\mathrm{trans}}={}&-100e_{p,t}^2+10I_{p,t}
  -0.3D_{\mathrm{pose},t}-0.1D_{\tau,t}-2D_{\mathrm{work},t}\\
  &-1000D_{z,t}-170D_{\mathrm{tilt},t},
\end{aligned}
\label{eq:supp-allegro-translation-reward}
\end{equation}
where $e_{p,t}=\lVert\mathbf{p}_t-\mathbf{p}^\star\rVert_2$ and $I_{p,t}=\mathbb{I}(e_{p,t}^2<5\times10^{-4})$. The regularizers are $D_{\mathrm{pose},t}=\lVert\mathbf{q}_t-\mathbf{q}_0\rVert_2^2$, $D_{\tau,t}=\lVert\boldsymbol{\tau}_t\rVert_2^2$, $D_{\mathrm{work},t}=(\sum_i|\tau_{t,i}||\dot q_{t,i}|)^2$, $D_{z,t}=\operatorname{clip}(0.48-p_{z,t},0,1)$, and $D_{\mathrm{tilt},t}=(c_{1y}-c_{2y})^2+(c_{1z}-c_{2z})^2$, where $\mathbf{c}_1$ and $\mathbf{c}_2$ are the cylinder endpoints. Terms present in the released implementation with zero weight are omitted from Eq.~\ref{eq:supp-allegro-translation-reward}. The student is then trained by action imitation rather than by optimizing this reward directly. It uses a 30-step temporal Transformer to infer an eight-dimensional latent from joint history and signed three-axis tactile measurements, followed by an MLP with widths $(512,256,128)$ that outputs 16 joint actions. We evaluate the released policy without retraining.

\paragraph{Allegro translation-and-rotation policy.}
For the translation-and-rotation task, we train a privileged PPO teacher initialized from the released oracle policy. The actor MLP has widths $(512,256,128)$ and receives the 99-dimensional task observation together with an eight-dimensional embedding of privileged simulator state produced by a $(256,128,8)$ MLP. PPO uses $\gamma=0.99$, $\lambda=0.95$, learning rate $10^{-4}$, minibatches of 4,096 transitions, three epochs, clipping parameter $0.2$, critic coefficient $4$, zero entropy coefficient, and gradient-norm clipping at $1.0$. The target is fixed to $+0.045\,\mathrm{m}$ translation and $-40^\circ$ long-axis rotation. Starting from Eq.~\ref{eq:supp-allegro-translation-reward}, the task replaces the translation-only completion bonus and adds rotation terms:
\begin{equation}
\begin{aligned}
  \tilde r_t^{\mathrm{trans+rot}}={}&\tilde r_t^{\mathrm{trans}}-10I_{p,t}
  +10I_{p,t}I_{\theta,t}-2e_{\theta,t}^2\\
  &+15v_{\mathrm{toward},t}+10I_{\theta,t}.
\end{aligned}
\label{eq:supp-allegro-translation-rotation-reward}
\end{equation}
Here $e_{\theta,t}$ is the wrapped long-axis angle error, $I_{\theta,t}=\mathbb{I}(e_{\theta,t}<10^\circ)$, and $v_{\mathrm{toward},t}$ is the signed angular speed toward the target, clipped to $[-1,1]\,\mathrm{rad\,s^{-1}}$ and set to zero inside the rotation tolerance. The joint-success reward therefore applies only when both $I_{p,t}$ and $I_{\theta,t}$ are one. For both Allegro teachers, the PPO implementation multiplies the complete environment reward by $0.01$ before storing it for optimization. We evaluate the saved policy with the highest recorded training return. The reference policy uses privileged observations as specified above.

\Needspace*{5\baselineskip}
\subsection{Astra Inference Protocol}
\label{sec:supp-dexterous-astra-inference}

\paragraph{Model and prompt.}
Every direct-control rollout uses \texttt{gpt-6-astra} with \texttt{xhigh} reasoning through the Responses API. Temperature is not explicitly configured, provider fallback is disabled, and each model request has a 900-s controller timeout. The fixed developer prompt defines the robot, task, coordinate conventions, observation fields, action contract, and the requirement to preserve the grasp while making progress toward the object-level target. It contains no demonstrations or RL actions. The per-step packet explicitly states the numerical target: world-frame axis and angular speed for Sharpa, Cartesian goal for translation, and Cartesian plus signed long-axis goal for the translation-and-rotation task.

\paragraph{Observation and temporal context.}
At every decision, Astra receives three synchronized RGB views together with named joint positions and velocities, the current joint target, object pose and velocity, target state, and the host's contact representation. Sharpa supplies per-fingertip contact validity and target angular velocity in both palm and world coordinates. Allegro supplies 16 signed three-axis tactile readings, or 48 scalar values, as well as task-native position and rotation errors and threshold indicators. Images are resized to at most 480 pixels on the longest edge, except translation case 000, which uses the recorded 384-pixel setting. Reward terms and cumulative reward are logged by the host but never placed in the model observation. The runs maintain one model thread for the physical episode, so earlier decisions remain in the conversation context. Changing the model thread never resets or replays the simulator state.

\paragraph{Action execution.}
The model must answer with a schema-validated tool call containing the current request identifier, one action vector, an integer repeat count, and a short action rationale. On Sharpa, the vector is a 22-dimensional normalized joint-target increment in $[-0.1,0.1]$; increments accumulate and the absolute normalized target is clipped to $[-1,1]$ before the unchanged EMA/PD controller is applied. On Allegro, the vector has 16 entries in $[-1,1]$, and the native controller applies $q_{\mathrm{target}}\leftarrow q_{\mathrm{target}}+0.04167a$ followed by joint-limit clipping. The repeat count $k\in\{1,\ldots,5\}$ holds the requested command for $k$ consecutive 0.05-s control steps. The next observation is generated only after those steps have executed.

Each rollout ends when the object is dropped, the task horizon is reached, or Astra reaches its limit of 100 decisions on Sharpa and 120 decisions on Allegro. Because an Astra command may span one to five control steps, this decision limit can be reached before the task horizon; RL instead produces one action at every 20-Hz control step.

\subsection{Pairing and Evaluation Protocol}
\label{sec:supp-dexterous-controllers}

\paragraph{RL inference and paired starts.}
Each frozen RL policy receives its native normalized observation and produces a deterministic action at every control step. Evaluation disables observation corruption and does not update weights, normalization statistics, or curriculum state. The RL controller never receives Astra's action history. Pairing holds the saved physical start and goal fixed while allowing the two controllers to retain their intended observations, architectures, and inference schedules.

\paragraph{Execution endpoints.}
Horizons are specified in Table~\ref{tab:supp-dexterous-suite}. Early-ending trajectories contribute only executed steps, without padding. Translation-and-rotation results use three endpoints: Astra at its decision limit, RL after the same number of control steps, and RL at the full 300-step horizon. These separate performance over matched simulated durations from eventual completion; per-case lengths appear in Section~\ref{sec:supp-dexterous-cases}.

\Needspace*{5\baselineskip}
\subsection{Evaluation Metrics}
\label{sec:supp-dexterous-metrics}

\paragraph{In-hand rotation.}
Let $q_o(t)$ and $q_g(t)$ denote the object and moving target quaternions. The benchmark's orientation error is their geodesic distance,
\begin{equation}
  e_q(t)=2\arccos\!\left(\left|\langle q_o(t),q_g(t)\rangle\right|\right).
\end{equation}
At-goal is the fraction of evaluated control steps for which $e_q(t)<0.1\,\mathrm{rad}$. To measure sustained motion, consecutive object-quaternion increments are unwrapped, projected onto world-frame $+Z$, and averaged over a trailing one-second window. Speed MAE is the mean absolute difference between that estimate and the commanded angular speed. Orientation error, At-goal, and Speed MAE are arithmetic means over executed control steps. Full records whether the specified analysis horizon is reached, and Drop records object loss within that window.

As Section~\ref{sec:dexterous-manipulation} explains, periodic orientation agreement need not imply sustained rotation; $e_q(t)$ must be interpreted with axial speed. Raw cumulative reward is diagnostic and is not compared across tasks.

\paragraph{In-hand translation and reorientation.}
Cylinder translation is evaluated by the Euclidean distance between the cylinder center and its Cartesian target at the final control step. Terminal success requires a position error below $22.4\,\mathrm{mm}$. The translation-and-rotation task uses the same position threshold together with an absolute wrapped long-axis rotation error below $10^\circ$ relative to the requested $-40^\circ$ change. Joint success requires both conditions at the same terminal endpoint. 

\subsection{Complete Per-Case Results}
\label{sec:supp-dexterous-cases}

The tables contain all 40 rollouts from five paired starts per task. Rotation means pool executed steps; Allegro means average terminal values over cases. \emph{Decisions} counts accepted outputs and equals \emph{Steps} for RL.

``Budget'' in Table~\ref{tab:supp-cylinder-cases} denotes exhaustion of Astra's 100 decisions.

\begin{table*}[htb]
\caption{\textbf{Cylinder rotation over the 20-s analysis window.} All five saved initial states are held out from the RL grasp bank.}
\label{tab:supp-cylinder-cases}

  \centering
  
  \small
  \setlength{\tabcolsep}{11.40pt}
  \begin{tabular}{llcrrrrr}
    \toprule
    \rowcolor{tablehead}
\reporthead{Case} & \reporthead{Controller} & \reporthead{End} & \reporthead{Steps} & \reporthead{Decisions} & \reporthead{At-goal $\uparrow$} & \reporthead{Error (rad) $\downarrow$} & \reporthead{Speed MAE $\downarrow$}\\
    \midrule
    \rowcolor{white}
000 & Astra Direct & Budget  & 223 & 100 & 0.45\% & 1.756 & 0.977 \\
\rowcolor{white}
& RL           & Horizon & 400 & 400 & 73.25\% & 0.193 & 0.208 \\
\rowcolor{tablestripe}
004 & Astra Direct & Drop    & 326 & 84  & 0.31\% & 1.532 & 0.906 \\
\rowcolor{tablestripe}
& RL           & Horizon & 400 & 400 & 77.00\% & 0.182 & 0.335 \\
\rowcolor{white}
008 & Astra Direct & Horizon & 400 & 96  & 0.75\% & 1.610 & 0.957 \\
\rowcolor{white}
& RL           & Horizon & 400 & 400 & 75.50\% & 0.201 & 0.297 \\
\rowcolor{tablestripe}
012 & Astra Direct & Horizon & 400 & 100 & 0.50\% & 1.632 & 0.940 \\
\rowcolor{tablestripe}
& RL           & Horizon & 400 & 400 & 79.50\% & 0.138 & 0.349 \\
\rowcolor{white}
016 & Astra Direct & Horizon & 400 & 100 & 0.50\% & 1.638 & 1.017 \\
\rowcolor{white}
& RL           & Horizon & 400 & 400 & 79.25\% & 0.153 & 0.190 \\
    \bottomrule
  \end{tabular}
\end{table*}

\begin{table*}[htb]
\caption{\textbf{Cuboid rotation over the 10-s analysis window.} Cases 000--003 overlap the RL initialization set; 004 is held out.}
\label{tab:supp-cuboid-cases}

  \centering
  
  \small
  \setlength{\tabcolsep}{11.29pt}
  \begin{tabular}{llcrrrrr}
    \toprule
    \rowcolor{tablehead}
\reporthead{Case} & \reporthead{Controller} & \reporthead{Split} & \reporthead{Steps} & \reporthead{Decisions} & \reporthead{At-goal $\uparrow$} & \reporthead{Error (rad) $\downarrow$} & \reporthead{Speed MAE $\downarrow$}\\
    \midrule
    \rowcolor{white}
000 & Astra Direct & overlap  & 200 & 50  & 0.50\%  & 0.798 & 0.143 \\
\rowcolor{white}
& RL           & overlap  & 200 & 200 & 81.00\% & 0.071 & 0.053 \\
\rowcolor{tablestripe}
001 & Astra Direct & overlap  & 200 & 56  & 6.50\%  & 0.917 & 0.168 \\
\rowcolor{tablestripe}
& RL           & overlap  & 200 & 200 & 51.00\% & 0.123 & 0.067 \\
\rowcolor{white}
002 & Astra Direct & overlap  & 200 & 54  & 5.50\%  & 0.690 & 0.139 \\
\rowcolor{white}
& RL           & overlap  & 200 & 200 & 65.00\% & 0.111 & 0.063 \\
\rowcolor{tablestripe}
003 & Astra Direct & overlap  & 200 & 53  & 7.00\%  & 0.714 & 0.147 \\
\rowcolor{tablestripe}
& RL           & overlap  & 200 & 200 & 51.00\% & 0.104 & 0.072 \\
\rowcolor{white}
004 & Astra Direct & held out & 200 & 59  & 2.50\%  & 0.592 & 0.117 \\
\rowcolor{white}
& RL           & held out & 200 & 200 & 69.50\% & 0.073 & 0.061 \\
    \bottomrule
  \end{tabular}
\end{table*}

\begin{figure*}[htb]
  \centering
  \includegraphics[width=\textwidth]{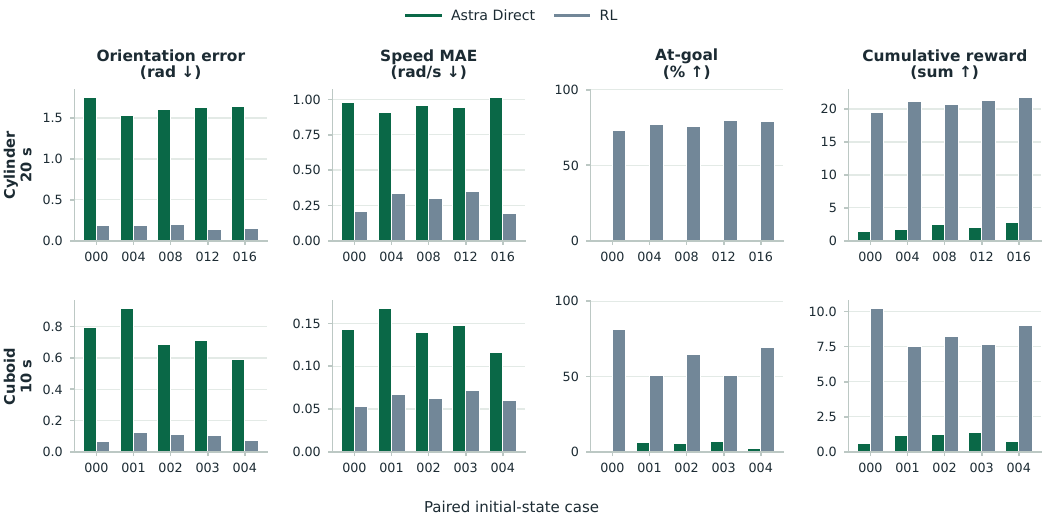}
  \caption{Per-case in-hand rotation performance. Columns report mean orientation error, axial-speed MAE, At-goal rate, and cumulative environment reward. Cylinder bars use all executed steps within the 20-s horizon; cuboid bars use the complete 10-s window. Reward is comparable only within each task row. Early termination and recording lengths are reported in Tables~\ref{tab:supp-cylinder-cases} and~\ref{tab:supp-cuboid-cases}.}
  \label{fig:supp-dexterous-per-case}
\end{figure*}

\begin{figure*}[htb]
  \centering
  \includegraphics[width=\textwidth]{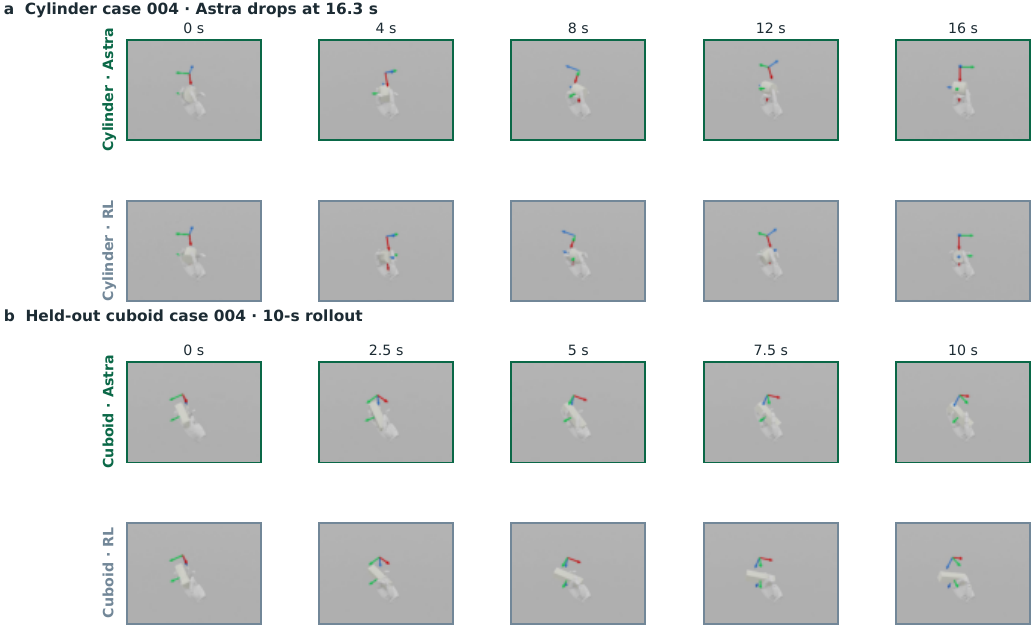}
  \caption{\textbf{Paired rotation sequences from identical saved starts.} (a) Cylinder case 004: Astra drops the object at 16.3\,s; RL completes 20\,s. (b) Cuboid case 004: Astra rotates gradually with little contact reconfiguration, while RL continues reorientation through 10\,s. Axes above the hand and on the object indicate world and object frames, respectively.}
  \label{fig:supp-dexterous-sequences}
\end{figure*}

Figures~\ref{fig:supp-dexterous-per-case} and~\ref{fig:supp-dexterous-sequences} support the contact analysis in Section~\ref{sec:dexterous-manipulation}. Contact transitions are assessed qualitatively from the trajectories.

Table~\ref{tab:supp-translation-cases} records no object drops during translation.

\begin{table*}[htb]
\caption{\textbf{Cylinder translation at the complete 15-s endpoint.} Pass denotes terminal position error below $22.4\,\mathrm{mm}$.}
\label{tab:supp-translation-cases}

  \centering
  
  \small
  \setlength{\tabcolsep}{19.38pt}
  \begin{tabular}{llrrrc}
    \toprule
    \rowcolor{tablehead}
\reporthead{Sample} & \reporthead{Controller} & \reporthead{Steps} & \reporthead{Decisions} & \reporthead{Final error (mm) $\downarrow$} & \reporthead{Outcome}\\
    \midrule
    \rowcolor{white}
000 & Astra Direct & 300 & 102 & 74.73  & Fail \\
\rowcolor{white}
& RL           & 300 & 300 & \textbf{10.02}  & \textbf{Pass} \\
\rowcolor{tablestripe}
001 & Astra Direct & 300 & 79  & \textbf{19.13}  & \textbf{Pass} \\
\rowcolor{tablestripe}
& RL           & 300 & 300 & 31.12  & Fail \\
\rowcolor{white}
002 & Astra Direct & 300 & 101 & 61.67  & Fail \\
\rowcolor{white}
& RL           & 300 & 300 & \textbf{9.84}   & \textbf{Pass} \\
\rowcolor{tablestripe}
003 & Astra Direct & 300 & 100 & 37.68  & Fail \\
\rowcolor{tablestripe}
& RL           & 300 & 300 & \textbf{14.99}  & \textbf{Pass} \\
\rowcolor{white}
004 & Astra Direct & 300 & 104 & 102.06 & Fail \\
\rowcolor{white}
& RL           & 300 & 300 & \textbf{20.50}  & \textbf{Pass} \\
    \bottomrule
  \end{tabular}
\end{table*}

\begin{table*}[htb]
\caption{\textbf{Translation-and-rotation terminal results.} All Astra endpoints occur after 120 decisions. Error pairs are position in millimeters / long-axis rotation in degrees. Outcomes are listed for Astra / matched RL / full-horizon RL.}
\label{tab:supp-translation-rotation-cases}

  \centering
  
  \small
  \setlength{\tabcolsep}{14.01pt}
  \begin{tabular}{crcccl}
    \toprule
    \rowcolor{tablehead}
\reporthead{Sample} & \reporthead{Astra steps} & \reporthead{Astra endpoint} & \reporthead{RL matched} & \reporthead{RL full, 15 s} & \reporthead{Outcome (A / M / F)}\\
    \midrule
    \rowcolor{white}
000 & 153 & 51.22 / 38.77 & 18.84 / 3.22 & 19.37 / 3.39 & Fail / \textbf{Pass} / \textbf{Pass} \\
\rowcolor{tablestripe}
001 & 129 & 38.12 / 28.56 & 15.53 / 3.57 & 15.38 / 3.55 & Fail / \textbf{Pass} / \textbf{Pass} \\
\rowcolor{white}
002 & 134 & 49.68 / 35.16 & 21.07 / 21.70 & 20.94 / 5.71 & Fail / Fail / \textbf{Pass} \\
\rowcolor{tablestripe}
003 & 120 & 50.60 / 27.41 & 10.42 / 9.56 & 11.26 / 1.87 & Fail / \textbf{Pass} / \textbf{Pass} \\
\rowcolor{white}
004 & 132 & 47.15 / 33.67 & 15.22 / 3.53 & 13.45 / 2.91 & Fail / \textbf{Pass} / \textbf{Pass} \\
    \bottomrule
  \end{tabular}
\end{table*}

\begin{figure*}[htb]
  \centering
  \includegraphics[width=\textwidth]{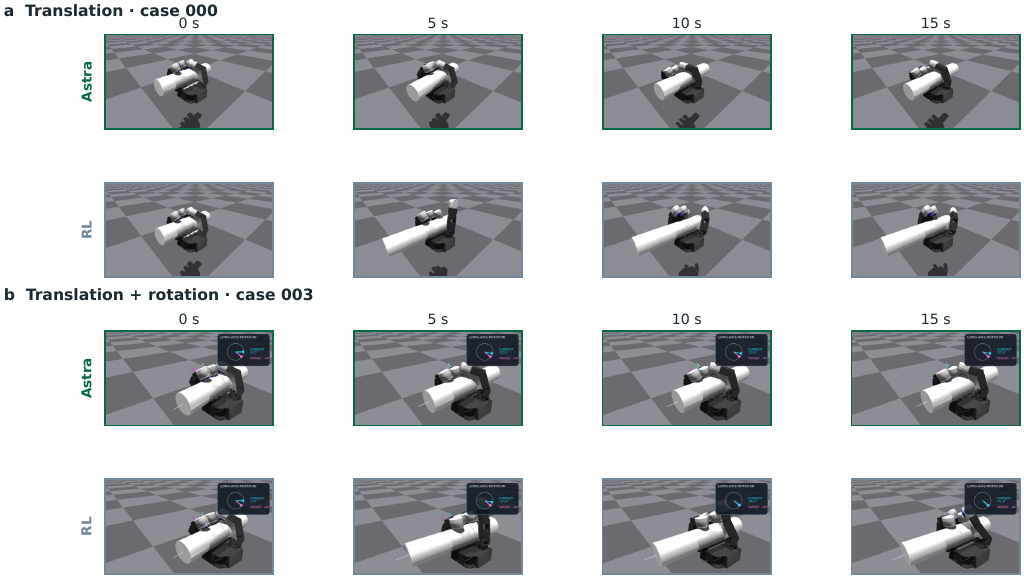}
  \caption{Qualitative Allegro rollouts from paired initial states. (a) Translation case 000. (b) Translation-and-rotation case 003, with a $-40^\circ$ long-axis target. Each row shows the same front view at 0, 5, 10, and 15 s. Astra reaches its 120-decision endpoint at 6.0 s in (b), so its terminal frame remains displayed while RL continues to 15 s.}
  \label{fig:supp-dexterous-allegro-sequences}
\end{figure*}

The terminal-error plots appear in Figure~\ref{fig:supp-dexterous-translation-results}; Figure~\ref{fig:supp-dexterous-allegro-sequences} adds representative rollout sequences.

%% file: sec/Z_navigation_mobile_suppl.tex
\section{Navigation and Mobile Manipulation: Protocols and Cases}
\label{app:navigation-mobile}

\subsection{Navigation Coverage and Termination}

The four navigation subsets each contain 50 episodes. The VLN-CE subsets of R2R and RxR distribute four or five episodes per scene across 11 scenes. The ObjectNav subsets balance scenes and goal categories: MP3D has two or three episodes per category across 21 categories; HM3D has eight or nine across six categories. The available source pools contain 1,839 R2R, 3,669 English-guide RxR, 2,195 MP3D, and 1,000 HM3D episodes. Balancing changes the source episode distribution, so the reported means characterize these evaluation subsets.

\begin{table}[htb]
\caption{\textbf{Navigation termination and travel.} Each row has 50 episodes. Steps count primitive turns, forward moves, and STOP, rather than model requests or waypoint submissions. Mean path measures translation; turning consumes steps without adding equivalent path length.}
\label{tab:navigation-behavior}

\centering\small
\setlength{\tabcolsep}{9.79pt}
\begin{tabular}{lrrrrr}
\toprule
\rowcolor{tablehead}
\reporthead{Task / dataset} & \reporthead{Successful STOP} & \reporthead{Unsuccessful STOP} & \reporthead{Step limit} & \reporthead{Mean steps} & \reporthead{Mean path (m)}\\
\midrule
\input{tables/navigation-behavior.tex}
\bottomrule
\end{tabular}

\end{table}

SPL~\cite{anderson2018evaluation} combines success $S$ and path efficiency as $S d/\max(d,\ell)$, where $d$ is initial shortest-path distance and $\ell$ is traveled distance. Unsuccessful episodes contribute zero; averages use all 50 episodes in the subset. VLN-CE route agreement follows nDTW and sDTW~\cite{ilharco2019dtw}, computed using FastDTW~\cite{salvador2007fastdtw}; sDTW additionally requires success. ObjectNav uses annotated viewpoints, and HM3D's 1\,m results rescore saved trajectories rather than rerunning the policy.

\input{sec/Z_navigation_baselines}

\subsection{RoboCasa Task Results and Paired Outcomes}

\label{app:robocasa-protocol}

\begin{table}[htb]
\caption{\textbf{All RoboCasa tasks and control horizons.} Every method uses five matched initial states per task. Horizons count 20\,Hz simulator steps. A single success changes a task's observed rate by 20 percentage points.}
\label{tab:robocasa-tasks}

\centering\small
\setlength{\tabcolsep}{14.1pt}
\begin{tabular}{llrrrr}
\toprule
\rowcolor{tablehead}
\reporthead{Group} & \reporthead{Task} & \reporthead{Horizon} & \reporthead{\pifive alone} & \reporthead{Direct} & \reporthead{Hybrid}\\
\midrule
\rowcolor{white}
\input{tables/robocasa-tasks.tex}
\midrule
All & 15 tasks & -- & 17/75 & 25/75 & 29/75 \\
\bottomrule
\end{tabular}

\end{table}

The task sampler sorts each frozen candidate group by \texttt{sha256(20260915|group|task)}, selects the first five tasks, and retains episodes 0--4 and their seeds. The ``seen'' labels refer to the local policy-training partition. All environments use \texttt{split=pretrain}; task composition and kitchen-layout split are different dimensions. Horizons are frozen local reference durations multiplied by 1.5.

\begin{table}[htb]
\caption{\textbf{Paired outcomes on 75 shared starts.} Hybrid adds optional \pifive. Two-sided exact McNemar tests~\cite{fagerland2013mcnemar} use discordant pairs. Exploratory $p$-values are unadjusted for multiple comparisons and account for neither within-task dependence nor harness differences.}
\label{tab:robocasa-paired}

\centering\small
\setlength{\tabcolsep}{15.11pt}
\begin{tabular}{lrrrrr}
\toprule
\rowcolor{tablehead}
\reporthead{Comparison ($A$ vs. $B$)} & \reporthead{Both succeed} & \reporthead{$A$ only} & \reporthead{$B$ only} & \reporthead{Both fail} & \reporthead{Exact $p$}\\
\midrule
\input{tables/robocasa-paired.tex}
\bottomrule
\end{tabular}

\end{table}

The 16 pairs where both Astra conditions succeed use an average of 1,110.4 steps with Hybrid and 1,256.6 with Direct; Hybrid uses fewer steps in 11 of these pairs. This duration comparison covers the jointly successful pairs. For the atomic-group comparison between Hybrid and Direct, a Bonferroni correction~\cite{dunn1961} over the three task groups raises the nominal $p$-value from 0.03125 to 0.09375.

\paragraph{Additional policy evaluation.}
An additional evaluation of the multitask checkpoint reports 403/2,500 successes across 50 tasks, or 16.12\%, and 132/750 on the matching 15-task subset, or 17.6\%. Table~\ref{tab:robocasa-results} uses the local baseline with shared initial states and a 20-step replanning cadence.

\subsection{RoboCasa Control Configurations, Information Access, and Resources}
\label{app:robocasa-controls}

\paragraph{Runtime and action semantics.}
The recorded environment uses RoboCasa 1.0.0~\cite{nasiriany2026robocasa365}, robosuite 1.5.2~\cite{zhu2020robosuite}, MuJoCo 3.3.1~\cite{todorov2012mujoco}, and Codex CLI 0.153.4. The policy is \texttt{pi05\_pretrain\_human300/multitask\_learning/75000}, with prediction horizon 50 and zero history images. The controller is the native \texttt{OperationalSpaceController}, based on operational-space control~\cite{khatib1987operational}. Pose quaternions use $xyzw$ order. Base and torso controls are normalized to $[-1,1]$; selecting arm/base mode does not automatically zero the other channels. Policy acceptance preserves its native continuous action values, whereas Direct exposes open/close gripper commands.

\paragraph{Harness configurations.}
Astra uses \texttt{xhigh} reasoning and 20-step prefixes. Resumed trajectories can span harness configurations. The conservative version compacts context at 96,000 tokens, summarizes policy proposals, and prefetches them. The guarded version also checks rendering and permits at most one action call per response, requiring intervening feedback. Runs use different hardware and service channels.

\begin{table}[!ht]
\caption{\textbf{Harness configuration coverage for the two Astra conditions.} Counts identify the configuration at the end of each trajectory; resumed trajectories may span multiple configurations. The independent policy does not use Astra context or proposal-review tools.}

\centering\small
\setlength{\tabcolsep}{22.37pt}
\begin{tabular}{lrr}
\toprule
\rowcolor{tablehead}
\reporthead{Harness configuration} & \reporthead{Direct} & \reporthead{Hybrid}\\
\midrule
\rowcolor{white}
Standard, \texttt{xhigh}, 20 steps & 2 & 19 \\
\rowcolor{tablestripe}
Conservative optimization & 0 & 8 \\
\rowcolor{white}
Conservative optimization with guards & 73 & 48 \\
\bottomrule
\end{tabular}

\end{table}

\paragraph{Information-boundary audit.}
An audit of 150 Astra sessions found no tool access to hidden object truth, complete physics state, reference trajectories, or action-guiding scores. It verifies mode-specific tools and zero policy queries or acceptances in Direct. Proprioception is supplied; pretraining is outside the audit. All 75 policy-only records retain proposals, actions, observations, outcomes, and matching initial-state fingerprints.

\begin{table}[!ht]
\caption{\textbf{Execution and request accounting.} Policy-only prefixes execute automatically, without Astra acceptance. Retained-chain requests differ from upstream requests and action segments. Counts cover saved request logs, including recovery history, and exclude discarded attempts; they are not totals over all inference calls.}
\label{tab:robocasa-cost}

\centering\small
\setlength{\tabcolsep}{18.36pt}
\begin{tabular}{lrrr}
\toprule
\rowcolor{tablehead}
\reporthead{Quantity} & \reporthead{\pifive alone} & \reporthead{Direct} & \reporthead{Hybrid}\\
\midrule
\rowcolor{white}
Episodes & 75 & 75 & 75 \\
\rowcolor{tablestripe}
Executed simulator steps & 140,958 & 130,665 & 130,549 \\
\rowcolor{white}
Policy proposals & 7,072 & 0 & 6,185 \\
\rowcolor{tablestripe}
Astra-authored action segments & 0 & 6,560 & 2,941 \\
\rowcolor{white}
Accepted policy segments & -- & 0 & 3,621 \\
\rowcolor{tablestripe}
Recorded model requests, retained chains & 0 & 7,910 & 8,941 \\
\rowcolor{white}
Median recorded model requests per episode & 0 & 85 & 108 \\
\bottomrule
\end{tabular}

\end{table}

Policy-only execution takes a median 121.6\,s after loading and scene reconstruction, excluding queueing, initialization, and infrastructure time. Astra timing includes inference and recovery waits. Both Astra conditions pause physics during reasoning; see Section~\ref{sec:resources}.

\paragraph{Reproducibility.}
Archives retain outcomes, initial-state IDs, audits, images, and checks on all 75 state triples.

%% file: tables/navigation-behavior.tex
\rowcolor{white}
VLN-CE / R2R & 39 & 11 & 0 & 85.2 & 12.03 \\
\rowcolor{tablestripe}
VLN-CE / RxR & 46 & 4 & 0 & 98.1 & 12.95 \\
\rowcolor{white}
ObjectNav / MP3D & 28 & 12 & 10 & 220.8 & 26.58 \\
\rowcolor{tablestripe}
ObjectNav / HM3D & 41 & 4 & 5 & 163.2 & 16.80 \\

%% file: sec/Z_navigation_baselines.tex
\Needspace*{5\baselineskip}
\subsection{Released Navigation Policies: Interfaces and Complete Results}
\label{app:navigation-open-models}

\paragraph{Fixed tasks and checkpoint selection.}
The released LightNav-0~\cite{wang2026lightnav}, Uni-NaVid~\cite{zhang2025uninavid}, and OmniNav Flow~\cite{xue2025omninav} policies each complete the same 200 episodes as Astra, without resampling or outcome-based exclusion. Source commits and checkpoints are fixed before evaluation. An eight-episode interface check tests valid execution, not navigation success; its outcomes are excluded. The archives retain code, manifests, and episode outputs.

\begin{table}[htb]
\caption{\textbf{Source commits and released checkpoints.} LightNav-0 and Uni-NaVid use their official Hugging Face releases; OmniNav uses the official ModelScope Flow release. Uni-NaVid is the \texttt{uninavid-7b-full-224-video-fps-1-grid-2} checkpoint. The open models are evaluated without local finetuning.}
\label{tab:navigation-open-revisions}

\centering\small
\setlength{\tabcolsep}{22.67pt}
\begin{tabular}{lll}
\toprule
\rowcolor{tablehead}
\reporthead{Released policy} & \reporthead{Source commit} & \reporthead{Checkpoint commit}\\
\midrule
\rowcolor{white}
LightNav-0 & \texttt{3015508b70fb} & \texttt{826dc5fbfa37} \\
\rowcolor{tablestripe}
Uni-NaVid 7B & \texttt{79ef5ea3fea1} & \texttt{0437222534b2} \\
\rowcolor{white}
OmniNav Flow & \texttt{e8b485953c65} & \texttt{f73a8d094f53} \\
\bottomrule
\end{tabular}

\end{table}

\paragraph{Observations and history.}
LightNav-0 and Uni-NaVid use the same $512\times384$, $90^\circ$ front camera at 0.88\,m as Astra. LightNav-0 uses its official trajectory prompt and aspect-preserving image preprocessing. Uni-NaVid uses its official visual processor and online feature cache, with intermediate RGB frames retained between two-action executions. All three released policies receive the simulator's route instruction or fixed goal-category instruction. Their input-history processing differs from Astra's persistent agent session.

OmniNav Flow receives real, simultaneous left, right, and front images; the side cameras are oriented $+90^\circ$ and $-90^\circ$ relative to the front camera. Each image is rendered at $720\times640$ with a $110^\circ$ horizontal field of view and camera height 0.88\,m, then processed by the released model's image and history pipeline. Side views do not require additional turns or consume primitive steps. No black-image substitutes are used in these evaluation episodes. The policy receives neither simulator pose nor evaluation distance. 

\paragraph{Action conversion and stopping.}
All policies use the primitives, 500-step budget, sliding settings, and STOP criteria in Section~\ref{sec:navigation}. Collisions interrupt the proposed sequence; budget exhaustion does not force STOP.

LightNav-0's residual-vector-quantized output is decoded into local future poses. The adapter selects the first pose with a representable forward or yaw component, rounds it to primitive actions, and executes turns before forward motion, bounded by 2\,m and $180^\circ$. The lateral component is not executed directly. A decoded zero trajectory yields STOP. A nonzero trajectory below primitive resolution produces one attainable primitive along its dominant component.

Uni-NaVid retains its VLN action template and sampling temperature 0.2, requests four actions, and executes the first two, stopping the sequence if STOP occurs. Forward, left, and right map directly to the benchmark primitives. Its visual cache resets at each episode. The released implementation notes that the L3MVN-derived ObjectNav imitation-learning component is omitted; our category-goal evaluation uses this public checkpoint.\footnote{\url{https://github.com/jzhzhang/Uni-NaVid}}

OmniNav Flow predicts five local waypoints, yaw angles, and arrival scores. The adapter uses the first waypoint's planar length and predicted yaw, quantizes them to the common primitives, and executes turns before forward motion with a 2\,m limit. A non-arrival output below action resolution produces one minimum primitive. STOP requires all five arrival scores to exceed 0.5. The evaluation omits the upstream loop's goal-distance-based stagnation rule and forced stopping. This condition uses the released vision-only Flow policy; it does not instantiate the paper's slow exploration planner. Invalid outputs raise execution errors rather than being converted to STOP; no such errors occur in the 600 completed episodes.

\paragraph{Evaluation configuration.}
The policies retain their released prompts and history processing. We replace the published LightNav-0 and Uni-NaVid camera settings with the shared front camera and discretize continuous outputs under the protocol in Section~\ref{sec:navigation}.

\paragraph{Published reference sources and coverage.}
Tables~\ref{tab:navigation-published-vln} and~\ref{tab:navigation-published-objectnav} use Qwen-RobotNav Tables~1 and~4, LightNav-0 Tables~III and~IV, NavFoM Table~1, Uni-NaVid Tables~II and~III, OmniNav Table~1, and CogNav Table~1~\cite{qwenrobotnav2026,wang2026lightnav,zhang2026embodied,zhang2025uninavid,xue2025omninav,cao2024cognav}. These are reported benchmark-split results under the published protocols. The source papers do not enumerate exact episode counts for every row. CogNav, for example, describes an MP3D validation set of approximately 1,800 episodes and 20 categories, whereas our 50 episodes are drawn from a 2,195-episode, 21-category pool. We therefore preserve the reported coverage rather than assign a common denominator. Qwen-RobotNav's closed-vocabulary ObjectNav results are marked RGB without assuming a camera count from its separate OVON experiment. The literature table excludes HM3D v1 and HM3D-OVON values from the HM3D v2 column.

\begin{table}[htb]
\caption{\textbf{Route agreement and termination on every fixed subset.} Each system--dataset row contains 50 episodes. nDTW and sDTW are percentages and apply only to VLN-CE. Successes, failed STOP, and step-limit terminations sum to 50 in every row. OmniNav Flow uses the separate three-view observation condition.}
\label{tab:navigation-open-diagnostics}

\centering\small
\setlength{\tabcolsep}{8.60pt}
\begin{tabular}{llrrrrr}
\toprule
\rowcolor{tablehead}
\reporthead{Task / dataset} & \reporthead{System} & \reporthead{Successes} & \reporthead{nDTW} & \reporthead{sDTW} & \reporthead{Failed STOP} & \reporthead{Step limit}\\
\midrule
\input{tables/navigation-open-diagnostics.tex}
\bottomrule
\end{tabular}

\end{table}

\begin{figure*}[htb]
\centering
\includegraphics[width=\textwidth]{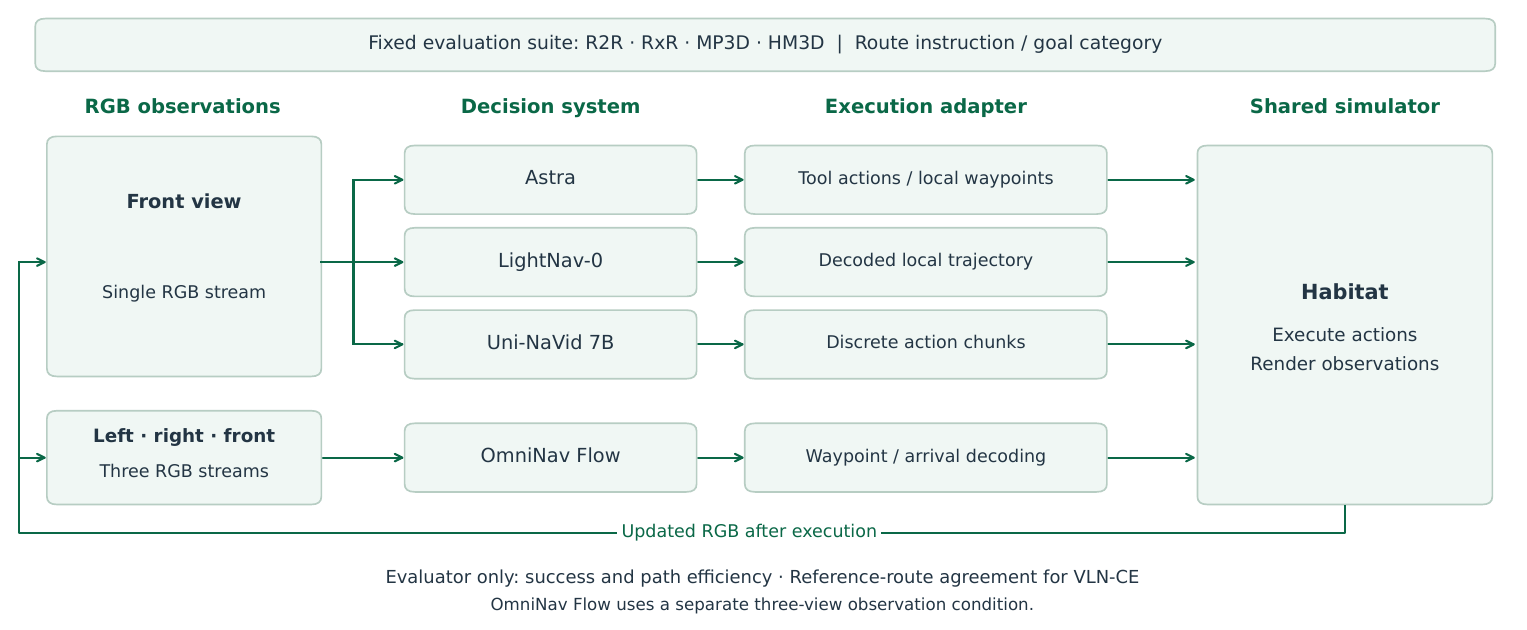}
\caption{\textbf{Navigation observation and execution adapters.} Astra, LightNav-0, and Uni-NaVid share the front-camera specification; OmniNav Flow uses three views. Habitat executes adapted outputs and returns RGB observations. Goal distances and metrics remain evaluator-only.}
\label{fig:navigation-open-models}
\end{figure*}

\input{tables/navigation-published-reference}

%% file: tables/navigation-open-diagnostics.tex
\rowcolor{white}
VLN-CE / R2R & Astra & 39/50 & 72.20 & 59.35 & 11 & 0 \\
\rowcolor{tablestripe}
VLN-CE / R2R & LightNav-0 & 29/50 & 72.77 & 48.89 & 21 & 0 \\
\rowcolor{white}
VLN-CE / R2R & Uni-NaVid 7B & 15/50 & 61.76 & 26.38 & 14 & 21 \\
\rowcolor{tablestripe}
VLN-CE / R2R & OmniNav Flow (3 views) & 28/50 & 71.18 & 49.23 & 22 & 0 \\
\midrule
\rowcolor{white}
VLN-CE / RxR & Astra & 46/50 & 84.73 & 80.42 & 4 & 0 \\
\rowcolor{tablestripe}
VLN-CE / RxR & LightNav-0 & 36/50 & 77.44 & 64.89 & 14 & 0 \\
\rowcolor{white}
VLN-CE / RxR & Uni-NaVid 7B & 17/50 & 56.34 & 27.73 & 19 & 14 \\
\rowcolor{tablestripe}
VLN-CE / RxR & OmniNav Flow (3 views) & 36/50 & 68.51 & 57.34 & 13 & 1 \\
\midrule
\rowcolor{white}
ObjectNav / MP3D & Astra & 28/50 & -- & -- & 12 & 10 \\
\rowcolor{tablestripe}
ObjectNav / MP3D & LightNav-0 & 10/50 & -- & -- & 6 & 34 \\
\rowcolor{white}
ObjectNav / MP3D & Uni-NaVid 7B & 4/50 & -- & -- & 17 & 29 \\
\rowcolor{tablestripe}
ObjectNav / MP3D & OmniNav Flow (3 views) & 2/50 & -- & -- & 33 & 15 \\
\midrule
\rowcolor{white}
ObjectNav / HM3D & Astra & 41/50 & -- & -- & 4 & 5 \\
\rowcolor{tablestripe}
ObjectNav / HM3D & LightNav-0 & 33/50 & -- & -- & 2 & 15 \\
\rowcolor{white}
ObjectNav / HM3D & Uni-NaVid 7B & 15/50 & -- & -- & 5 & 30 \\
\rowcolor{tablestripe}
ObjectNav / HM3D & OmniNav Flow (3 views) & 3/50 & -- & -- & 17 & 30 \\

%% file: tables/navigation-published-reference.tex
\begin{table*}[htbp]
\caption{\textbf{VLN-CE: subset-to-full-benchmark reference comparison.} Published rows report R2R and RxR Val-Unseen results under the published protocols; the Astra row uses only 50 fixed episodes per dataset. Values are percentages. Each row retains its source protocol for observations, actions, and scoring. Full-benchmark scope follows each paper; exact episode counts are not supplied for every reference.}
\label{tab:navigation-published-vln}

\centering\small
\setlength{\tabcolsep}{18.48pt}
\begin{tabular}{llrrrr}
\toprule
\rowcolor{tablehead}
&   &  \multicolumn{2}{c}{\reporthead{R2R}}  &  \multicolumn{2}{c}{\reporthead{RxR}} \\
\cmidrule(lr){3-4}\cmidrule(lr){5-6}
\rowcolor{tablehead}
\reporthead{Method} & \reporthead{Observations} & \reporthead{SR} & \reporthead{SPL} & \reporthead{SR} & \reporthead{SPL}\\
\midrule
\rowcolor{tablegroup}
\multicolumn{6}{l}{\emph{Published full-benchmark references (published protocols)}} \\
\rowcolor{tablestripe}
Uni-NaVid~\cite{zhang2025uninavid} & 1 RGB & 47.0 & 42.7 & 48.7 & 40.9 \\
\rowcolor{white}
NavFoM~\cite{zhang2026embodied} & 1 RGB & 56.2 & 51.2 & 57.4 & 49.4 \\
\rowcolor{tablestripe}
LightNav-0~\cite{wang2026lightnav} & 1 RGB & 68.5 & 62.8 & 73.6 & 64.5 \\
\rowcolor{white}
Qwen-RobotNav-4B~\cite{qwenrobotnav2026} & 1 RGB & 66.9 & 60.5 & 71.3 & 61.5 \\
\rowcolor{tablestripe}
Qwen-RobotNav-8B~\cite{qwenrobotnav2026} & 1 RGB & 65.7 & 59.6 & 73.4 & 63.5 \\
\rowcolor{white}
NavFoM~\cite{zhang2026embodied} & 4 RGB & 61.7 & 55.3 & 64.4 & 56.2 \\
\rowcolor{tablestripe}
OmniNav~\cite{xue2025omninav} & Multi-view RGB & 69.5 & 66.1 & 73.6 & 62.0 \\
\rowcolor{white}
Qwen-RobotNav-8B~\cite{qwenrobotnav2026} & Panoramic RGB & 72.1 & 66.6 & 76.5 & 65.7 \\
\midrule
\rowcolor{tablegroup}
\multicolumn{6}{l}{\emph{This work: fixed subsets, 50 episodes per dataset}} \\
\rowcolor{white}
\textbf{Astra (this work)} & 1 RGB & 78.00 & 65.27 & 92.00 & 77.25 \\
\bottomrule
\end{tabular}

\end{table*}

\begin{table*}[htbp]
\caption{\textbf{Object-goal navigation: subset-to-full-benchmark reference comparison.} Published rows report full validation-benchmark results; the Astra row uses only 50 fixed episodes per dataset. Values are percentages. Coverage and protocols differ, as in Table~\ref{tab:navigation-published-vln}. RGB without a view count preserves the source\textquotesingle{}s unspecified camera configuration. A dash denotes no result for that dataset/version; HM3D v1 and HM3D-OVON scores are excluded from HM3D v2.}
\label{tab:navigation-published-objectnav}

\centering\small
\setlength{\tabcolsep}{18.94pt}
\begin{tabular}{llrrrr}
\toprule
\rowcolor{tablehead}
&   &  \multicolumn{2}{c}{\reporthead{MP3D}}  &  \multicolumn{2}{c}{\reporthead{HM3D v2}} \\
\cmidrule(lr){3-4}\cmidrule(lr){5-6}
\rowcolor{tablehead}
\reporthead{Method} & \reporthead{Observations} & \reporthead{SR} & \reporthead{SPL} & \reporthead{SR} & \reporthead{SPL}\\
\midrule
\rowcolor{tablegroup}
\multicolumn{6}{l}{\emph{Published full-benchmark references (published protocols)}} \\
\rowcolor{tablestripe}
CogNav~\cite{cao2024cognav} & RGB-D + pose & 46.6 & 16.1 & -- & -- \\
\rowcolor{white}
LightNav-0~\cite{wang2026lightnav} & 1 RGB & 53.3 & 21.2 & 79.5 & 43.7 \\
\rowcolor{tablestripe}
Qwen-RobotNav-4B~\cite{qwenrobotnav2026} & RGB & 52.2 & 16.0 & 75.6 & 30.6 \\
\rowcolor{white}
Qwen-RobotNav-8B~\cite{qwenrobotnav2026} & RGB & 48.8 & 17.7 & 71.2 & 33.0 \\
\midrule
\rowcolor{tablegroup}
\multicolumn{6}{l}{\emph{This work: fixed subsets, 50 episodes per dataset}} \\
\rowcolor{white}
\textbf{Astra (this work)} & 1 RGB & 56.00 & 22.43 & 82.00 & 43.69 \\
\bottomrule
\end{tabular}

\end{table*}

%% file: tables/robocasa-tasks.tex
\rowcolor{white}
Atomic seen & CoffeeSetupMug & 600 & 3/5 & 0/5 & 0/5 \\
\rowcolor{white}
& OpenDrawer & 750 & 3/5 & 2/5 & 3/5 \\
\rowcolor{white}
& OpenStandMixerHead & 450 & 1/5 & 0/5 & 2/5 \\
\rowcolor{white}
& PickPlaceDrawerToCounter & 750 & 2/5 & 0/5 & 3/5 \\
\rowcolor{white}
& PickPlaceSinkToCounter & 900 & 4/5 & 5/5 & 5/5 \\
\midrule
\rowcolor{tablestripe}
Composite seen & DeliverStraw & 2550 & 0/5 & 1/5 & 1/5 \\
\rowcolor{tablestripe}
& KettleBoiling & 1500 & 0/5 & 1/5 & 3/5 \\
\rowcolor{tablestripe}
& PackIdenticalLunches & 3900 & 1/5 & 1/5 & 2/5 \\
\rowcolor{tablestripe}
& SearingMeat & 4350 & 0/5 & 1/5 & 1/5 \\
\rowcolor{tablestripe}
& WashLettuce & 1650 & 2/5 & 0/5 & 0/5 \\
\midrule
\rowcolor{white}
Composite unseen & BreadSelection & 1950 & 0/5 & 4/5 & 2/5 \\
\rowcolor{white}
& GatherTableware & 2250 & 0/5 & 0/5 & 1/5 \\
\rowcolor{white}
& HeatKebabSandwich & 2700 & 0/5 & 2/5 & 1/5 \\
\rowcolor{white}
& MakeIceLemonade & 3000 & 0/5 & 3/5 & 2/5 \\
\rowcolor{white}
& RecycleBottlesByType & 2850 & 1/5 & 5/5 & 3/5 \\

%% file: tables/robocasa-paired.tex
\rowcolor{white}
Hybrid vs. Direct & 16 & 13 & 9 & 37 & 0.5235 \\
\rowcolor{tablestripe}
Hybrid vs. $\pi_{0.5}$ & 11 & 18 & 6 & 40 & 0.0227 \\
\rowcolor{white}
Direct vs. $\pi_{0.5}$ & 7 & 18 & 10 & 40 & 0.1849 \\